\documentclass[letterpaper, times]{elsarticle}
\makeatletter
\def\ps@pprintTitle{%
 \let\@oddhead\@empty
 \let\@evenhead\@empty
 \def\@oddfoot{\thepage\hfill\footnotesize\itshape\today}
 \let\@evenfoot\@oddfoot}
\makeatother

\usepackage[margin=2.54cm]{geometry}
\usepackage{amsthm}
\usepackage{xcolor}
\usepackage{mathtools}
\usepackage{graphicx}
\usepackage{amssymb}
\usepackage{caption}
\usepackage{subcaption}
\usepackage{tcolorbox}
\usepackage{scrextend}
\usepackage{listings}
\usepackage{bm}
\usepackage{float}
\usepackage[colorlinks]{hyperref}
\usepackage{multirow}
\usepackage[linesnumbered,lined,boxed,ruled,vlined]{algorithm2e}
\usepackage{afterpage}
\usepackage{etoolbox}
\usepackage{dutchcal}
\usepackage{makecell}
\usepackage{cleveref}
\usepackage{lineno}
\usepackage[pages=some,placement=top]{background}

\backgroundsetup{
	scale=1,
	color=black,
	opacity=1.0,
	angle=0,
	contents={%
 		\includegraphics[width=\paperwidth,height=2cm]{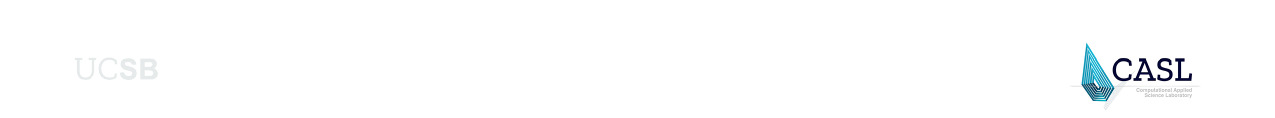}
	}%
}

\hypersetup{
	colorlinks=true,      		
	linkcolor=blue,				
	citecolor=blue,				
	filecolor=black,      		
	urlcolor=red,				
	bookmarks=false,
	pdffitwindow=true,
	pdfpagelayout=SinglePage
}

\SetAlCapNameFnt{\small}	
\SetAlCapFnt{\small}
\SetAlFnt{\small}

\definecolor{navy}{RGB}{2, 48, 71}
\definecolor{aqua}{RGB}{33, 158, 188}

\newcommand{\colorsection}[1]{\sffamily\color{navy}\section{#1}\color{black}\rmfamily}
\newcommand{\colorsubsection}[1]{\sffamily\color{navy}\subsection{#1}\color{black}\rmfamily}
\newcommand{\colorsubsubsection}[1]{\sffamily\color{navy}\subsubsection{#1}\color{black}\rmfamily}

\patchcmd{\abstract}{Abstract}{\sffamily{Abstract}\rmfamily}{}{}
\abstracttitle{\sffamily\textcolor{navy}{Abstract}}

\newcommand{\etal}{et al.\ }

\newcommand{\ode}[2]{ \frac{\mathrm{d}#1}{\mathrm{d}#2} }

\newcommand{\pde}[2]{ \frac{\partial #1}{\partial #2} }

\newcommand{\vv}[1]{ \boldsymbol{\mathbf{#1}} }

\newcommand{\eten}[2]{ #1\times 10^{#2} }

\let\OLDthebibliography\thebibliography
\renewcommand\thebibliography[1]{
  \OLDthebibliography{#1}
  \setlength{\parskip}{1pt}
  \setlength{\itemsep}{1pt plus 0.3ex}
}

\theoremstyle{remark}
\newtheorem{remark}{Remark}

\begin{document}

\title{\textcolor{aqua}{\sffamily\bfseries Error-correcting neural networks for semi-Lagrangian advection in the level-set method}}

\author[1]{\textcolor{gray}{Luis \'{A}ngel} Larios-C\'{a}rdenas\corref{cor1}}
\ead{lal@cs.ucsb.edu}

\author[1,2]{\textcolor{gray}{Fr\'{e}d\'{e}ric} Gibou}
\ead{fgibou@ucsb.edu}

\cortext[cor1]{Corresponding author}

\address[1]{\textcolor{darkgray}{Computer Science Department, University of California, Santa Barbara, CA 93106, USA}}
\address[2]{\textcolor{darkgray}{Mechanical Engineering Department, University of California, Santa Barbara, CA 93106, USA}}

\BgThispage


\begin{abstract}

We present a machine learning framework that blends image super-resolution technologies with passive, scalar transport in the level-set method.  Here, we investigate whether we can compute on-the-fly, data-driven corrections to minimize numerical viscosity in the coarse-mesh evolution of an interface.  The proposed system's starting point is the semi-Lagrangian formulation.  And, to reduce numerical dissipation, we introduce an error-quantifying multilayer perceptron.  The role of this neural network is to improve the numerically estimated surface trajectory.  To do so, it processes localized level-set, velocity, and positional data in a single time frame for select vertices near the moving front.  Our main contribution is thus a novel machine-learning-augmented transport algorithm that operates alongside selective redistancing and alternates with conventional advection to keep the adjusted interface trajectory smooth.  Consequently, our procedure is more efficient than full-scan convolutional-based applications because it concentrates computational effort only around the free boundary.  Also, we show through various tests that our strategy is effective at counteracting both numerical diffusion and mass loss.  In simple advection problems, for example, our method can achieve the same precision as the baseline scheme at twice the resolution but at a fraction of the cost.  Similarly, our hybrid technique can produce feasible solidification fronts for crystallization processes.  On the other hand, tangential shear flows and highly deforming simulations can precipitate bias artifacts and inference deterioration.  Likewise, stringent design velocity constraints can limit our solver's application to problems involving rapid interface changes.  In the latter cases, we have identified several opportunities to enhance robustness without forgoing our approach's basic concept.  Despite these circumstances, we believe all the above assets make our framework attractive to parallel level-set algorithms.  Its appeal resides in the possibility of avoiding further mesh refinement and decreasing expensive communications between computing nodes.
\end{abstract}

\begin{keyword}
\textcolor{gray}{\small machine learning \sep semi-Lagrangian advection \sep error modeling \sep neural networks \sep level-set method \sep Stefan problem}
\end{keyword}

\maketitle



\colorsection{Introduction}
\label{sec:Introduction}

In the last few years, we have witnessed unprecedented progress in scientific disciplines thanks to machine learning.  Advancements in these areas have sped up, especially with the democratization of computing resources and increased data accessibility.  In particular, thriving machine learning applications \cite{A18, Mehta19, Hands-onMLwithScikit-LearnKerasAndTF19} have emerged from computer vision and image processing \cite{Turk;Pentland;Eigenfaces;1991, SDenoisingAutoEncoders10, AlexNet12, Dong;Loy;He;SuperResolution;2014, GAdversarialNets14, U-Net;2015, ResNet16, NLC17-Ventricle-Segmentation, LevelSetAsDeepRNN18}, computer graphics \cite{Xie;etal;TempoGAN;2018}, natural language processing and information retrieval \cite{Word2Vec13, BGJM17-Word-Vectors-with-Subword-Info, Elmo18, ALM17-Sentence-Embeddings}, and sequencing and language translation \cite{LSTM;1997, GRU14, Transformer17}.  Not long ago, machine learning and neural networks began carving their paths through the classic numerical methodologies, leading to promising ramifications in mathematics, physics, and engineering \cite{IntroToNeuralMethodsForDiffEqns15}.  It all started with the groundbreaking work by Lagaris \etal \cite{NNetForODEsPDEs98, NNetForPDEs00} to approximate the solution to boundary-value problems.

More recent machine learning incursions in computational science have brought about a broad spectrum of novel applications.  Among these, physics-informed neural networks (PINNs) \cite{Raissi17a, Raissi18, Raissi;PINN;2019} have fueled a large body of scientific deep learning developments.   The motivation behind PINNs is to uncover the governing dynamics in spatiotemporal data sets (e.g., \cite{Hu;etal;DCSNN-elliptic-PDEs;2021, Pakravan;etal;SolvingInversePDEs;2021}).  Also, another group of techniques has found a niche in conservation-law simulations.  Contributions in this branch include classifiers for near-discontinuity regions \cite{TroubledCellIndicator18}, shock detectors for complex flows \cite{ShockDetector20}, and under-resolved-stencil discriminators for surface reconstruction \cite{Buhendwa;Bezgin;Adams;IRinLSwithML;2021}.  These models have assisted with balancing efficiency and accuracy by enabling their systems to switch between solvers depending on field smoothness and other attributes.

Practitioners, too, have turned to machine learning to address challenging tasks in free-boundary problems (FBPs) \cite{Friedman10}.  Some implicit formulations widely used for solving FBPs are the volume-of-fluid (VOF) \cite{Hirt;Nichols:81:Volume-of-Fluid-VOF-}, the phase-field \cite{Fix:1983aa, Langer:1986aa}, and the level-set methods \cite{Osher1988}.  In VOF technologies, we can find a representative combination of computational fluid dynamics (CFD) and machine learning in the work by Despr\'es and Jourdren with algorithms for bimaterial compressible Euler calculations \cite{DespresJourdren;MLDesignOfVOF;20}.  Similarly, Qi \etal \cite{CurvatureML19} and Patel \etal \cite{VOFCurvature3DML19} have optimized neural networks to estimate curvature at the center of two- and three-dimensional volume-fraction stencils.  Besides curvature computation, the interface geometry is likewise hard to characterize because of the discontinuous nature of the VOF representation.  Ataei \etal \cite{NPLIC20}, for example, have proposed a neural piece-wise linear interface-construction (NPLIC) system to deal with this difficulty.  Their alternative approach has outperformed the complex analytical PLIC procedures \cite{Youngs;VOF-PLIC;1982} while being nearly as accurate at faster evaluation rates.

The level-set method has also benefited from data-driven technologies.  In this formulation, researchers have actively sought efficacious solutions to long-standing difficulties derived from the absence of inbuilt mechanisms that secure well-balancedness \cite{Popinet;NumModelsOfSurfTension;18} and preserve mass \cite{MKZ98}.  Such problems can get exacerbated in coarse meshes and when the free boundary undergoes severe stretching or tearing \cite{Sussman;Puckett:00:A-Coupled-Level-Set-}.  First, well-balancedness relates to surface tension models and their ability to recover equilibrium solutions  \cite{Popinet;NumModelsOfSurfTension;18}.  It rests on estimating curvature accurately \textit{at} the interface, regardless of mesh size.  Although the level-set framework provides a straightforward relation to calculate curvature, it does not always yield satisfactory approximations when the level-set field lacks sufficient smoothness and regularity.  With this in mind, we have recently proposed network-only \cite{LALariosFGibou;LSCurvatureML;2021} and hybrid inference systems \cite{Larios;Gibou;HybridCurvature;2021} that tackle the curvature problem in low-resolution grids.  Our investigation has revealed that blending traditional schemes with multilayer perceptrons produces better results than taking each component alone.  In like manner, Buhendwa \etal \cite{Buhendwa;Bezgin;Adams;IRinLSwithML;2021} have developed a machine-learning interface-reconstruction (IR) strategy that computes volume fractions and apertures.  Their procedure has proven effective for under-resolved regions while recovering the conventional IR's accuracy for well-resolved sectors.

Often, artificial mass loss and numerical diffusion occur simultaneously.  Further, the coarser the mesh, the more that one loses mass to under-resolution.  Consequently, mesh refinement is the preferred option to minimize numerical dissipation.  In order to refine a grid efficiently, researchers have proposed nonuniform discretization schemes to increase the resolution only next to the interface \cite{Berger;Oliger:84:Adaptive-mesh-refine, Strain1999}.  Adaptive Cartesian grids, for instance, have served as the basis for high-order robust level-set algorithms and tools in both serial \cite{Min;Gibou:06:A-second-order-accur, Min;Gibou;Ceniceros:06:A-supra-convergent-f, Min;Gibou:07:A-second-order-accur, Chen;Min;Gibou:07:A-Supra-Convergent-F, Min;Gibou:07:Geometric-integratio, Min;Gibou:08:Robust-second-order-, Chen;Min;Gibou:09:A-numerical-scheme-f} and distributed computing systems \cite{Burstedde;Wilcox;Ghattas:11:p4est:-Scalable-Algo, Mirzadeh;etal:16:Parallel-level-set}.  Over the years, practitioners have also combined level-set technologies with other numerical frameworks to improve mass conservation.  Among these hybrid approaches, we can find the coupled level-set and volume-of-fluid \cite{Sussman;Puckett:00:A-Coupled-Level-Set-}, the particle level-set \cite{Enright;Fedkiw;Ferziger;etal:02:A-Hybrid-Particle-Le}, the coupled volume-of-fluid and level-set \cite{VOSET;2010}, and the marker level-set \cite{MarkerLevelSet;2007} methods.  Other computational scientists have opted for in-place augmentation by either adding volume reinitialization \cite{Salih;Ghosh;LS-VolReinit;2013} or incorporating source/sink terms \cite{Yuan;etal;LS-Source-Sink;2018} into the standard formulation.  However, little has been done to investigate mass loss from a data-driven perspective.  In this manuscript, we bridge the level-set method with machine learning to offset numerical viscosity and preserve mass in low-resolution grids.

The present study draws inspiration from image super-resolution methodologies.  The image super-resolution problem involves learning an end-to-end mapping between low- and high-resolution images \cite{Dong;Loy;He;SuperResolution;2014}.  In scientific computing, this concept is strongly connected with training low-resolution models to imitate the rules in their high-resolution counterparts \cite{Bar-Sinai;Hoyer;Hickey;Brenner;LearningData-DrivenPDEs;2019, Zhuang;etal;LrndDiscForPassSclrAdvctn2D;2021}.  This idea, in particular, offers promising opportunities to circumvent the prohibitive costs of solving PDEs in grids with considerably small mesh sizes.  In computer graphics, for instance, Xie \etal \cite{Xie;etal;TempoGAN;2018} have designed temporally coherent generative adversarial neural networks.  Their networks learn to produce volumetric frames from a low-resolution field containing passively advected density and velocity data.  Likewise, Liu \etal \cite{Liu;etal;DLMthdsSuperResoltnReconstTurbFlows;2020} have reconstructed turbulent flows from coarse spatiotemporal data with the aid of convolutional and multi-scale residuals blocks.  Other influential developments in CFD include the learned discretizations of \cite{Bar-Sinai;Hoyer;Hickey;Brenner;LearningData-DrivenPDEs;2019} and \cite{Zhuang;etal;LrndDiscForPassSclrAdvctn2D;2021}.  In \cite{Zhuang;etal;LrndDiscForPassSclrAdvctn2D;2021}, Zhuang \etal extend their data-driven discretizations in \cite{Bar-Sinai;Hoyer;Hickey;Brenner;LearningData-DrivenPDEs;2019} to two-dimensional passive scalar advection in turbulent flows.  Instead of using predefined finite-volume coefficients, Zhuang \etal have trained convolutional neural networks that emit optimal coefficients based on high-resolution simulations.  Their machine learning approach thus keep an exceptional level of accuracy, even when the solution is under-resolved with classic methods.  Also, Pathak \etal \cite{Pathak;etal;MLToAugCoarseGridCFD;2020} have developed a PDE-machine-learning strategy to enhance the accuracy of coarse high-$Re$ turbulent-flow simulations.  Their framework's key component is an \textit{error-correcting} U-Net \cite{U-Net;2015} convolutional network.  This architecture helps restore fine-scale details in a high-resolution estimate of system variables and simultaneously corrects the error introduced during the coarse-grid simulation.

In this manuscript, we introduce a data-driven strategy that extends the notions in \cite{Zhuang;etal;LrndDiscForPassSclrAdvctn2D;2021} and \cite{Pathak;etal;MLToAugCoarseGridCFD;2020} to the level-set method.  More precisely, we seek to minimize numerical viscosity in the coarse-mesh evolution of an interface to improve the accuracy of the solution to an FBP.  To this end, we consider the semi-Lagrangian formulation \cite{Courant;Isaacson;Rees:52:On-the-Solution-of-N, Wiin-Nielsen;SemiLagrangian;1958, Fletcher;SemiLagrangianGeoScience;2020} as our starting point.  Semi-Lagrangian schemes are popular in the level-set community because of their unconditional stability and trivial nonuniform-grid implementation.  Here, we blend these ordinary schemes with an error-quantifying neural network that corrects the numerically estimated surface trajectory.  Unlike \cite{Zhuang;etal;LrndDiscForPassSclrAdvctn2D;2021, Pathak;etal;MLToAugCoarseGridCFD;2020, Liu;etal;DLMthdsSuperResoltnReconstTurbFlows;2020}, our model is a plain multilayer perceptron.  This network processes localized level-set, velocity, and positional data in a single time frame for select vertices near the moving front.  Our main contribution is thus a novel machine-learning-augmented transport algorithm that operates alongside selective redistancing and alternates with standard advection to keep the adjusted interface trajectory smooth.  Consequently, our procedure is more efficient than full-scan convolutional-based applications because it concentrates computational effort only around the free boundary.  Also, we show through various tests that our strategy is effective at counteracting both numerical diffusion and mass loss.  Some of these assessments reveal that our method can achieve the same precision as the baseline scheme at twice the resolution but at a fraction of the cost.  Similarly, we prove that our hybrid technique can produce feasible solidification fronts for crystallization processes.  In addition, our approach exhibits great generalization for simulation times that exceed the durations employed during training.  Together, these assets make our framework attractive to parallel level-set algorithms.  Its appeal resides in the possibility of avoiding further mesh refinement and decreasing expensive communications between computing nodes.

We have organized the paper as follows.  \Crefrange{sec:TheLevelSetMethod}{sec:SemiLagrangianAdvection} describe the level-set method, quadtree Cartesian grids, and semi-Lagrangian advection schemes.  Then, we state our methodology and related algorithms for training and deployment in \Cref{sec:Methodology}.  After these, \Cref{sec:Results} evaluates our approach through several standard test cases.  Finally, we discuss our results, limitations, and possible avenues for future work in \Cref{sec:Conclusions}.


\colorsection{The level-set method}
\label{sec:TheLevelSetMethod}

The level-set method \cite{Osher1988} is an Eulerian formulation for capturing and tracking interfaces that undergo complex topological changes.  The level-set representation denotes an interface by $\Gamma \doteq \{\vv{x} : \phi(\vv{x}) = 0\}$, where $\phi(\vv{x}) : \mathbb{R}^n \mapsto \mathbb{R}$ is a higher-dimensional, Lipschitz relation known as the \textit{level-set function}.  In this framework, the zero-isocontour partitions the computational domain $\Omega \subseteq \mathbb{R}^n$ into the non-overlapping interior and exterior regions defined by $\Omega^- \doteq \{\vv{x} : \phi(\vv{x}) < 0\}$ and $\Omega^+ \doteq \{\vv{x} : \phi(\vv{x}) > 0\}$.  Then, given some velocity field $\vv{u}(\vv{x})$, we can evolve $\phi(\vv{x})$ and $\Gamma$ by solving the \textit{level-set equation}:

\begin{equation}
\phi_t + \vv{u}\cdot\nabla\phi = 0.
\label{eq:LevelSetEquation}
\end{equation}

Assuming that $\phi(\vv{x})$ remains sufficiently smooth after advection, one can use the following straightforward expressions to compute normal vectors and mean curvature for any point $\vv{x} \in \Omega$:

\begin{equation}
\hat{\vv{n}}(\vv{x}) = \frac{\nabla\phi(\vv{x})}{||\nabla\phi(\vv{x})||}, \quad \kappa(\vv{x}) = \nabla \cdot \frac{\nabla\phi(\vv{x})}{||\nabla\phi(\vv{x})||}.
\label{eq:NormalAndCurvature}
\end{equation}

When $\vv{u}(\vv{x})$ does not depend directly on $\phi(\vv{x})$, \cref{eq:LevelSetEquation} is linear, and we can solve it with a semi-Lagrangian approach \cite{Courant;Isaacson;Rees:52:On-the-Solution-of-N, Wiin-Nielsen;SemiLagrangian;1958}.  Compared with other numerical schemes, the semi-Lagrangian method is unconditionally stable.  It is also trivial to implement on adaptive Cartesian grids, whereas higher-order advection solvers are challenging.  Combining the semi-Lagrangian formulation with nonuniform meshes is thus convenient because it allows us to solve \cref{eq:LevelSetEquation} in level-set applications with very high grid resolutions.  

In general, there is an infinite number of functions for which $\Gamma$ describes the same zero level set.  Usually, one chooses $\phi(\vv{x})$ as a signed distance function because it simplifies computations (e.g., $||\nabla \phi(\vv{x})|| = 1$) and leads to robust numerical results \cite{Sussman;Smereka;Osher:94:A-Level-Set-Approach}.  Also, signed distance functions are beneficial because they are uniquely determined as the viscosity solutions to the Eikonal equation \cite{Min:10:On-reinitializing-le}.  Recent studies \cite{LALariosFGibou;LSCurvatureML;2021, Larios;Gibou;HybridCurvature;2021} suggest that these functions can improve the accuracy of machine learning estimations, too.  However, as a numerical simulation progresses, a signed distance function can deteriorate quickly, developing noisy features that compromise stability and precision.  For this reason, it is customary to redistance $\phi(\vv{x})$ frequently by solving the pseudo-time transient \textit{reinitialization equation} \cite{Sussman;Smereka;Osher:94:A-Level-Set-Approach}

\begin{equation}
\phi_\tau + \texttt{sgn}(\phi^0)(||\nabla\phi|| - 1) = 0,
\label{eq:Reinitialization}
\end{equation}
where $\tau$ is a pseudo-time stepping variable, $\phi^0$ is the starting level-set function, and $\texttt{sgn}(\cdot)$ is a smoothed-out signum function.

Typically, one uses a TVD Runge--Kutta scheme in time and a Godunov discretization in space to solve \cref{eq:Reinitialization} to a steady state (i.e., $\phi_\tau = 0$ and $||\nabla\phi|| = 1$).  For efficiency, we often stop this iterative process after a prescribed number of steps $\nu$, depending on the application and how close $\phi^0$ is to a signed distance function.  For a detailed description of the reinitialization algorithm, we refer the reader to the work of Min and Gibou \cite{Min;Gibou:07:A-second-order-accur} and Mirzadeh \etal \cite{Mirzadeh;etal:16:Parallel-level-set}.  In both cases, the authors redesigned the redistancing procedure for adaptive Cartesian grids, which we discuss in \Cref{sec:AdaptiveCartesianGrids}.  For a complete presentation of the level-set technologies, the reader may consult Osher, Fedkiw, and Sethian's classic texts \cite{Osher;Fedkiw:02:Level-Set-Methods-an, Sethian:99:Level-set-methods-an} and the latest review by Gibou \etal \cite{GFO18}.


\colorsection{Adaptive Cartesian grids}
\label{sec:AdaptiveCartesianGrids}

Unlike uniform meshes, adaptive Cartesian grids significantly reduce the cost of solving FBPs with level-set methods.  They do so by increasing the spatial resolution only close to $\Gamma$, where accuracy is needed the most \cite{Strain1999}.  In this work, we discretize a computational domain $\Omega$ with the help of standard quadtrees and signed distance level-set functions to digitize the corresponding grid $\mathcal{G}$.  

\begin{figure}[t]
	\centering
	\includegraphics[width=0.9\textwidth]{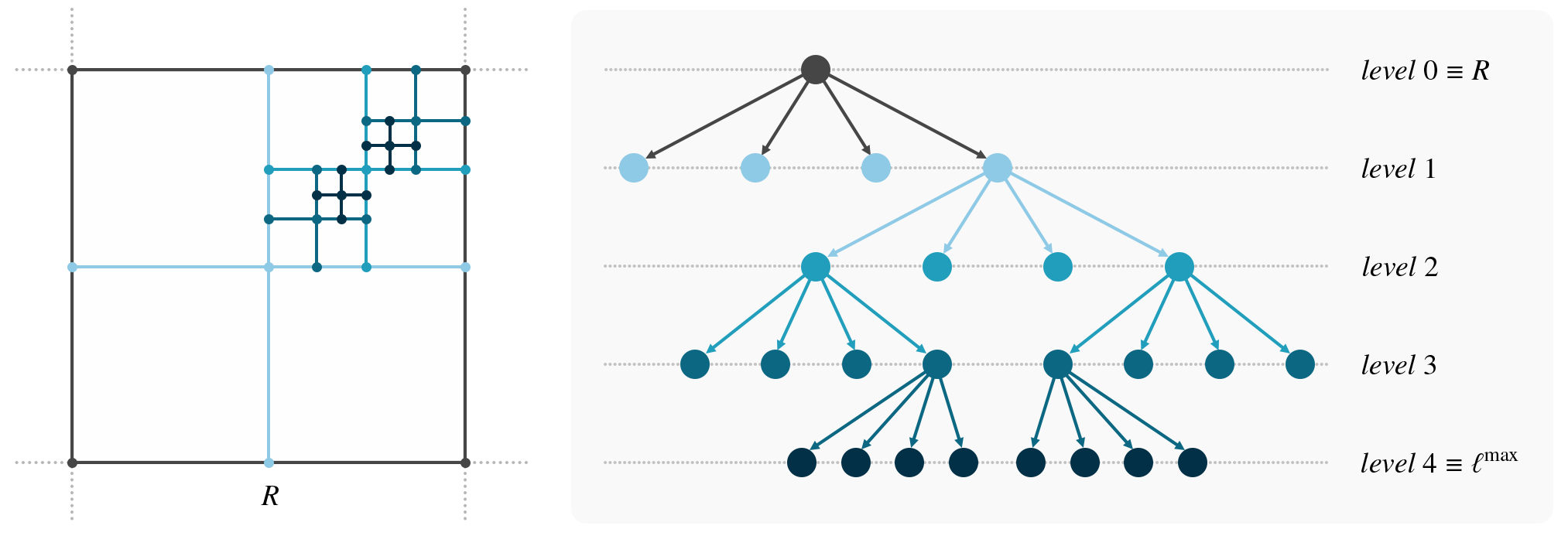}
	\caption{A (non-graded) quadtree Cartesian grid (left) and its data structure representation (right).  The quadtree covers a region $R \subseteq \Omega$ corresponding to its root ($level~0$).  We denote grid nodes as small dots at the intersections of the quadtree cells.  (Color online.)}
	\label{fig:Quadtree}
\end{figure}

A quadtree is a rooted data structure composed of discrete cells covering a rectangular region $R \subseteq \Omega$.  A tree cell $C$ has four children (i.e., quadrants) or is a leaf \cite{BKOS00}.  Also, each cell contains a list with its vertex coordinates, nodal level-set values, and other application-dependent data.  \Cref{fig:Quadtree} illustrates a quadtree with its cells organized into $0 \leqslant L = \ell^{\max} + 1$ levels.  Given an $L$-level quadtree, we can perform several operations efficiently, like searching in $\mathcal{O}(\ln(\ell^{\max}))$ time and sorting \cite{Strain1999}.  In addition, we can determine the mesh size or minimum cell width $h$ by establishing a relationship between $\Omega$'s side lengths and the maximum level of refinement $\ell^{\max}$.

As with most level-set applications based on quadtree Cartesian grids, we have adopted Min's extended Whitney decomposition \cite{Min2004} to refine the mesh near the zero-isocontour.  To construct or update a quadtree, we begin at its root $R$ and recursively subdivide its cells according to their distance to $\Gamma$ until we reach level $\ell^{\max}$.  In particular, we mark cell $C$ for refinement if the condition

\begin{equation}
\min_{v \in \textrm{vertices}(C)}{|\phi(v)|} \leqslant \textrm{Lip}(\phi)\cdot\texttt{diag}(C)
\label{eq:RefinementCriterion}
\end{equation}
is valid, where $v$ is a vertex, $\textrm{Lip}(\phi)$ is the level-set function's Lipschitz constant (set to $1.2$), and $\texttt{diag}(C)$ is $C$'s diagonal length.  Conversely, we mark any cells for coarsening whenever they cannot fulfill the above criterion \cite{Mirzadeh;etal:16:Parallel-level-set}.

Our research extends the parallel level-set methods provided in \cite{Mirzadeh;etal:16:Parallel-level-set}, which rely on Burstedde and coauthors' {\tt p4est} library \cite{Burstedde;Wilcox;Ghattas:11:p4est:-Scalable-Algo}.  {\tt p4est} is a suite of highly scalable routines for refining, coarsening, partitioning, and load-balancing grids.  This library represents the adaptive grid $\mathcal{G}$ as a \textit{forest} of abutting quadtrees rooted in individual cells of a coarse parent grid known as the \textit{macromesh}.  The most important contribution of Mirzadeh \etal in \cite{Mirzadeh;etal:16:Parallel-level-set} is a parallel interpolation procedure.  This algorithm exploits {\tt p4est}'s ghost layering and global nodal indexing by recreating the entire forest as a per-process hierarchical representation.  As shown in \Cref{sec:Methodology}, we have employed their local replication and multi-process interpolation mechanisms to realize semi-Lagrangian advection.  For more details about {\tt p4est} and the level-set methods for distributed-memory systems using efficient {\tt MPI} communications \cite{MPI14}, we refer the reader to their respective manuscripts and handouts.


\colorsection{Semi-Lagrangian advection}
\label{sec:SemiLagrangianAdvection}

Semi-Lagrangian schemes \cite{Wiin-Nielsen;SemiLagrangian;1958} are extensions of the Courant--Isaacson--Rees method \cite{Courant;Isaacson;Rees:52:On-the-Solution-of-N}.  Practitioners have verified their convergence for several FBPs with passive transport where \cref{eq:LevelSetEquation} is hyperbolic.  For parabolic PDEs---where the Courant--Friedrichs--Lewy (CFL) condition requires $\Delta t \sim \mathcal{O}(h^2)$ for most explicit methods---semi-Lagrangian schemes can also converge with $\Delta t \sim \mathcal{O}(h)$ if one integrates frequent reinitialization and velocity smoothing \cite{Strain1999}.

Because of their versatility, unconditional stability, and ease of implementation, practitioners have resorted to semi-Lagrangian methods for solving FBPs in both uniform and adaptive grids.  Semi-Lagrangian schemes, however, are not conservative and often lead to mass loss when applied to the evolution of the level-set function \cite{GFO18}.  

The basis of this numerical technique is that we begin with an Eulerian mesh with known grid points and level-set values at the current time step.  Also, we know the vertex locations in the next time frame, but we must solve for the unknown new level-set values at these points \cite{Fletcher;SemiLagrangianGeoScience;2020}.  To approximate the solution to \cref{eq:LevelSetEquation} at these locations, we can integrate the system

\begin{equation}
\left\{
\begin{split}
	\ode{X(t)}{t} & = \vv{u}(X(t)) \\
	\ode{\phi(X(t), t)}{t} & = 0
\end{split}\right.
\label{eq:LevelSetEquationAlongCharacteristics}
\end{equation}
along the characteristic curves $X(t)$ backward in time.  Thus, if $\mathcal{G}^{n+1}$ is the computational grid at time $t^{n+1} = t^n + \Delta t$,\footnote{We use the traditional shorthand temporal discretization $\psi^n \equiv \psi(t^n)$ for variable $\psi$.} we start at a known \textit{arrival} grid point $\vv{x}_a^{n+1}$ lying on the trajectory $X^{n+1}$ and trace it back to its \textit{departure} point $\vv{x}_d$ in the upwind direction.  Then, we compute its level-set value as in \cite{Mirzadeh;etal:16:Parallel-level-set} with

\begin{equation}
\phi^{n+1}(\vv{x}_a^{n+1}) = \phi^{n+1}(X^{n+1}) = \phi(X(t^{n+1}), t^{n+1}) = \phi(X(t^n), t^n) = \phi^n(X^n) = \phi^n(\vv{x}_d),
\label{eq:SLLevelSetValue}
\end{equation}
where the second-order accurate midpoint method provided in \cite{Xiu;Karniadakis:01:A-Semi-Lagrangian-Hi, Min;Gibou:07:A-second-order-accur},

\begin{subequations}
\begin{align}
\hat{\vv{x}} &= \vv{x}_a^{n+1} - \frac{\Delta t}{2}\vv{u}^n(\vv{x}_a^{n+1}) \label{eq:SLMidpoint} \\
\vv{x}_d     &= \vv{x}_a^{n+1} - \Delta t \vv{u}^{n+\frac{1}{2}}(\hat{\vv{x}}), \label{eq:SLDeparturePoint}
\end{align}
\label{eq:SLMidpointDeparturePoint}
\end{subequations}
locates the departure point using the intermediate velocity

\begin{equation}
\vv{u}^{n+\frac{1}{2}} = \frac{3}{2}\vv{u}^n - \frac{1}{2}\vv{u}^{n-1}.
\label{eq:SLIntermediateVelocity}
\end{equation}

\begin{figure}[t]
	\centering
	\includegraphics[width=7cm]{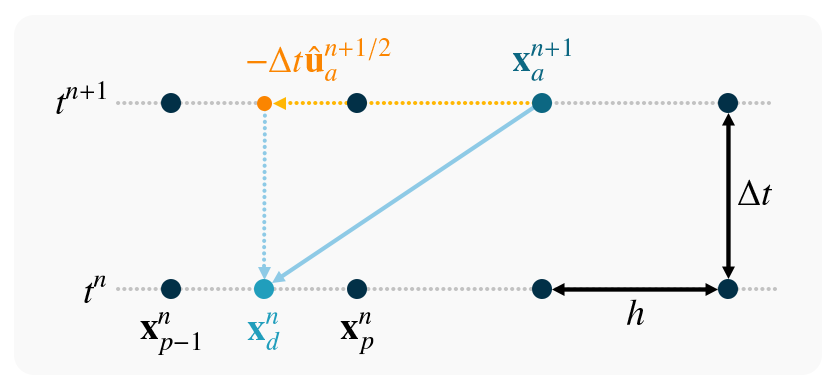}
	\caption{Illustration of backtracking with a one-dimensional semi-Lagrangian scheme.  $\vv{x}_a^{n+1}$ is the arrival grid point, $\vv{x}_d^n$ is its departure point lying between $\vv{x}_p^n$ and $\vv{x}_{p-1}^n$, and $\hat{\vv{u}}_a^{n+1/2}$ is the intermediate velocity from the midpoint method.  The space-time characteristic appears as a light blue solid arrow and maps $\vv{x}_a^{n+1}$ to $\vv{x}_d^n$.  Adapted from \cite{Fletcher;SemiLagrangianGeoScience;2020}.  (Color online.)}
	\label{fig:SemiLagrangian}
\end{figure}

Since $\hat{\vv{x}}$ and $\vv{x}_d$ do not necessarily coincide with grid points, we must interpolate the appropriate fields from $\mathcal{G}^n$ and $\mathcal{G}^{n-1}$ to calculate $\vv{u}^{n+\frac{1}{2}}(\hat{\vv{x}})$ and $\phi^n(\vv{x}_d)$.  \Cref{fig:SemiLagrangian} shows this situation for a one-dimensional semi-Lagrangian scheme.  In this work, we employ bilinear and quadratic interpolation to sample scalar and vector fields on quadtree Cartesian grids.  As detailed by Strain \cite{Strain1999}, if $C \in \mathcal{G}$ is an $h$-by-$h$ quadtree cell with lower-left coordinates $(x_0, y_0)$, we can calculate $\phi(\vv{x})$ at $(x,y) = (x_0 + \alpha h, y_0 + \beta h) \in C$ using the piecewise bilinear interpolation form

\begin{equation}
\phi(x,y) = (1-\alpha)(1-\beta)\phi_{00} + (1-\alpha)(\beta)\phi_{01} + (\alpha)(1-\beta)\phi_{10} + (\alpha)(\beta)\phi_{11},
\label{eq:BilinearInterpolation}
\end{equation}
where $\phi_{ij} = \phi(x_0 + ih, y_0 + jh)$ is a vertex value in $C$.  Similarly, as noted by Min and Gibou \cite{Min;Gibou:07:A-second-order-accur}, we can get a quadratic interpolation procedure by correcting \cref{eq:BilinearInterpolation} with second-order derivatives.  Hence,

\begin{equation}
\begin{split}
\phi(x,y) =&~ (1-\alpha)(1-\beta)\phi_{00} + (1-\alpha)(\beta)\phi_{01} + (\alpha)(1-\beta)\phi_{10} + (\alpha)(\beta)\phi_{11} \\
           &~ - h^2\frac{(\alpha)(1-\alpha)}{2}\phi_{xx} - h^2\frac{(\beta)(1-\beta)}{2}\phi_{yy},
\end{split}
\label{eq:QuadraticInterpolation}
\end{equation}
where $\phi_{xx}$ and $\phi_{yy}$ are bilinearly interpolated at $(x,y)$ from their corresponding vertex values for $C$.  The reader may consult \cite{Min;Gibou:07:A-second-order-accur} for more information about the standard finite-difference formulations to discretize $\phi_{xx}$ and $\phi_{yy}$ on non-graded, adaptive meshes.  The textbook by \cite{Fletcher;SemiLagrangianGeoScience;2020} also provides a detailed overview of semi-Lagrangian schemes and their applications.  In the present study, we have instrumented \cref{eq:BilinearInterpolation,eq:QuadraticInterpolation} within Algorithm \href{https://www.sciencedirect.com/science/article/pii/S002199911630242X\#fg0050}{2} in \cite{Mirzadeh;etal:16:Parallel-level-set} for heterogeneous computing systems.


\colorsection{Methodology}
\label{sec:Methodology}

In this section, we state our data-driven strategy to improve the accuracy of semi-Lagrangian schemes in quadtree Cartesian grids.  Our approach blends machine learning with the level-set technologies of \cite{Min;Gibou:07:A-second-order-accur} and \cite{Mirzadeh;etal:16:Parallel-level-set} to solve a specialized instance of the image super-resolution problem \cite{Dong;Loy;He;SuperResolution;2014}.  The idea behind our framework stems from the fact that highly resolved grids yield much better solutions to FBPs than their coarser counterparts.  Here, as in \cite{Pathak;etal;MLToAugCoarseGridCFD;2020}, we investigate the possibility of computing on-the-fly corrections for a coarse-grid interface trajectory so that it closely follows its corresponding evolution in a finer mesh.

\colorsubsection{Problem definition}
\label{subsec:ProblemDefinition}

Let $\mathcal{S}^n$ be the state of the solution to an FBP at time $t^n$.  $\mathcal{S}$ embodies a computational grid besides its nodal level-set, velocity, pressure, and temperature values, among others.  Further, define {\tt FBPEqnSolver()} as a sequence of nonlinear operators acting on $\mathcal{S}^n$.  {\tt FBPEqnSolver()} prepares the advection of $\phi^n \in \mathcal{S}^n$ by producing the intermediate state $\hat{\mathcal{S}}^n = \texttt{FBPEqnSolver(}\mathcal{S}^n\texttt{)}$.  Then, given a time step $\Delta t$ and a {\tt SemiLagrangian()} procedure that performs the operations described in \Cref{sec:SemiLagrangianAdvection}, the statement

\begin{equation}
\mathcal{S}^{n+1} = \texttt{SemiLagrangian(FBPEqnSolver(}\mathcal{S}^n\texttt{)}, \Delta t\texttt{)}
\label{eq:SemiLagrangianStep}
\end{equation}
finalizes the solution at time $t^{n+1}$.  In particular, the \texttt{SemiLagrangian()} method arrives at $\mathcal{S}^{n+1}$ by advancing the sampled level-set function from $\phi^n$ to $\phi^{n+1}$. Also, it updates the underlying quadtree Cartesian mesh from $\mathcal{G}^n$ to $\mathcal{G}^{n+1}$.  For a technical description of the {\tt SemiLagrangian()} subroutine, the reader may consult Algorithm \href{https://www.sciencedirect.com/science/article/pii/S002199911630242X\#fg0060}{3} in \cite{Mirzadeh;etal:16:Parallel-level-set}.

Next, consider a simulation that starts by discretizing $\Omega$ with a coarse mesh $\mathcal{G}_c^0$ and a fine mesh $\mathcal{G}_f^0$, where $\ell_c^{\max} < \ell_f^{\max}$, and $h_c > h_f$ (see \cref{fig:Quadtree})\footnote{Unless otherwise stated, the subscript $c$ refers to a coarse-grid feature/component and $f$ to the fine-grid counterpart.}.  In addition, assume that the nodal coordinates in $\mathcal{G}_c^n$ are a proper subset of the vertex coordinates in $\mathcal{G}_f^n$ for all $t^n$.  If $\mathcal{S}_c(t^0) = \mathcal{S}_f(t^0)$ is the FBP's initial condition\footnote{In practice, it suffices that $\mathcal{S}_c(t^0)$ and $\mathcal{S}_f(t^0)$ are approximately equal along a narrow region of interest (e.g., in a shell around $\Gamma_c^0$).} involving the simultaneous coarse- and fine-grid solutions, then

\begin{equation}
\mathcal{S}_c(t^0+\delta) = \texttt{SemiLagrangian(}\hat{\mathcal{S}}_c^0, \delta\texttt{)} \neq \texttt{SemiLagrangian(}\hat{\mathcal{S}}_f^0, \delta\texttt{)} = \mathcal{S}_f(t^0+\delta)
\label{eq:SemiLagrangianComparison}
\end{equation}
after some finite time interval $\delta > 0$.  More precisely, $\phi_c(t^0+\delta) \neq \phi_f(t^0+\delta)$, and $\Gamma_c(t^0+\delta) \neq \Gamma_f(t^0+\delta)$, mainly because of the numerical diffusion in the {\tt SemiLagrangian()} function.  Consequently, both the coarse- and the fine-grid solutions will diverge as the simulation progresses.

In this study, if $\bar{\delta} \sim \mathcal{O}(h_c)$ is a small time interval and the velocity field satisfies $\max{||\vv{u}_c(\vv{x})||} = 1$, $\forall \vv{x} \in \Omega$, we claim that one can characterize $\mathcal{S_c}(t^0+\bar{\delta})$'s deviation from $\mathcal{S_f}(t^0+\bar{\delta})$ with

\begin{equation}
\phi_f(t^0+\bar{\delta}) = \phi_c(t^0+\bar{\delta}) + \bar{\varepsilon},
\label{eq:LevelSetDeviation}
\end{equation}
where $\bar{\varepsilon}$ is the coarse level-set error.

Therefore, if we can model $\bar{\varepsilon}$ in \cref{eq:LevelSetDeviation} at regular intervals $\bar{\delta}$, we will be able to correct $\mathcal{S}_c$, improve its precision, and reduce artificial mass loss.  As introduced in \Cref{sec:AdaptiveCartesianGrids}, we can further simplify this error model by concentrating our effort only at vertices next to $\Gamma$, where accuracy is critical.


\colorsubsection{Error-correcting neural networks for semi-Lagrangian advection}
\label{subsec:ErrorCorrectingNNetsForSLAdvect}

Our goal is to design a function that quantifies the \textit{local error} $\bar{\varepsilon}$ incurred by $\phi_c^n(\vv{x})$'s semi-Lagrangian transport for any vertex next to the interface.  In this model, we should leverage the statistical information available from various fields in $\hat{\mathcal{S}}_c^n$ in the neighborhood of $\Gamma_c^n$.  Upon evaluating these fields, our function should estimate (with high precision) how much numerical dissipation in $\phi_c^{n+1}(\vv{x})$ has taken place as we advanced the simulation from $t^n$ to $t^{n+1}$.

Let $\mathcal{F}_{c,f}(\cdot)$ be our \textit{error-correcting} function for advected level-set values near $\Gamma_c^n$.  Here, we follow state-of-the-art machine learning approaches in CFD \cite{Xie;etal;TempoGAN;2018, Liu;etal;DLMthdsSuperResoltnReconstTurbFlows;2020, Pathak;etal;MLToAugCoarseGridCFD;2020, Zhuang;etal;LrndDiscForPassSclrAdvctn2D;2021} and materialize $\mathcal{F}_{c,f}(\cdot)$ as a neural network.  More specifically, we resort to image super-resolution methodologies to estimate $\bar{\varepsilon}$ and construct $\mathcal{F}_{c,f}(\cdot)$ by abstracting the difference between the coarse-mesh semi-Lagrangian advection of $\phi_c^n(\vv{x})$ and the numerical reference trajectory in a fine grid.

\begin{figure}[t]
	\centering
	\includegraphics[width=\textwidth]{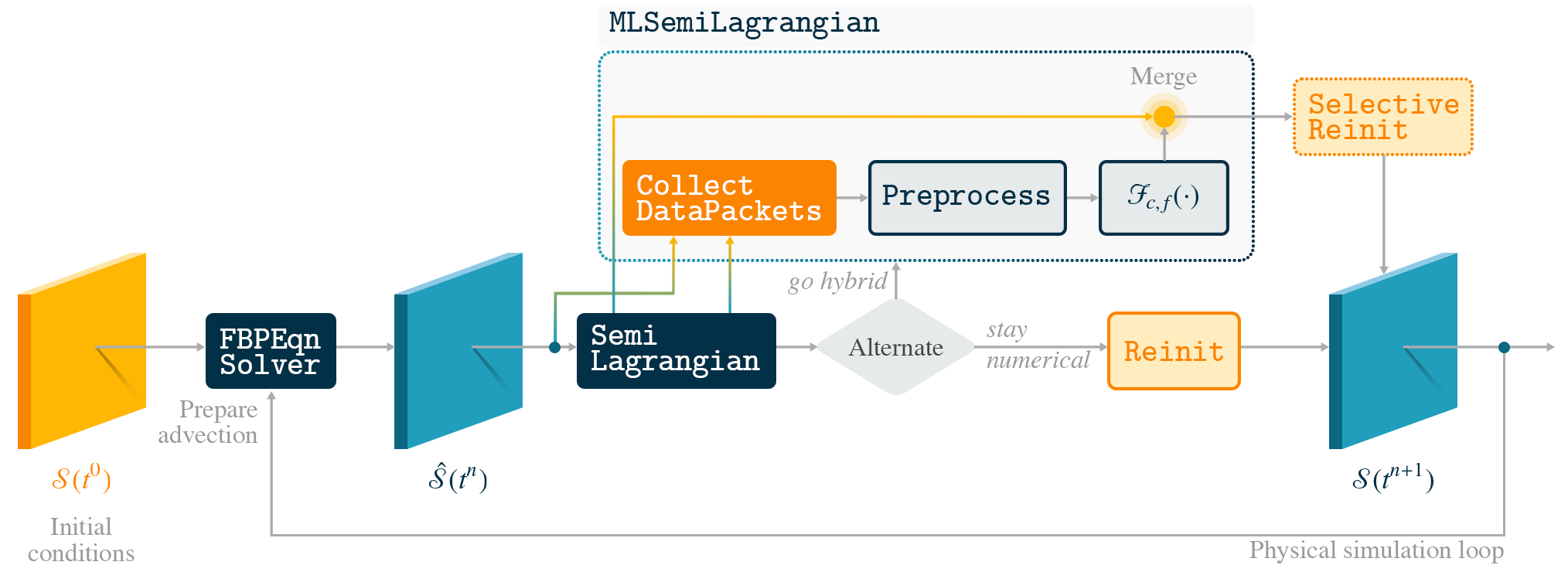}
	\caption{Overview of our hybrid semi-Lagrangian solver. The {\tt MLSemiLagrangian()} routine couples numerical advection with the error-correcting neural network $\mathcal{F}_{c,f}(\cdot)$ to improve the level-set accuracy of grid points next to $\Gamma^n$.  A full description of {\tt MLSemiLagrangian()} is provided in \Cref{alg:MLSemiLagrangian}.  Our solver alternates between standard transport and machine-learning-induced advection.  When using the neural model, we apply selective reinitialization to protect the level-set values in a subset of the corrected trajectory.  The alternating mechanism helps smooth out the level-set values near the evolving front since the {\tt MLSemiLagrangian()} module cannot guarantee regularity on its own.  (Color online.)}
	\label{fig:Overview}
\end{figure}

\Cref{fig:Overview} describes how $\mathcal{F}_{c,f}(\cdot)$ interacts with a conventional FBP solver.  The {\tt MLSemiLagrangian()} block is the main contribution of this research.  First, this module gathers \textit{data packets} from intermediate state information and {\tt SemiLagrangian()}-advected level-set values for grid nodes near the interface at time $t^n$.  Next, a custom subroutine preprocesses these packets and routes them to $\mathcal{F}_{c,f}(\cdot)$ for evaluation.  Finally, $\mathcal{F}_{c,f}(\cdot)$ yields improved local estimations to $\phi_c^{n+1}(\vv{x})$ that are subsequently re-integrated to the {\tt SemiLagrangian()} computations.  The advection algorithm within the {\tt MLSemiLagrangian()} method is iterative and involves successive mesh coarsening and refining steps until $\mathcal{G}_c^{n+1}$ converges (see Section \href{https://www.sciencedirect.com/science/article/pii/S002199911630242X\#se0050}{3.2} in \cite{Mirzadeh;etal:16:Parallel-level-set}).  For efficiency, we cache the machine-learning-corrected level-set values from the first iteration.  Then, we \textit{merge} them with the proposed numerical estimations before each re-gridding operation.  In addition to the {\tt MLSemiLagrangian()} unit, our framework includes a selective reinitialization function that masks out and protects a portion of the enhanced trajectory from the perturbations arising when solving \cref{eq:Reinitialization}.  As in \cite{Mirzadeh;etal:16:Parallel-level-set}, we integrate \cref{eq:Reinitialization} using a second-order accurate TVD-RK scheme \cite{Shu;Osher:88:Efficient-Implementa} with explicit, adaptive time-stepping and a Godunov discretization with quadratic interface localization in space \cite{Min;Gibou:07:A-second-order-accur}.  \Cref{fig:Overview} also depicts an \textit{alternating mechanism} between standard advection and machine-learning-induced transport.  Such a mechanism is fundamental to (1) counteract the lack of physical constraints in $\mathcal{F}_{c,f}(\cdot)$ and (2) regularize the adjusted level set values next to $\Gamma_c^n$.  The following sections detail how all these elements interplay within our hybrid advection system.

\begin{figure}[t]
	\centering
	\includegraphics[width=\textwidth]{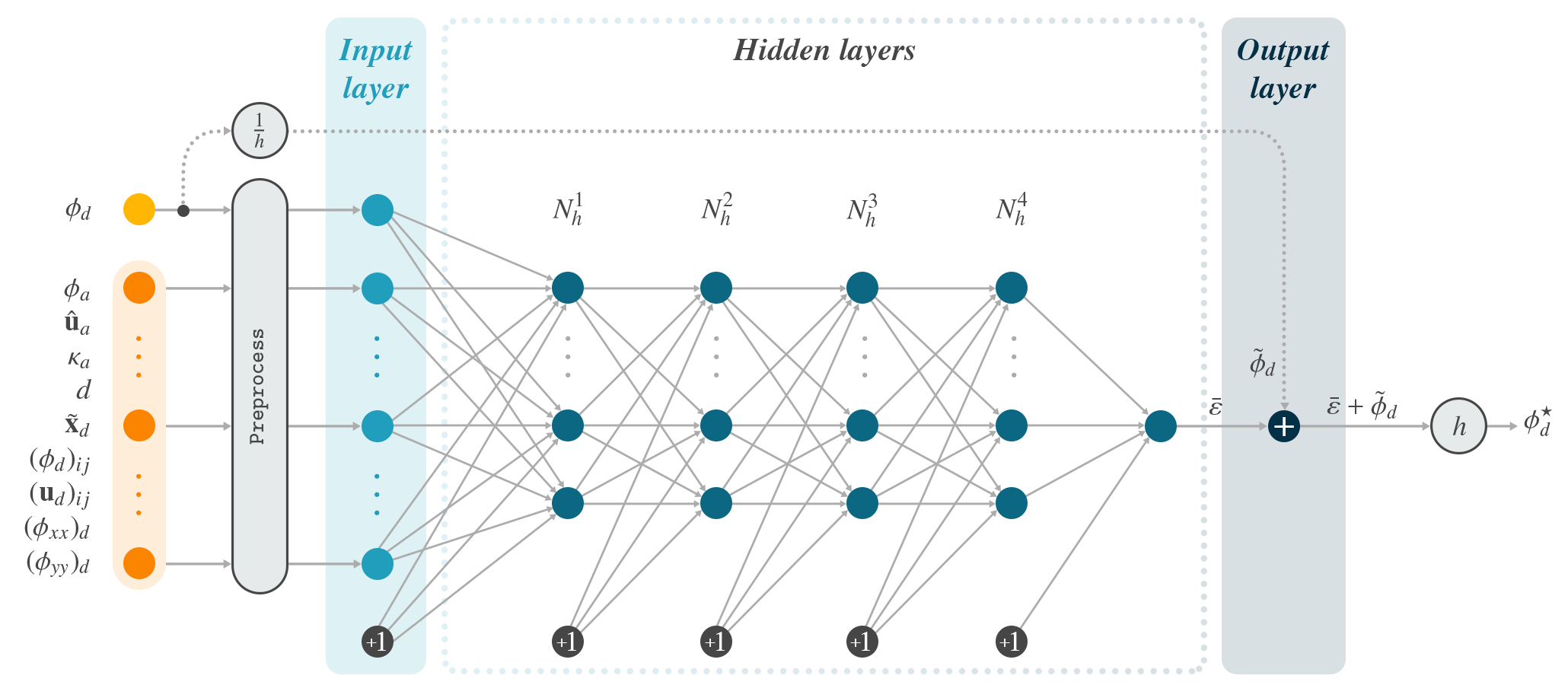}
	\caption{The error-correcting neural network $\mathcal{F}_{c,f}(\cdot)$ employed in the {\tt MLSemiLagrangian()} module of \cref{fig:Overview}.  We also show the {\tt Preprocess()} subroutine and the normalization/denormalization operations for the level-set value $\phi_d$ estimated with the semi-Lagrangian scheme of \Cref{sec:SemiLagrangianAdvection}.  In general, we expect the corrected output $\phi_d^\star$ to be closer to the level-set value computed with a fine grid.  (Color online.)}
	\label{fig:ECNet}
\end{figure}

Our neural model resembles a localized version of the solver in \cite{Pathak;etal;MLToAugCoarseGridCFD;2020} and adapts the hybrid strategy in \cite{Zhuang;etal;LrndDiscForPassSclrAdvctn2D;2021} to the level-set framework.  \Cref{fig:ECNet} outlines $\mathcal{F}_{c,f}(\cdot)$'s architecture.  Unlike \cite{Pathak;etal;MLToAugCoarseGridCFD;2020, Zhuang;etal;LrndDiscForPassSclrAdvctn2D;2021}, $\mathcal{F}_{c,f}(\cdot)$ is an ordinary multilayer perceptron that outputs a corrected level-set value after digesting a small input vector.  Besides a typical feedforward architecture for estimating $\bar{\varepsilon}$, $\mathcal{F}_{c,f}(\cdot)$ features a skip connection that carries the $h$-normalized\footnote{We indicate $h$-normalized variables with a tilde $\tilde{~}$ accent.} level-set value $\tilde{\phi}_d$ to a non-trainable additive neuron that computes $\bar{\varepsilon} + \tilde{\phi}_d$.  In the end, upon applying $h$-denormalization (subsumed in $\mathcal{F}_{c,f}(\cdot)$'s box in \cref{fig:Overview}), we produce a better level-set approximation, $\phi_d^\star$, for advancing the moving front.  Compared to \cite{Liu;etal;DLMthdsSuperResoltnReconstTurbFlows;2020}, $\mathcal{F}_{c,f}(\cdot)$ exploits only low-resolution data from one step, thus relieving the hybrid solver from buffering temporal information beyond time $t^n$.  \Cref{fig:ECNet} also displays the preprocessing module, which transforms incoming data according to patterns observed during training.  Similar to our findings in \cite{Larios;Gibou;HybridCurvature;2021}, the {\tt Preprocess()} subroutine is crucial for the {\tt MLSemiLagrangian()} procedure because it favors learning convergence and increases $\mathcal{F}_{c,f}(\cdot)$'s accuracy.  We discuss the {\tt Preprocess()} building block alongside its technical implementation in \Cref{subsubsec:TechnicalAspects}.

Our goal is to ensure that the free boundary motion from $\Gamma^n$ to $\Gamma^{n+1}$ is highly accurate.  For this reason, we reserve $\mathcal{F}_{c,f}(\cdot)$ for data packets collected for points $\vv{x}_a^{n+1}$ lying next to the \textit{known} interface at time $t^n$.  As shown in \cref{fig:ECNet}, the corresponding input for a qualified vertex $\vv{x}_a^{n+1}$ to $\mathcal{F}_{c,f}(\cdot)$ is a preprocessed version of data available to the {\tt SemiLagrangian()} subroutine.  The latter is the basis to compute $\phi^{n+1}(\vv{x}_a^{n+1}) = \phi^n(\vv{x}_d)$.  Among these data, we find level-set values, distances, velocity components, and second-order derivatives.  All of them take the role of arguments in \crefrange{eq:SLLevelSetValue}{eq:QuadraticInterpolation}.  We will describe the data-extraction procedure later in \Cref{subsec:DataPacketCollection}.

In order to specify a data packet formally, suppose we trace a vertex located at $\vv{x}_a$ near $\Gamma^n$ back to its departure point\footnote{For data-packet specific attributes, we use the subscript $a$ to denote information at the arrival point and $d$ for the departure point.} $\vv{x}_d$.  To do so, we move in the $-\Delta t\,\hat{\vv{u}}_a$ direction according to \cref{eq:SLMidpointDeparturePoint}, where $\hat{\vv{u}}_a \equiv \vv{u}(\hat{\vv{x}}) \equiv \vv{u}^n(\hat{\vv{x}})$.  This last expression holds because we assume that $\vv{u}^{n+\frac{1}{2}} \equiv \vv{u}^n$ in \cref{eq:SLIntermediateVelocity}.  In particular, we can find $\hat{\vv{x}}$ using only the velocity field at $t^n$ and \cref{eq:SLMidpoint} followed by a quadratic interpolation step to evaluate $\vv{u}^n(\hat{\vv{x}})$.  Then, we can readily compute $\vv{x}_d$ with \cref{eq:SLDeparturePoint}.  Since $\mathcal{F}_{c,f}(\cdot)$ does not consume entries from previous solution states, from now on, we will drop the $n$ superscript to simplify the notation.  

Next, let $C$ be the $h$-by-$h$ square cell at the maximum level of refinement that \textit{owns} $\vv{x}_d$.  From $C$, we can build a data packet

\begin{equation}
\mathcal{p} = \left(\begin{array}{rlcrl}
	                    \phi_a: & \textrm{level-set value at }\vv{x}_a                 &~& (\vv{u}_d)_{ij}: & \textrm{nodal velocities} \\
	            \hat{\vv{u}}_a: & \textrm{midpoint velocity at }\hat{\vv{x}}           &~&   (\phi_{xx})_d: & |\partial^2\phi/\partial x^2| \textrm{ bilinearly interpolated at }\vv{x}_d \\
	                         d: & \textrm{distance from }\vv{x}_d\textrm{ to }\vv{x}_a &~&   (\phi_{yy})_d: & |\partial^2\phi/\partial y^2| \textrm{ bilinearly interpolated at }\vv{x}_d \\
	(\tilde{x}_d, \tilde{y}_d): & \vv{x}_d\textrm{ coordinates w.r.t. }(x_0, y_0)      &~&        \kappa_a: & \textrm{curvature bilinearly interpolated at }\Gamma \\
	             (\phi_d)_{ij}: & \textrm{nodal level-set values}                      &~&          \phi_d: & \textrm{numerical level-set value at }\vv{x}_d \\
\end{array}\right) \in \mathbb{R}^{22},
\label{eq:DataPacket}
\end{equation}
where $(\psi_d)_{ij} = (\psi_d)(x_0 + ih, y_0 + jh)$ is a vertex value, $i,j \in \{0, 1\}$, and $(x_0, y_0)$ is $C$'s lower-left corner.  

\begin{figure}[t]
	\centering
	\begin{subfigure}[b]{5cm}
		\includegraphics[width=\textwidth]{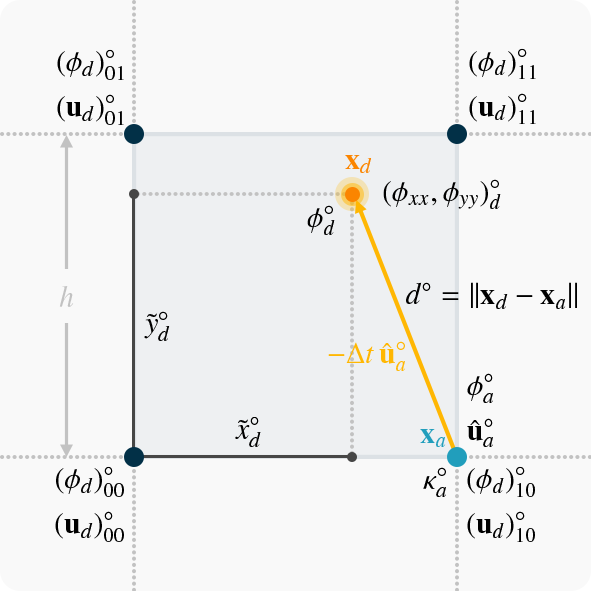}
		\caption{\footnotesize Original data packet $\mathcal{p}^\circ$}
		\label{fig:OriginalSample}
	\end{subfigure}
	~
	\begin{subfigure}[b]{5cm}
		\includegraphics[width=\textwidth]{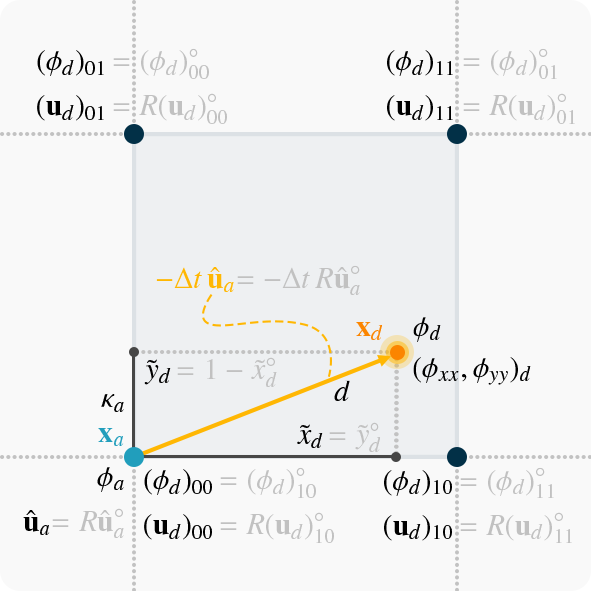}
		\caption{\footnotesize Reoriented data packet $\mathcal{p}$ (standard form)}
		\label{fig:ReorientedSample}
	\end{subfigure}
	~
	\begin{subfigure}[b]{5cm}
		\includegraphics[width=\textwidth]{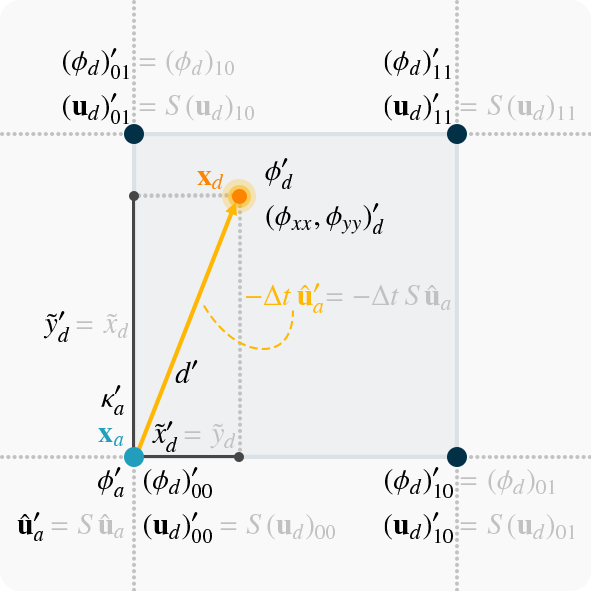}
		\caption{\footnotesize Augmented data packet $\mathcal{p}'$}
		\label{fig:AugmentedSample}
	\end{subfigure}
	\caption{An $h$-by-$h$ grid cell $C$ at the maximum level of refinement owning the departure point $\vv{x}_d$ (in orange) for vertex $\vv{x}_a$ (in light blue) next to $\Gamma$.  Original sampled data from a numerical simulation appears in (a), denoted with a $\circ$ superscript.  Our inference system digests only numerical values shown in black.  After reorientation, we obtain the machine-learning-suitable data packet in (b), where $R$ is a rotation transformation (assuming that $\kappa_a^\circ < 0$).  Invariant data to rotation transfers directly from (a) to (b), and we do not indicate any equality explicitly (e.g., $\phi_d = \phi_d^\circ$).  (c) illustrates data augmentation by reflection about the line $y = x + \beta$, with $\beta \in \mathbb{R}$, going through $\vv{x}_a$.  A permutation matrix $S$ swaps the elements of multicomponent reoriented data in (b) (e.g., velocities).  Invariant data to reflection transfers directly from (b) to (c) (e.g., $(\phi_d)_{00}' = (\phi_d)_{00}$).  We denote augmented data with a prime superscript.  (Color online.)}
	\label{fig:Sample}
\end{figure}

\Cref{fig:OriginalSample} describes a data packet in a spatial context, where a $\circ$ superscript differentiates the original from the \textit{standard form} suitable for machine learning.  \Cref{fig:OriginalSample} also shows that $\vv{x}_a$ always coincides with one of $C$'s vertices if we consider a unit $CFL$ constant and a velocity maximum-unit-norm restriction.  The previous statement about $\vv{x}_a$ is true regardless of grid resolution and is essential for narrowing the problem at hand.

In earlier research \cite{Larios;Gibou;HybridCurvature;2021}, we exploited symmetry in the curvature problem to facilitate the neural-network design and data-set composition.  In particular, we confirmed that training only on the \textit{negative curvature spectrum} favored efficiency, reduced architectural complexity, and accelerated learning.  These benefits hinged on available, inexpensive mechanisms to gauge a stencil's convexity and flip the sign of level-set values and inferred curvature accordingly.  Here, we rely on this strategy to simplify $\mathcal{F}_{c,f}(\cdot)$'s topology while preserving at least the same expressive power as a full-curvature-spectrum model.  Thus, the idea is to calculate $\kappa_a^\circ$ in \cref{fig:OriginalSample} by discretizing \cref{eq:NormalAndCurvature} using finite differences.  Then, we estimate curvature at

\begin{equation}
\vv{x}_a^\Gamma = \vv{x}_a - \phi(\vv{x}_a) \frac{\nabla\phi(\vv{x}_a)}{||\nabla\phi(\vv{x}_a)||}
\label{eq:XOnGamma}
\end{equation}
with bilinear interpolation. $\vv{x}_a^\Gamma$ is a rough approximation for $\vv{x}_a$'s projection onto $\Gamma^n$.  Next, we perform negative-curvature normalization by negating $\mathcal{p}^\circ$'s level-set values if $\kappa_a^\circ > 0$.  After this step, we proceed to the last preparations before leaving the data packet in a well-suited form for the {\tt Preprocess()} module.  Finally, upon post-processing $\mathcal{F}_{c,f}(\cdot)$'s output in \cref{fig:ECNet}, we check $\kappa_a^\circ$ and restore the sign for the machine-learning-corrected level-set value.

We have also resorted to \textit{sample reorientation} to promote neural learning.  This technique helps minimize feature variations and has proven effective in elementary computer vision applications, such as face recognition with \textit{eigenfaces} \cite{Turk;Pentland;Eigenfaces;1991, Parker;CS170A;2016}.  \Cref{fig:ReorientedSample} shows a data packet $\mathcal{p}$ after reorienting the original $\mathcal{p}^\circ$, assuming that $\kappa_a^\circ < 0$.  Standard-forming $\mathcal{p}^\circ$ requires rotating its stencil until the angle between the horizontal axis and $-\hat{\vv{u}}_a$ lies between $0$ and $\pi / 2$.  Reorientation does not change intrinsic level-set values, curvature, and distance between $\vv{x}_a$ and $\vv{x}_d$; however, it affects vector quantities by pre-multiplying them with a rotation transformation $R(\theta)$, where $\theta = \pm k \pi/2$, and $k \in \{0, 1, 2\}$.  After reorientation, the arrival point becomes the origin of a local coordinate system that contains all the information we need in its first quadrant.

Following our description of the error-correcting multilayer perceptron, we present the {\tt MLSemiLagrangian()} procedure in \Cref{alg:MLSemiLagrangian} next.  This algorithm combines $\mathcal{F}_{c,f}(\cdot)$ with the parallel semi-Lagrangian schemes in \cite{Mirzadeh;etal:16:Parallel-level-set}.  Its formal parameters\footnote{For consistency, we represent one-element nodal variables as $m$-vectors in lowercase bold faces (e.g., $\vv{\phi}^n$) and variables with $d > 1$ values per node as $d$-by-$m$ matrices in caps (e.g., $U^n$).  In the {\tt p4est} terminology, $m$ is the number of independent vertices that $\mathcal{G}^n$ is aware of.} include $\mathcal{F}_{c,f}(\cdot)$, the quadtree data structure $\mathcal{G}^n \in \hat{\mathcal{S}}^n$, and the scalar-/vector-field vertex values necessary to construct the neural network inputs and advance the moving front.  Furthermore, \Cref{alg:MLSemiLagrangian} assumes that we have optimized $\mathcal{F}_{c,f}(\cdot)$ for reducing numerical diffusion in the advected level-set field for nodes near $\Gamma^n$ in a mesh with an $\ell_c^{\max}$ maximum level of refinement.  Similar to its numerical counterpart, the {\tt MLSemiLagrangian()} method produces new level-set values and updates the adaptive mesh.  In addition, it generates an array of vertex coordinates that we should protect during selective reinitialization.  As outlined above, our goal is to output nodal level-set values in $\vv{\phi}^{n+1}$ with coordinates in $\mathcal{C}^{n+1}$, closely resembling the free boundary motion in a higher-resolution grid.

\begin{remark}
We provide \Cref{alg:MLSemiLagrangian} and the rest of the routines in the most general way.  That is, we have used conventional formulae to determine the time step $\Delta t$ based on a user-defined $CFL$ constant and a possibly unbounded velocity field over $\Omega$.  However, to validate our assumptions and narrow down the learning problem, we have set $CFL = 1$ and $\max{||\vv{u}||} = 1$ for all nodes in $\mathcal{G}$.  Therefore, as given by the first instruction of \Cref{alg:MLSemiLagrangian}, we have that $\Delta t = h$, and we can thus recover the data characterization provided in \cref{fig:Sample}.
\label{rmk:Constraints}
\end{remark}


\begin{algorithm}[!t]
\SetAlgoLined
\SetKwFunction{reconstruct}{Reconstruct}
\SetKwFunction{interpolate}{Interpolate}
\SetKwFunction{hmin}{hmin}
\SetKwFunction{mpiallreduce}{MPI\_Allreduce}
\SetKwFunction{collectdatapackets}{CollectDataPackets}
\SetKwFunction{preprocess}{Preprocess}
\SetKwFunction{updateghostvalues}{UpdateGhostValues}
\SetKwFunction{computedeparturepoints}{ComputeDeparturePoints}
\SetKwFunction{interpolate}{Interpolate}
\SetKwFunction{adjustlswithml}{AdjustLevelSetWithMLSolution}
\SetKwFunction{refineandcoarsen}{refineAndCoarsen}
\SetKwFunction{partition}{partition}
\SetKwFunction{getlistofkeys}{getListOfKeys}
\SetKwFunction{append}{append}
\SetKwFunction{sign}{Sign}
\SetKwFunction{getnodeswithflow}{GetCoordsWithNegativeFlow}

\KwIn{error-correcting neural network for maximum coarse refinement level $c$ and maximum fine refinement level $f$, $\mathcal{F}_{c,f}$; grid structure, $\mathcal{G}^n$; nodal level-set values, $\vv{\phi}^n$; nodal velocities, $U^n$; level-set second spatial derivatives, $\Phi_{xx}^n$; nodal normal unit vectors, $N^n$; nodal curvature values, $\vv{\kappa}^n$; $CFL$ number.}
\KwResult{updated nodal structure, $\mathcal{G}^{n+1}$, nodal level-set values, $\vv{\phi}^{n+1}$, and array of coordinates updated with $\mathcal{F}_{c,f}(\cdot)$ to be protected with selective reinitialization, $\mathcal{C}^{n+1}$.}
\BlankLine

$h \leftarrow \mathcal{G}^n.$\hmin{}$, \quad \Delta t_\ell \leftarrow CFL \cdot h/ \max{||U^n||}$\tcp*[r]{Stepping variables}
$\Delta t \leftarrow$ \mpiallreduce{$\Delta t_\ell$, {\tt MPI\_MIN}}\tcp*[r]{Retrieve minimum $\Delta t$ across processes}
$\mathcal{H}^n \leftarrow$ \reconstruct{$\mathcal{G}^n$}\tcp*[r]{Refer to Algorithm \href{https://www.sciencedirect.com/science/article/pii/S002199911630242X\#fg0030}{1} in \cite{Mirzadeh;etal:16:Parallel-level-set}}
\BlankLine
			
\tcp{Prepare computation of machine-learning-corrected level-set values for nodes next to $\Gamma^n$}
$(\mathcal{P},\, \mathcal{C}) \leftarrow$ \collectdatapackets{$\mathcal{G}^n$, $\mathcal{H}^n$, $\vv{\phi}^n$, $U^n$, $\Phi_{xx}^n$, $N^n$, $\vv{\kappa}^n$, $\Delta t$}\tcp*[r]{See \Cref{alg:CollectDataPackets}}
$\mathcal{S}_\kappa \leftarrow [~], \quad B \leftarrow [~]$\tcp*[r]{List of true sample curvature signs and matrix of network input samples}

\ForEach{data packet $\mathcal{p} \in \mathcal{P}$}{

	$\mathcal{S}_\kappa.$\append{\sign{$\mathcal{p}.\kappa_a$}}\;
	transform $\mathcal{p}$, so that $\mathcal{p}.\kappa_a$ is negative\;
	rotate $\mathcal{p}$, so that the angle of $-\mathcal{p}.\hat{\vv{u}}_a$ lies between $0$ and $\pi/2$\;
	\BlankLine
	
	\tcp{Produce two network input samples for each data packet to improve accuracy}
	$B.$\append{$[$\preprocess{$\mathcal{p}$, $h$},~ $\frac{1}{h}\mathcal{p}.\phi_d]$}\;
	let $\mathcal{p}'$ be the reflected data packet about line $y = x + \beta$ going through the arrival point\;
	$B.$\append{$[$\preprocess{$\mathcal{p'}$, $h$},~ $\frac{1}{h}\mathcal{p'}.\phi_d]$}\;
	
}
\BlankLine

\tcp{Launch batch network predictions}
$\mathcal{O} \leftarrow \mathcal{F}_{c,f}(B)$\;\label{alg:MLSemiLagrangian.prediction}
\BlankLine

\tcp{Collect neural predictions and broadcast them across processes}
$\mathcal{M}_\phi \leftarrow \varnothing$\tcp*[r]{Map of (local and ghost) node coordinates to $\phi$ values computed with $\mathcal{F}_{c,f}(\cdot)$}
\For{$\iota \leftarrow 0$ to $(|\mathcal{P}| - 1)$}{

	$\phi_d^\star \leftarrow \frac{h}{2}\left(\mathcal{O}[2\iota] + \mathcal{O}[2\iota + 1]\right)$\tcp*[r]{Average neural prediction}
	\BlankLine
	
	\If{$\frac{1}{h}|\phi_d^\star - \mathcal{P}[\iota].\phi_d| > 0.15$ or $|\phi_d^\star - \mathcal{P}[\iota].\phi_a| \geqslant h$}{
		$\phi_d^\star \leftarrow \mathcal{P}[\iota].\phi_d$\tcp*[r]{Catch possible divergence in neural prediction}
	}
	
	fix sign of $\phi_d^\star$ according to $\mathcal{S}_\kappa[\iota]$\;
	$\mathcal{M}_\phi[~\mathcal{C}[\iota]~] \leftarrow \phi_d^\star$ only if $\phi_d^\star$ did not revert to numerical approximation\;
}
\updateghostvalues{$\mathcal{M}_\phi$}\tcp*[r]{Use MPI to gather values for ghost nodes updated via $\mathcal{F}_{c,f}(\cdot)$}
\BlankLine

\tcp{Couple machine-learning-corrected trajectory with the rest of the level-set values}
$\mathcal{G}_0^{n+1} \leftarrow \mathcal{G}^n$\;

\While{true}{
	$X_d \leftarrow$ \computedeparturepoints{$\mathcal{G}_0^{n+1}$, $U^n$, $\Delta t$}\tcp*[r]{Using \cref{eq:SLMidpointDeparturePoint,eq:SLIntermediateVelocity}}
	$\vv{\phi}^{n+1} \leftarrow$ \interpolate{$\mathcal{H}^n$, $\vv{\phi}^n$, $X_d$}\tcp*[r]{Refer to Algorithm \href{https://www.sciencedirect.com/science/article/pii/S002199911630242X\#fg0050}{2} in \cite{Mirzadeh;etal:16:Parallel-level-set}}
	\adjustlswithml{$\vv{\phi}^{n+1}$, $\mathcal{M}_\phi$}\tcp*[r]{Correct $\phi$ for nodes with coordinates in $\mathcal{M}_\phi$}
	$\mathcal{G}^{n+1} \leftarrow \mathcal{G}_0^{n+1}.$\refineandcoarsen{$\vv{\phi}^{n+1}$}\tcp*[r]{Using criterion in \cref{eq:RefinementCriterion}}
	
	\eIf{$\mathcal{G}^{n+1} \neq \mathcal{G}_0^{n+1}$}{
		$\mathcal{G}^{n+1}.$\partition{}\;
		$\mathcal{G}_0^{n+1} \leftarrow \mathcal{G}^{n+1}$\;
	}{
		\textbf{break}\;
	}
}
\BlankLine

$\mathcal{W}^{n+1} \leftarrow $\getnodeswithflow{$\mathcal{H}^n$, $U^n$, $N^n$, $\mathcal{G}^{n+1}$, $\vv{\phi}^{n+1}$}\tcp*[r]{Points `lagging behind' $\Gamma^{n+1}$}
$\mathcal{C}^{n+1} \leftarrow \mathcal{M}_\phi.$\getlistofkeys{}$\, \cap\,\mathcal{W}^{n+1}$\tcp*[r]{Vertices to protect during reinitialization}
\Return $(\mathcal{G}^{n+1},\, \vv{\phi}^{n+1},\, \mathcal{C}^{n+1})$\;

\caption{\small $(\mathcal{G}^{n+1},\, \vv{\phi}^{n+1},\, \mathcal{C}^{n+1}) \leftarrow$ {\tt MLSemiLagrangian(}$\mathcal{F}_{c,f}(\cdot)$, $\mathcal{G}^n$, $\vv{\phi}^n$, $U^n$, $\Phi_{xx}^n$, $N^n$, $\vv{\kappa}^n$, $CFL${\tt )}: Update level-set values $\vv{\phi}^{n+1}$ from $\vv{\phi}^n$ using a numerical semi-Lagrangian scheme with error correction provided by $\mathcal{F}_{c,f}(\cdot)$.}
\label{alg:MLSemiLagrangian}
\end{algorithm}

The first step to incorporate machine learning into semi-Lagrangian advection in \Cref{alg:MLSemiLagrangian} is to collect data packets in $\mathcal{P}$ and their coordinates in $\mathcal{C}$ for vertices next to $\Gamma^n$.  To do so, we get the local hierarchical reconstruction $\mathcal{H}^n$ and retrieve the information in \cref{eq:DataPacket,fig:OriginalSample} by calling {\tt CollectDataPackets()}.  We discuss {\tt CollectDataPackets()} in some detail in \Cref{subsec:DataPacketCollection}.  Then, we apply negative-curvature normalization and reorientation for each data packet $\mathcal{p} \in \mathcal{P}$ to arrive at the standard form referenced in \cref{fig:ReorientedSample}.  At the same time, we populate the ancillary array $\mathcal{S}_\kappa$ to record whether $\vv{x}_a$ is proximal to a concave or convex interface region.  For every machine-learning-suitable packet, we generate two samples: one with the preprocessed standard blob $\mathcal{p}$ and another with the augmented data packet $\mathcal{p}'$.  For example, \cref{fig:AugmentedSample} illustrates the reflected data packet for the reoriented $\mathcal{p}$ shown in \cref{fig:ReorientedSample}.  The rationale for double-sampling is that the departure point's level-set value does not change if we reflect its blob about a slope-one, straight line going through the arrival point.  By leveraging this invariance, we can increase the accuracy of $\phi_d^\star$ if we average the predictions for the corresponding input samples

\begin{equation}
[\texttt{Preprocess(}\mathcal{p},\, h\texttt{)},\, \tfrac{1}{h}\mathcal{p}.\phi_d] \quad \textrm{and} \quad 
[\texttt{Preprocess(}\mathcal{p}',\, h\texttt{)},\, \tfrac{1}{h}\mathcal{p}'.\phi_d].
\label{eq:Samples}
\end{equation}
Notice that these samples are two-part input vectors that match $\mathcal{F}_{c,f}(\cdot)$'s expected format.  Recently, practitioners have shown that such symmetry-preserving transformations and succeeding averaging can improve machine-learning IR in the level-set method \cite{Buhendwa;Bezgin;Adams;IRinLSwithML;2021}.  Here, we have realized these transformations with a few permutations over $\mathcal{p}$'s vertex values.

For efficiency, we accommodate all the samples in matrix $B$.  Then, we launch $\mathcal{F}_{c,f}(\cdot)$ in batch mode to predict the $h$-normalized, error-corrected level-set values for the columns in $B$.  Upon $h$-denormalization, averaging, and sign-restoration using $\mathcal{S}_\kappa$, we recover the improved $\phi_d^\star$.  Even though we expect $\phi_d^\star$ to be more accurate than $\phi_d$, outlying input patterns might lead to wild predictions in some scenarios.  To account for these cases, we revert to $\phi_d$ whenever $\mathcal{F}_{c,f}(\cdot)$ predicts $\Gamma^n$ should move by more than $h$ or if the $h$-relative difference between $\phi_d$ and $\phi_d^\star$ is over 15\%.  Neural network divergence is not uncommon in numerical experiments, and we have reported extreme cases when approximating mean curvature \cite{Larios;Gibou;HybridCurvature;2021}.  In this research, we mitigate such a vulnerability by taking simple precautions.

While collecting $\phi_d^\star$, we simultaneously build a hash map $\mathcal{M}_\phi$ that links nodal coordinates to machine-learning-computed level-set values.  $\mathcal{M}_\phi$ works as a cache that prevents calling $\mathcal{F}_{c,f}(\cdot)$ more than once.  Also, it is essential for data synchronization across processes and sharing what locally aware nodal $\phi_d^\star$ values should persist after re-gridding.  To instrument this map and update the ghost layer, we have used {\tt MPI} \cite{MPI14}, {\tt PETSc} vectors \cite{Balay;Brown;Buschelman;etal:12:PETSc-Web-page}, and the infrastructure in {\tt p4est}\cite{Burstedde;Wilcox;Ghattas:11:p4est:-Scalable-Algo}.

The concluding part of \Cref{alg:MLSemiLagrangian} merges the machine learning corrections with the discrete level-set function advected numerically in the rest of the mesh.  The coupling occurs in an iterative process based on Algorithm \href{https://www.sciencedirect.com/science/article/pii/S002199911630242X\#fg0060}{3} in \cite{Mirzadeh;etal:16:Parallel-level-set}.  In each iteration, we compute the departure points $X_d$ and their level-set values $\vv{\phi}^{n+1}$ using \crefrange{eq:SLLevelSetValue}{eq:SLIntermediateVelocity} and interpolation from data at time $t^n$.  The {\tt Interpolate()} method entails a series of distributed actions that rely on synchronization and $\mathcal{H}^n$ to retrieve scattered information from different sites.  Then, we replace the numerical level-set values with the machine-learning ones in $\mathcal{M}_\phi$ for independent nodes whose coordinates match with a key in the cache.  Afterward, we refine and coarsen the new grid to produce $\mathcal{G}^{n+1}$ according to the criterion in \cref{eq:RefinementCriterion}.  We execute these operations repeatedly until $\mathcal{G}^{n+1}$ converges.  Usually, convergence requires no more steps than the maximum number of subdivisions in any quadtree \cite{Mirzadeh;etal:16:Parallel-level-set}.  

Upon exiting the update loop, we gather the vertex coordinates in $\mathcal{M}_\phi$ and intersect them with $\mathcal{W}^{n+1}$, which results from calling {\tt GetCoordsWithNegativeFlow()}.  This ancillary subroutine locates vertices within a narrow band of half-width $2h\sqrt{2}$ around $\Gamma^{n+1}$, where the angular condition

\begin{equation}
\mathrm{acos}\left(-\texttt{sign}\left(\phi^{n+1}(\vv{x})\right) \hat{\vv{n}}^n(\vv{x}) \cdot \hat{\vv{u}}^n(\vv{x})\right) \leqslant \theta_{\mathcal{W}}
\label{eq:WithTheFlowCondition}
\end{equation}
holds for $\theta_{\mathcal{W}} = 19\pi / 36 = 95^\circ$.  In this expression, $\vv{x}$ is a point in $\mathcal{G}^{n+1}$, and $\hat{\vv{n}}$ and $\hat{\vv{u}}$ are the normal and velocity unit vectors interpolated from data at $t^n$.  Intuitively, \cref{eq:WithTheFlowCondition} selects nodes lagging behind the motion of the new interface.  Our goal is to protect enhanced level-set values in $\vv{\phi}^{n+1}$ from indiscriminate redistancing on points on one side of $\Gamma^{n+1}$ at the current iteration.  Then, we resort to the conventional semi-Lagrangian scheme to smooth out the moving front from the opposite side at $t^{n+2}$.  In other words, improving the free boundary trajectory is not a task of the {\tt MLSemiLagrangian()} function alone; it demands a delicate balance between machine learning correction, selective reinitialization, and pure numerical advection.  In particular, we should enable \Cref{alg:MLSemiLagrangian} whenever it is its turn in the alternating strategy, and the constraints described in \Cref{rmk:Constraints} are valid (i.e., $\Delta t = h$, $CFL=1$, and $\max\|\vv{x}\| = 1$ for all $\vv{x} \in \Omega$).  These conditions guarantee that the input information for $\mathcal{F}_{c,f}(\cdot)$ is compatible with the data collected during training.  Furthermore, as shown in \Cref{subsec:Training}, alternating between solvers is necessary not only at the inference stage but also when optimizing our numerical-error estimator.


\colorsubsection{Data-packet collection for learning and inference}
\label{subsec:DataPacketCollection}

The {\tt CollectDataPackets()} subroutine in \Cref{alg:CollectDataPackets} is a straightforward procedure that exploits the multi-process interpolation mechanism of \cite{Mirzadeh;etal:16:Parallel-level-set} to extract information from local cells owning departure points.  This information is necessary during inference (see \Cref{alg:MLSemiLagrangian}) and for constructing the training data set (see \Cref{subsec:Training}). 

\begin{figure}[t]
	\centering
	\includegraphics[width=6cm]{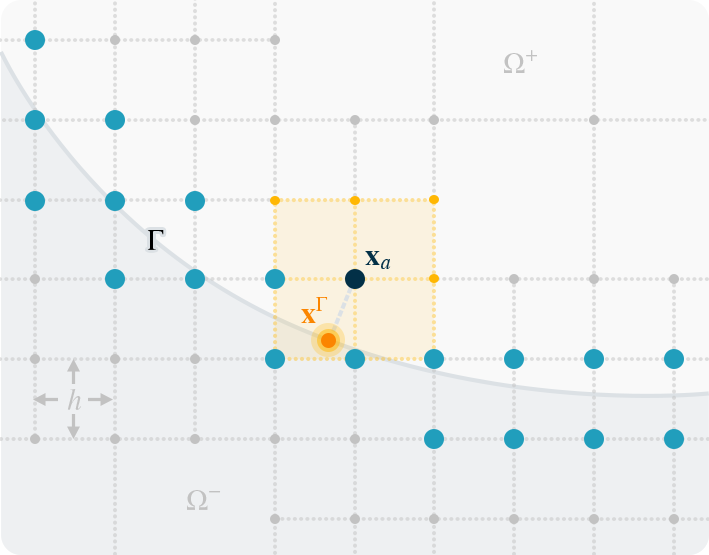}
	\caption{Sampled nodes adjacent to a concave interface region in a quadtree mesh.  Vertices with $h$-uniform nine-point stencils appear in light blue.  We gather these nodes with the {\tt GetLocalNodesNextToGamma()} subroutine inside \Cref{alg:CollectDataPackets}.  Among them, we shade in yellow the stencil of the (arrival) grid point $\vv{x}_a$ colored in dark blue.  The rough estimation to the closest point on $\Gamma$ to $\vv{x}_a$ is $\vv{x}^\Gamma$ (shown in orange).  (Color online.)}
	\label{fig:Nodes}
\end{figure}

We begin \Cref{alg:CollectDataPackets} by extracting a set $\mathcal{N}$ of locally owned nodes near $\Gamma$.  Every node $\mathcal{n} \in \mathcal{N}$ with coordinates $\mathcal{n}.\vv{x} = (\mathcal{n}.x, \mathcal{n}.y)$ has an $h$-uniform, nine-point stencil and satisfies at least one of the four conditions: $\phi(\mathcal{n}.x, \mathcal{n}.y) \cdot \phi(\mathcal{n}.x \pm h, \mathcal{n}.y) \leqslant 0$, or $\phi(\mathcal{n}.x, \mathcal{n}.y) \cdot \phi(\mathcal{n}.x, \mathcal{n}.y \pm h) \leqslant 0$.  \Cref{fig:Nodes} portrays a group of vertices typically found in $\mathcal{N}$.

For each node $\mathcal{n} \in \mathcal{N}$, first, we compute its departure point $\vv{x}_d$ and midpoint velocity $\hat{\vv{u}}$ with \cref{eq:SLMidpointDeparturePoint,eq:SLIntermediateVelocity} via quadratic interpolation.  Next, we use $\vv{x}_d$ and $\hat{\vv{u}}$ to check the validity of $\mathcal{n}$ and decide if we should continue processing it.  A grid node is valid when its intermediate velocity is nonzero, and its departure point lies within a cell $C$ at the maximum level of refinement.  Alongside the $CFL$ and velocity-norm restrictions, these new constraints help confine the problem's complexity and favor data consistency.  

After validation, we fetch a partially populated $\mathcal{p}$ using the information available for $\mathcal{n}$ in $C$.  To do so, we employ the {\tt FechDataPacket()} subroutine, which extends the multi-process {\tt Interpolate()} function.  Upon calling {\tt FechDataPacket()}, $\mathcal{p}$ will hold data for $(\tilde{x}_d, \tilde{y}_d)$, $(\phi)_{ij}$, $(\vv{u}_d)_{ij}$, $(\phi_{xx})_d$, and $(\phi_{yy})_d$.  Then, we fill in the missing fields in $\mathcal{p}$ by computing $\mathcal{n}$'s distance $d$ from $\vv{x}_d$ and querying its nodal level-set value $\phi_a$.  In addition, we bilinearly interpolate $\kappa$ at $\mathcal{n}$'s projection onto $\Gamma$, as specified by \cref{eq:XOnGamma}, and use quadratic interpolation to retrieve $\phi_d$ at $\vv{x}_d$, as given by \cref{eq:SLMidpointDeparturePoint}.  Once more, note that our interpolative tasks require a local hierarchical reconstruction $\mathcal{H}$ to determine which partition owns $\vv{x}_d$.  For a detailed description of the distributed interpolation procedure, we refer the reader to Algorithm \href{https://www.sciencedirect.com/science/article/pii/S002199911630242X\#fg0050}{2} in \cite{Mirzadeh;etal:16:Parallel-level-set}.  By the end of our \Cref{alg:CollectDataPackets}, we will have all the valid-vertex data packets in $\mathcal{P}$ and their respective coordinates (i.e., arrival points) in $\mathcal{C}$.


\begin{algorithm}[!t]
\SetAlgoLined
\SetKwFunction{getlocalnodesnexttogamma}{GetLocalNodesNextToGamma}
\SetKwFunction{computedeparturepointandmidvel}{ComputeDeparturePointAndMidVel}
\SetKwFunction{isinvalid}{IsInvalid}
\SetKwFunction{fetchdatapacket}{FetchDataPacket}
\SetKwFunction{interpolate}{Interpolate}
\SetKwFunction{append}{append}

\KwIn{grid structure, $\mathcal{G}$; tree hierarchy, $\mathcal{H}$; nodal level-set values, $\vv{\phi}$; nodal velocities, $U$; level-set second spatial derivatives, $\Phi_{xx}$; nodal normal unit vectors, $N$; nodal curvature values, $\vv{\kappa}$; time step, $\Delta t$.}
\KwResult{array of data packets, $\mathcal{P}$; array of coordinates, $\mathcal{C}$.}
\BlankLine

$\mathcal{N} \leftarrow$ \getlocalnodesnexttogamma{$\mathcal{G}$, $\vv{\phi}$}\;
$\mathcal{P} \leftarrow [~], \quad \mathcal{C} \leftarrow [~]$\;
\BlankLine

\ForEach{node $\mathcal{n} \in \mathcal{N}$}{
	$(\vv{x}_d,\, \hat{\vv{u}}) \leftarrow$ \computedeparturepointandmidvel{$\mathcal{G}$, $U$, $\mathcal{n}$, $\Delta t$}\tcp*[r]{Using \cref{eq:SLMidpointDeparturePoint,eq:SLIntermediateVelocity}}
	\lIf{$||\hat{\vv{u}}|| \leqslant \epsilon$ or \isinvalid{$\mathcal{G}$, $\vv{x}_d$}}{skip node $n$}
	\BlankLine
	
	$\mathcal{p} \leftarrow$ \fetchdatapacket{$\mathcal{H}$, $\vv{\phi}$, $\Phi_{xx}$, $\vv{x}_d$, $h$}\tcp*[r]{Populate most of $\mathcal{p}$ data via interpolation}
	\BlankLine
	
	\tcp{Fill in missing information in $\mathcal{p}$}
	$\mathcal{p}.d \leftarrow ||\vv{x}_d - \mathcal{n}.\vv{x}||$\tcp*[r]{Distance between arrival and departure points}
	$\mathcal{p}.\hat{\vv{u}}_a \leftarrow \hat{\vv{u}}$\tcp*[r]{Midpoint velocity used at arrival point}
	$\mathcal{p}.\phi_a \leftarrow \vv{\phi}[\mathcal{n}]$\tcp*[r]{Level-set value at arrival point}
	$\vv{x}^\Gamma \leftarrow \mathcal{n}.\vv{x} - \vv{\phi}[\mathcal{n}](N[\mathcal{n}])$\tcp*[r]{Using \cref{eq:XOnGamma,eq:NormalAndCurvature}}
	$\mathcal{p}.\kappa_a \leftarrow$ \interpolate{$\mathcal{H}$, $\vv{\kappa}$, $\vv{x}^\Gamma$}\tcp*[r]{Curvature at closest point to $\mathcal{n}.\vv{x}$ on $\Gamma$}
	$\mathcal{p}.\phi_d \leftarrow$ \interpolate{$\mathcal{H}$, $\vv{\phi}$, $\vv{x}_d$}\tcp*[r]{Numerical baseline level-set value}
	\BlankLine
	
	$\mathcal{P}.$\append{$\mathcal{p}$}\;
	$\mathcal{C}.$\append{$\mathcal{n}.\vv{x}$}\;
}				
\Return $(\mathcal{P},\, \mathcal{C})$\;

\caption{\small $(\mathcal{P},\, \mathcal{C}) \leftarrow$ {\tt CollectDataPackets(}$\mathcal{G}$, $\mathcal{H}$, $\vv{\phi}$, $U$, $\Phi_{xx}$, $N$, $\vv{\kappa}$, $\Delta t${\tt )}: Collect data packets for nodes next to $\Gamma$.}
\label{alg:CollectDataPackets}
\end{algorithm}


\colorsubsection{Training}
\label{subsec:Training}

The motivation for developing our error-correcting neural network stems from the promising results in machine learning applications that rest on image super-resolution methodologies \cite{Dong;Loy;He;SuperResolution;2014, Xie;etal;TempoGAN;2018, Bar-Sinai;Hoyer;Hickey;Brenner;LearningData-DrivenPDEs;2019, Liu;etal;DLMthdsSuperResoltnReconstTurbFlows;2020, Pathak;etal;MLToAugCoarseGridCFD;2020, Zhuang;etal;LrndDiscForPassSclrAdvctn2D;2021}.  Under these paradigms, to recover the information lost on a low-resolution grid, one should solve the same problem on a highly resolved mesh and assess the difference between both systems.  In other words, the coarse-grid solution should mimic the fine-grid solution as much as possible with the help of the available data.  In our case, we have already stated what information we can leverage when tracking the evolution of a moving front.  Therefore, to train $\mathcal{F}_{c,f}(\cdot)$, we must assemble a data set $\mathcal{D}$ with curated low-resolution input data packets and high-resolution (departure) output target level-set values.  Later, we can contrast the inferred $\phi_d^\star$ values in \cref{fig:ECNet} with their reference counterparts to optimize $\mathcal{F}_{c,f}(\cdot)$ via stochastic gradient descent.

\colorsubsubsection{Data-set generation}
\label{subsubsec:DataSetGeneration}

First, we explain how to build the learning data set $\mathcal{D}$.  The high-level recipe is straightforward as long as we consider just one coarse resolution at a time.  If various mesh sizes are necessary, one must produce their corresponding data sets separately and train their respective neural networks.  This idea would be the semi-Lagrangian-scheme equivalence to the curvature neural-network dictionaries introduced in previous studies \cite{LALariosFGibou;LSCurvatureML;2021, Larios;Gibou;HybridCurvature;2021}.  Now, assuming that we discretize $\Omega$ with adaptive Cartesian grids, let $\ell_c^{\max}$ and $\ell_f^{\max}$ be the coarse- and fine-mesh maximum levels of refinement (refer to \cref{fig:Quadtree}).  As stated in \Cref{subsec:ProblemDefinition}, we presume that $\ell_c^{\max} < \ell_f^{\max}$ and $h_c > h_f$.  Thus, to populate $\mathcal{D}$, it suffices to set up a \textit{direct numerical simulation} (DNS) with a predefined level-set function.  Then, one must advect its interface simultaneously on a coarse and a fine grid with an external velocity field and collect samples periodically until some $t^{end}$.  To accommodate the alternating mechanism described in \Cref{subsec:ErrorCorrectingNNetsForSLAdvect} and illustrated in \cref{fig:Overview}, we note that data collection takes place every other iteration.  Furthermore, we can promote pattern variation and generalization if we repeatedly perform these steps for random configurations.  This randomization technique is the backbone of the {\tt GenerateDataSet()} procedure presented in \Cref{alg:GenerateDataSet}.

\Cref{alg:GenerateDataSet} is combinatorial.  For each velocity field, we draw a level-set-function base parameter from a range of evenly distributed values.  Then, we spawn the proper initial level-set configuration at $N_C$ different places within the computational domain.  For simplicity, we use circular-interface level-set functions whose base parameters are their radii, and initial configurations are their interface centers.  In our case, $N_F = 7$ velocity fields and $N_C = 4$ locations per radius yield well-sized data sets from simulations that last up to $t^{end} = 0.5$.  Also, to minimize concurrency issues, we carry out \Cref{alg:GenerateDataSet} in a uni-process system that can fit a computational $2$-by-$2$ discretized $\Omega \equiv [-1, +1]^2$ with four unit-square quadtrees.

As mentioned before, if $CFL = 1$ and $\max{||\vv{u}(\vv{x})||} = 1$, $\forall \vv{x} \in \Omega$, all nodes of interest will coincide with one vertex in their departure-point-owning cells (see \cref{fig:Sample}).  By enforcing this constraint, we ensure that those departure points never lie farther than two cells away from $\Gamma^n$.  Consequently, it is enough to require a coarse-grid uniform band around $\Gamma^n$ of half-width $B_c = 2$ (measured in cell diagonals) in anticipation of \Cref{alg:CollectDataPackets}'s validity test.  For accuracy, we also impose a uniform band of half-width $B_f$ in the fine grid.  While $B_c$ aids with data consistency and problem simplification, $B_f$ favors precision when querying target level-set values.  In particular, $B_f$ is relevant to how we interleave advection of low- and high-resolution level-set values: first, we advance the fine-grid level-set function; then, we use that updated trajectory to determine which reference (i.e., more accurate) level-set values should become targets to the sampled coarse-grid arrival points.  Therefore, if $\Delta t_c$ is the coarse time step, the advected-fine-mesh band around $\Gamma_f(t^n + \Delta t_c)$ must be wide enough to encompass the coarse-grid, selected nodes at time $t^n$.  The expression for setting $B_f$ in the preamble of \Cref{alg:GenerateDataSet} reassures these  properties and a bit more.  Since we have discretized $\Omega$ with unit-square Cartesian quadtrees, the ratio $h_c : h_f$ from coarse to fine mesh size is a power of two.  Accordingly, after advecting the high-resolution level-set function, we find that the sampled coarse-grid points overlap with vertices within the uniform band around the new fine-grid zero-isocontour.  The latter is a desirable feature when assembling $\mathcal{D}$, considering that we use interpolation to fetch information from the fine mesh.


\begin{algorithm}[!t]
\SetAlgoLined
\SetKwFunction{linspace}{LinSpace}
\SetKwFunction{levelset}{LevelSet}
\SetKwFunction{genrandvelfield}{GenerateVelocityField}
\SetKwFunction{gengrid}{GenerateGrid}
\SetKwFunction{evaluate}{Evaluate}
\SetKwFunction{reinitialize}{Reinitialize}
\SetKwFunction{intadvectandcollectsamples}{IntAdvectAndCollectSamples}

\KwIn{maximum refinement levels per unit length, $\ell_c^{\max}$ and $\ell_f^{\max}$; final time per configuration, $t^{end}$; number of iterations for level-set redistancing, $\nu$; number of random velocity fields, $N_F$; number of circular-interface level-set functions with the same radius per velocity field, $N_C$; uniform band half-width around coarse interface, $B_c$; coarse grid reset frequency, $R_{freq}$; $CFL$ number.}
\KwResult{data set with learning samples, $\mathcal{D}$.}
\BlankLine

define $\Omega \equiv [-1,+1]^2 \subset \mathbb{R}^2$, with 2 quadtrees in each Cartesian direction\;

$B_f \leftarrow \frac{7}{4} B_c \cdot 2^{\left(\ell_f^{\max} - \ell_c^{\max} - 1\right)}$\tcp*[r]{Uniform band half-width around fine grid}
\BlankLine

$h_c \leftarrow 2^{-\ell_c^{\max}}, \quad h_f \leftarrow 2^{-\ell_f^{\max}}$\tcp*[r]{Coarse and fine mesh sizes}
$r_{\min} \leftarrow 5h_c, \quad r_{\max} \leftarrow 0.25$\tcp*[r]{Circular-interface radial bounds}
$N_r \leftarrow \left\lceil 3 \left(\frac{r_{\max} - r_{\min}}{h_c}\right) \right\rceil + 1$\tcp*[r]{Number of distinct radii}
\BlankLine

$\mathcal{D} \leftarrow \varnothing$\;
\BlankLine

\tcp{Perform multiple simulations for $N_F$ velocity fields}
\For{$i \leftarrow 1$ to $N_F$}{
	
	$\mathcal{V}_i(\cdot) \leftarrow$ \genrandvelfield{}\tcp*[r]{Generate a random, divergence-free velocity field}
	normalize $\mathcal{V}_i(\cdot)$\tcp*[r]{and normalize it so that $\max{||\mathcal{V}_i(\vv{x})||} = 1, \forall \vv{x} \in \Omega$}
	\BlankLine
	
	\tcp{Evaluate $N_r$ evenly spaced interface radii $r \in [r_{min}, r_{max}]$}
	\ForEach{radius $r_\circ$ in \linspace{$r_{min}$, $r_{max}$, $N_r$}}{
		\BlankLine
	
		\tcp{Repeat each radius $r_\circ$ as many as $N_C$ times}
		\For{$j \leftarrow 1$ to $N_C$}{
			
			$(x_\circ, y_\circ) \sim \mathcal{U}(-1/2, +1/2)$\tcp*[r]{Random center}
			$\phi_\circ(\cdot) \leftarrow$ \levelset{$r_\circ$, $x_\circ$, $y_\circ$}\tcp*[r]{Build circular-interface level-set function}
			\BlankLine
			
			$\mathcal{G}_c^0 \leftarrow$ \gengrid{$\Omega$, $\phi_\circ(\cdot)$, $\ell_c^{\max}$, $B_c$}\tcp*[r]{Build initial grids}
			$\mathcal{G}_f^0 \leftarrow$ \gengrid{$\Omega$, $\phi_\circ(\cdot{x})$, $\ell_f^{\max}$, $B_f$}\;
			\BlankLine
			
			$\vv{\phi}_c^0 \leftarrow$ \evaluate{$\phi_\circ(\cdot)$, $\mathcal{G}_c^0$}, $\quad U_c^0 \leftarrow$ \evaluate{$\mathcal{V}_i(\cdot)$, $\mathcal{G}_c^0$}\tcp*[r]{Sample level-set function}
			$\vv{\phi}_f^0 \leftarrow$ \evaluate{$\phi_\circ(\cdot)$, $\mathcal{G}_f^0$}, $\quad U_f^0 \leftarrow$ \evaluate{$\mathcal{V}_i(\cdot)$, $\mathcal{G}_f^0$}\tcp*[r]{and velocity field}
			\BlankLine
			
			$\vv{\phi}_c^0 \leftarrow$ \reinitialize{$\mathcal{G}_c^0$, $\vv{\phi}_c^0$, $\nu$}, $\quad \vv{\phi}_f^0 \leftarrow$\reinitialize{$\mathcal{G}_f^0$, $\vv{\phi}_f^0$, $B_c \cdot \nu$}\;
			\BlankLine
			
			\tcp{Collect learning samples into $\mathcal{D}_c$ (see \Cref{alg:IntAdvectAndCollectSamples})}
			$\mathcal{D}_c \leftarrow$ \intadvectandcollectsamples{$\mathcal{G}_c^0$, $\mathcal{G}_f^0$, $\vv{\phi}_c^0$, $\vv{\phi}_f^0$, $U_c^0$, $U_f^0$, $t^{end}$, $\nu$, $\mathcal{V}_i(\cdot)$, $B_c$, $B_f$, $h_c$, $h_f$, $R_{freq}$, $CFL$}\;
			$\mathcal{D} \leftarrow \mathcal{D}\, \cup\, \mathcal{D}_c$\;
		}
	}   
}
\Return $\mathcal{D}$\;

\caption{\small $\mathcal{D} \leftarrow$ {\tt GenerateDataSet(}$\ell_c^{\max}$, $\ell_f^{\max}$, $t^{end}$, $\nu$, $N_F$, $N_C$, $B_c$, $R_{freq}$, $CFL${\tt )}: Generate learning data set.}
\label{alg:GenerateDataSet}
\end{algorithm}

Following the initialization of $h_c$ and $h_f$, we set the bounds for the interface radius.  With these mesh sizes, we also define $N_r$, which is the number of radii to evaluate for every velocity field.  Here, $N_r$ is equivalent to over three times as many coarse cells fit between $r_{\max}$ and $r_{\min}$.  Then, we enter the outermost data production loop, where we repeatedly generate divergence-free, random velocity fields $\mathcal{V}_i(\cdot)$, for $i = 1, 2, ..., N_F$.  In this study, we have realized $\mathcal{V}_i(\cdot)$ as a smooth field constructed from superpositions of sinusoidal waves \cite{Saad;Sutherland;CommOnRandomVelFields;2016}, as implemented by \cite{Bar-Sinai;Hoyer;Hickey;Brenner;LearningData-DrivenPDEs;2019} and \cite{Zhuang;etal;LrndDiscForPassSclrAdvctn2D;2021}.  Subsequently, we normalize these fields so that $\max{||\mathcal{V}_i(\vv{x})||} = 1$, for all $\vv{x} \in \Omega$.

After defining $\mathcal{V}_i(\cdot)$, we instantiate circular-interface level-set functions of the form

\begin{equation}
\phi_\circ(\vv{x}) = \|\vv{x} - \vv{x}_\circ\|^2 - r_\circ^2
\label{eq:CircularLevelSetFunction}
\end{equation}
with evenly spaced radii $r_\circ$ from $r_{\min}$ to $r_{\max}$.  For each radius, we choose $N_C$ random centers $\vv{x}_\circ = (x_\circ,\, y_\circ)^T$ within the unit square $[-1/2, +1/2]^2 \subset \Omega$.  Then, we use the resulting $\phi_\circ(\vv{x})$ to create the starting coarse and fine grids, $\mathcal{G}_c^0$ and $\mathcal{G}_f^0$, with uniform bands around $\Gamma^0$.  Next, we evaluate $\phi_\circ(\cdot)$ and $\mathcal{V}_i(\cdot)$ on both meshes and conclude with preparatory reinitialization before interleaving advection and collecting samples.  To improve the accuracy on the fine grid, we use $B_c$ times as many redistancing operations on the nodal level-set values $\vv{\phi}_f^0$ as we do on the coarse-grid $\vv{\phi}_c^0$.  In this study, we have selected $\nu = 10$ iterations for solving \cref{eq:Reinitialization}, which is typical for many level-set applications.  For example, Mirzadeh \etal opted for 20 steps in \cite{Mirzadeh;etal:16:Parallel-level-set}, but we have chosen half of it because \cref{eq:CircularLevelSetFunction} is not far from a sign distance function, and we only need accurate results within a small band around $\Gamma$.  In this regard, the practitioner could choose a larger $\nu$ to improve level-set smoothness but should keep in mind that redistancing is costly when one uses high-order schemes \cite{Chene;Min;Gibou:08:Second-order-accurat}.  $\nu$ is thus another parameter limiting the scope of $\mathcal{F}_{c,f}(\cdot)$; the reason is that one must determine $\nu$ ahead of time depending on the accuracy-performance tradeoff.  To conclude, \Cref{alg:GenerateDataSet} launches the {\tt IntAdvectAndCollectSamples()} procedure and accumulates the output samples into the learning data set $\mathcal{D}$.


\begin{algorithm}[!t]
\SetAlgoLined
\SetKwFunction{evaluate}{Evaluate}
\SetKwFunction{semilagrangian}{SemiLagrangian}
\SetKwFunction{advectfinegrid}{AdvectFineGrid}
\SetKwFunction{advectcoarsegrid}{AdvectCoarseGrid}
\SetKwFunction{secondspatialderivatives}{SecondSpatialDerivatives}
\SetKwFunction{collectdatapackets}{CollectDataPackets}
\SetKwFunction{ramp}{Ramp}
\SetKwFunction{fittofinegrid}{FitToFineGrid}
\SetKwFunction{selectivereinit}{SelectiveReinitialization}
\SetKwFunction{computecurvatureandnormal}{ComputeCurvatureAndNormal}
\SetKwFunction{reconstruct}{Reconstruct}
\SetKwFunction{interpolate}{Interpolate}
\SetKwFunction{adjustlswithml}{AdjustLevelSetWithMLSolution}
\SetKwFunction{getnodeswithflow}{GetCoordsWithNegativeFlow}

\KwIn{initial coarse grid and its level-set values and velocities, $\mathcal{G}_c^0$, $\vv{\phi}_c^0$, and $U_c^0$; initial fine grid and its nodal level-set values and velocities, $\mathcal{G}_f^0$, $\vv{\phi}_f^0$, and $U_f^0$; final advection time, $t^{end}$; number of iterations for level-set redistancing, $\nu$; velocity field, $\mathcal{V}(\cdot)$; uniform band half-width around coarse interface, $B_c$; uniform band half-width around fine interface, $B_f$; coarse mesh size, $h_c$; fine mesh size, $h_f$; coarse-grid reset frequency, $R_{freq}$; $CFL$ number.}
\KwResult{coarse-grid learning data set, $\mathcal{D}_c$.}
\BlankLine

$t_c^n \leftarrow 0, \quad \Delta t_c \leftarrow CFL \cdot h_c / \max{||\mathcal{V}(\vv{x})||}, \quad iter \leftarrow 0$\tcp*[r]{Coarse-grid stepping variables}
$(\mathcal{G}_c^n,\, \vv{\phi}_c^n,\, U_c^n) \leftarrow (\mathcal{G}_c^0,\, \vv{\phi}_c^0,\, U_c^0), \quad 
(\mathcal{G}_f^n,\, \vv{\phi}_f^n,\, U_f^n) \leftarrow (\mathcal{G}_f^0,\, \vv{\phi}_f^0,\, U_f^0)$\;
$\mathcal{D}_c \leftarrow \varnothing$\;
\BlankLine
			
\tcp{Interleave advection of $\vv{\phi}_c$ and $\vv{\phi}_f$ using their respective structures in $\mathcal{G}_c$ and $\mathcal{G}_f$}
\While{$t_c^n < t^{end}$}{
	adjust $\Delta t_c$ if $t_c^n + \Delta t_c > t^{end}$\;
	\BlankLine
	
	$(\mathcal{G}_f^{n+1},\, \vv{\phi}_f^{n+1},\, U_f^{n+1}) \leftarrow$ \advectfinegrid{$\mathcal{G}_f^n$, $\vv{\phi}_f^n$, $U_f^n$, $t_c^n$, $t_c^n + \Delta t_c$, $\nu$, $\mathcal{V}(\cdot)$, $B_c$, $B_f$, $h_f$, $CFL$}\tcp*[r]{See \Cref{alg:AdvectFineGrid}}
	\BlankLine
	
	\tcp{Alternating between sampling and standard advection}
	\eIf{$2$ divides $iter$ evenly}{
		
		$(\vv{\kappa}_c^n,\, N_c^n) \leftarrow$ \computecurvatureandnormal{$\mathcal{G}_c^n$, $\vv{\phi}_c^n$}\tcp*[r]{Nodal curvature and normal unit vectors}
		
		$\Phi_{c,xx}^n \leftarrow$ \secondspatialderivatives{$\mathcal{G}_c^n$, $\vv{\phi}_c^n$}\tcp*[r]{$\phi_{xx}$ and $\phi_{yy}$ at the nodes of $\mathcal{G}_c^n$}

		$\mathcal{H}_f^{n+1} \leftarrow$ \reconstruct{$\mathcal{G}_f^{n+1}$}, $\quad\mathcal{H}_c^n \leftarrow$ \reconstruct{$\mathcal{G}_c^n$}\tcp*[r]{Fine and coarse tree hierarchies}
		\BlankLine
	
		\tcp{Collect and process data packets to create learning samples}	
		$(\mathcal{P},\, \mathcal{C}) \leftarrow$ \collectdatapackets{$\mathcal{G}_c^n$, $\mathcal{H}_c^n$, $\vv{\phi}_c^n$, $U_c^n$, $\Phi_{c,xx}^n$, $N_c^n$, $\vv{\kappa}_c^n$, $\Delta t_c$}\tcp*[r]{See \Cref{alg:CollectDataPackets}}
		$\mathcal{M}_\phi \leftarrow \varnothing$\tcp*[r]{Map from node coordinates to $\phi$ values interpolated from $\vv{\phi}_f^{n+1}$}
	
		\ForEach{data packet $\mathcal{p} \in \mathcal{P}$}{
		
			let $\vv{x}_a$ be the coordinates (i.e., arrival point) corresponding to $\mathcal{p}$ in $\mathcal{C}$\;
		
			$\tilde{\phi}_d^* \leftarrow$ \interpolate{$\mathcal{H}_f^{n+1}$, $\vv{\phi}_f^{n+1}$, $\vv{x}_a$}$/ h_c$\tcp*[r]{Use advected fine grid to get normalized $\phi_d^*$}
			
			let $\vv{\xi}$ be the learning tuple $\left(\mathcal{p}, \tilde{\phi}_d^*\right)$ with inputs $\mathcal{p}$ and expected output $\tilde{\phi}_d^*$\;
			transform $\vv{\xi}$, so that $\mathcal{p}.\kappa_a$ is negative\;
			rotate $\vv{\xi}$, so that the angle of $-\mathcal{p}.\hat{\vv{u}}_a$ lies between $0$ and $\pi/2$\;
			add tuple $\vv{\xi}$ to $\mathcal{D}_c$\;
			\BlankLine
		
			let $\vv{\xi}'$ be the reflected tuple about line $y = x + \beta$ going through $\vv{x}_a$\tcp*[r]{Data augmentation}
			add $\vv{\xi}'$ to $\mathcal{D}_c$\;
			\BlankLine
		
			$\mathcal{M}_\phi[\,\vv{x}_a\,] \leftarrow h_c\tilde{\phi}_d^*$\tcp*[r]{Caching expected or improved level-set values}
		}
		\BlankLine
	
		\eIf{$R_{freq}$ divides $(iter + 1)$ evenly}{
			$(\mathcal{G}_c^{n+1},\, \vv{\phi}_c^{n+1}) \leftarrow$ \fittofinegrid{$\mathcal{G}_c^n$, $\vv{\phi}_c^n$, $\mathcal{G}_f^{n+1}$, $\vv{\phi}_f^{n+1}$}\tcp*[r]{Reset coarse grid}
		}{
			$(\mathcal{G}_c^{n+1},\, \vv{\phi}_c^{n+1}) \leftarrow$ \semilagrangian{$\mathcal{G}_c^n$, $\vv{\phi}_c^n$, $U_c^n$, $U_c^n$, $CFL$}\tcp*[r]{Refer to Algorithm \href{https://www.sciencedirect.com/science/article/pii/S002199911630242X\#fg0060}{3} in \cite{Mirzadeh;etal:16:Parallel-level-set}}
			\adjustlswithml{$\vv{\phi}_c^{n+1}$, $\mathcal{M}_\phi$}\tcp*[r]{Correct $\phi^{n+1}$ for nodes with coords in $\mathcal{M}_\phi$}
		}
		\BlankLine
	
		$\mathcal{W} \leftarrow$ \getnodeswithflow{$\mathcal{H}_c^n$, $U_c^n$, $N_c^n$, $\mathcal{G}_c^{n+1}$, $\vv{\phi}_c^{n+1}$}\tcp*[r]{Points `lagging behind' $\Gamma_c^{n+1}$}
		$\vv{\phi}_c^{n+1} \leftarrow$ \selectivereinit{$\mathcal{G}_c^{n+1}$, $\vv{\phi}_c^{n+1}$, $\mathcal{C}\, \cap\, \mathcal{W}$, $\nu$}\;
	}{
		$(\mathcal{G}_c^{n+1},\, \vv{\phi}_c^{n+1}) \leftarrow$ \advectcoarsegrid{$\mathcal{G}_c^n$, $\vv{\phi}_c^n$, $U_c^n$, $\mathcal{G}_f^{n+1}$, $\vv{\phi}_f^{n+1}$, $\nu$, $iter$, $R_{freq}$, $CFL$}\tcp*[r]{See \Cref{alg:AdvectCoarseGrid}}
	}
	\BlankLine
	
	$U_c^{n+1} \leftarrow$ \evaluate{$\mathcal{V}(\cdot)$, $\mathcal{G}_c^{n+1}$}\tcp*[r]{Resample velocity field}
	$iter \leftarrow iter + 1, \quad t_c^n \leftarrow t_c^n + \Delta t_c$\tcp*[r]{and update stepping variables}
	$(\mathcal{G}_c^n,\, \vv{\phi}_c^n,\, U_c^n) \leftarrow (\mathcal{G}_c^{n+1},\, \vv{\phi}_c^{n+1},\, U_c^{n+1}), \quad 
(\mathcal{G}_f^n,\, \vv{\phi}_f^n,\, U_f^n) \leftarrow (\mathcal{G}_f^{n+1},\, \vv{\phi}_f^{n+1},\, U_f^{n+1})$\;
}
\Return $\mathcal{D}_c$\;

\caption{\small $\mathcal{D}_c \leftarrow$ {\tt IntAdvectAndCollectSamples(}$\mathcal{G}_c^0$, $\mathcal{G}_f^0$, $\vv{\phi}_c^0$, $\vv{\phi}_f^0$, $U_c^0$, $U_f^0$, $t^{end}$, $\nu$, $\mathcal{V}(\cdot)$, $B_c$, $B_f$, $h_c$, $h_f$, $R_{freq}$, $CFL${\tt )}: Interleave advection of coarse and fine grids and collect learning samples.}
\label{alg:IntAdvectAndCollectSamples}
\end{algorithm}

We present the coarse- and fine-grid simultaneous advection subroutine to assemble $\mathcal{D}$ in \Cref{alg:IntAdvectAndCollectSamples}.  Our method follows the natural flow of a DNS on $\mathcal{G}_c$ with interleaved updates to the level-set function on $\mathcal{G}_f$.  In addition, it integrates the alternation depicted in \cref{fig:Overview} and the periodic resetting operations necessary for trajectory synchronization.  The formal parameters of \Cref{alg:IntAdvectAndCollectSamples} include the initial-condition level-set vectors and velocity matrices.  Also, it requires the divergence-free velocity function $\mathcal{V}(\cdot)$ and the coarse- and fine-grid data structures.  As before, our constraints help define the coarse-grid time step as $\Delta t_c = h_c$ (and $\Delta t_f = h_f$) since $\max{||\mathcal{V}(\vv{x})|| = 1}$ for all points in $\Omega$.

In each iteration, the main simulation loop in \Cref{alg:IntAdvectAndCollectSamples} first advances the fine-grid level-set function from $\vv{\phi}_f^n \equiv \vv{\phi}_f(t_c^n)$ to $\vv{\phi}_f^{n+1} \equiv \vv{\phi}_f(t_c^{n+1}) \equiv \vv{\phi}_f(t_c^n + \Delta t_c)$.  That way, we know which reference level-set values are expected for arrival points next to the interface in $\mathcal{G}_c^n$ during sampling.  The {\tt AdvectFineGrid()} subroutine retrieves the updated fine mesh and its nodal level-set values and velocities.  \Cref{alg:AdvectFineGrid} describes the actions implied by this module.  Briefly, \Cref{alg:AdvectFineGrid} is an iterative process based on the {\tt SemiLagrangian()} scheme of \cite{Mirzadeh;etal:16:Parallel-level-set}.  It transports $\vv{\phi}_f^n$, reinitializes the new level-set values, and reevaluates $\mathcal{V}(\cdot)$ on the updated grid at regular time intervals.  About redistancing on $\mathcal{G}_f$, it is worth emphasizing that we need $B_f \cdot \nu$ iterations in the last fine-grid advection step.  The motivation behind this idea is to ensure that $\vv{\phi}_f^{n+1}$ accurately resembles a signed distance function as far as $B_c$ coarse cells away from $\Gamma_f$.


\begin{algorithm}[!t]
\SetAlgoLined
\SetKwFunction{evaluate}{Evaluate}
\SetKwFunction{reinitialize}{Reinitialize}
\SetKwFunction{semilagrangian}{SemiLagrangian}

\KwIn{fine grid and its nodal level-set values and velocities, $\mathcal{G}_f^0$, $\vv{\phi}_f^0$, and $U_f^0$; initial advection time, $t_f^0$; ending advection time, $t_f^F$; number of iterations for level-set redistancing, $\nu$; velocity field, $\mathcal{V}(\cdot)$; uniform band half-width around coarse interface, $B_c$; uniform band half-width around fine interface, $B_f$; mesh size, $h_f$; $CFL$ number.}
\KwResult{updated fine grid structure, $\mathcal{G}_f^F$, nodal level-set values, $\vv{\phi}_f^F$, and nodal velocities, $U_f^F$, at time $t_f^F$.}
\BlankLine

$t_f^n \leftarrow t_f^0, \quad \Delta t_f \leftarrow CFL \cdot h_f / \max{||\mathcal{V}(\vv{x})||}$\tcp*[r]{Stepping variables}
$(\mathcal{G}_f^n,\, \vv{\phi}_f^n,\, U_f^n) \leftarrow (\mathcal{G}_f^0,\, \vv{\phi}_f^0,\, U_f^0)$\;
\BlankLine
				
\While{$t_f^n < T$}{
	adjust $\Delta t_f$ if $t_f^n + \Delta t_f > t_f^F$\;
				
	$(\mathcal{G}_f^{n+1},\, \vv{\phi}_f^{n+1}) \leftarrow$ \semilagrangian{$\mathcal{G}_f^n$, $\vv{\phi}_f^n$, $U_f^n$, $U_f^n$, $CFL$}\tcp*[r]{Refer to Algorithm \href{https://www.sciencedirect.com/science/article/pii/S002199911630242X\#fg0060}{3} in \cite{Mirzadeh;etal:16:Parallel-level-set}}
	\eIf{$t_f^n + \Delta t_f < t_f^F$}{
		$\vv{\phi}_f^{n+1} \leftarrow$ \reinitialize{$\mathcal{G}_f^{n+1}$, $\vv{\phi}_f^{n+1}$, $B_c \cdot \nu$}\;
	}{
		$\vv{\phi}_f^{n+1} \leftarrow$ \reinitialize{$\mathcal{G}_f^{n+1}$, $\vv{\phi}_f^{n+1}$, $B_f \cdot \nu$}\tcp*[r]{Increase redistancing accuracy in last step}
	}
	$U_f^{n+1} \leftarrow$ \evaluate{$\mathcal{V}(\cdot)$, $\mathcal{G}_f^{n+1}$}\tcp*[r]{Resample velocity field}
	\BlankLine
	
	$t_f^n \leftarrow t_f^n + \Delta t_f$\;
	$(\mathcal{G}_f^n,\, \vv{\phi}_f^n,\, U_f^n) \leftarrow (\mathcal{G}_f^{n+1},\, \vv{\phi}_f^{n+1},\, U_f^{n+1})$\;
}
$(\mathcal{G}_f^F,\,  \vv{\phi}_f^F,\, U_f^F) \leftarrow (\mathcal{G}_f^n,\, \vv{\phi}_f^n,\, U_f^n)$\;
\Return $(\mathcal{G}_f^F,\, \vv{\phi}_f^F,\, U_f^F)$\;

\caption{\small $(\mathcal{G}_f^F,\, \vv{\phi}_f^F,\, U_f^F) \leftarrow$ {\tt AdvectFineGrid(}$\mathcal{G}_f^0$, $\vv{\phi}_f^0$, $U_f^0$, $t_f^0$, $t_f^F$, $\nu$, $\mathcal{V}(\cdot)$, $B_c$, $B_f$, $h_f$, $CFL${\tt )}: Advect fine-grid level-set values from $\vv{\phi}_f^0$ to $\vv{\phi}_f^F$ and update nodal velocities.}
\label{alg:AdvectFineGrid}
\end{algorithm}

After advecting $\vv{\phi}_f^n$ to $\vv{\phi}_f^{n+1}$, we decide whether one should collect samples for the current iteration.  If not, we simply call the {\tt AdvectCoarseGrid()} module (see \Cref{alg:AdvectCoarseGrid}) and retrieve the updated grid and (reinitialized) nodal level-set values into $\mathcal{G}_c^{n+1}$ and $\vv{\phi}_c^{n+1}$.  The {\tt AdvectCoarseGrid()} subroutine is equivalent to a single step in \Cref{alg:AdvectFineGrid} performed on $\mathcal{G}_c$ with an additional reset operation.  We will provide more details and the justification for the {\tt FitToFineGrid()} reset statement below as we dive into data sampling.

In the case of a data-collection iteration, we start by computing coarse-grid nodal curvature values, normal unit vectors, and level-set second-order derivatives.  Then, we fetch training input data through the {\tt CollectDataPackets()} procedure listed in \Cref{alg:CollectDataPackets}.  The sampling portion in {\tt IntAdvectAndCollectSamples()} replicates the first group of statements in our {\tt MLSemiLagrangian()} algorithm.  However, instead of constructing a batch of preprocessed entries, we accumulate learning tuples using the data packets in $\mathcal{P}$ as inputs.  We denote a learning tuple by

\begin{equation}
\vv{\xi} \doteq \left( \mathcal{p}, \tilde{\phi}_d^* \right ) \in \mathbb{R}^{23},
\label{eq:LearningTuple}
\end{equation}
where $\mathcal{p}$ appears in \cref{eq:DataPacket}, and $\tilde{\phi}_d^*$ is the $h_c$-normalized reference level-set value at the departure point.  To calculate $\tilde{\phi}_d^*$, we turn to quadratic interpolation from $\vv{\phi}_f^{n+1}$.  This strategy leverages $\vv{\phi}_f^{n+1}$'s higher accuracy and the fact that the coarse-grid vertices around $\Gamma_c^n$ are a proper subset of the fine-mesh uniform band around $\Gamma_f^{n+1} \equiv \Gamma_f(t^n + \Delta t_c)$.  Similar to \Cref{alg:MLSemiLagrangian}, we reorient $\mathcal{p} \in \vv{\xi}$ into its standard form and apply negative-curvature normalization.  At the same time, we store $\phi_d^*$ for next-to-$\Gamma_c^n$ nodes in a cache, $\mathcal{M}_\phi$. Later, we will merge these target values with the rest of the elements in $\vv{\phi}_c^{n+1}$ advected numerically.

In addition to accumulating reoriented-input samples, as in \cref{fig:ReorientedSample}, we exploit symmetry and invariance in data-packet fields to supplement $\mathcal{D}$ with augmented pairs.  Data augmentation via rotation and reflection has proven effective for image classifiers with convolutional networks \cite{A18, Hands-onMLwithScikit-LearnKerasAndTF19}.  Here, we augment samples by reflecting $\mathcal{p} \in \vv{\xi}$ about the line $y = x + \beta$ going through the origin of the local coordinate system centered at $\mathcal{p}$'s arrival point (see \cref{fig:AugmentedSample}).

Next, we advance the coarse-grid level-set values after collecting learning samples for time $t^n$ into the cumulative set $\mathcal{D}_c$.  To evolve $\vv{\phi}_c^n$ to $\vv{\phi}_c^{n+1}$ and update $\mathcal{G}_c^n$, we could have only called the {\tt SemiLagrangian()} procedure and merged the reference $\phi_d^*$ values from $\mathcal{M}_\phi$ back into $\vv{\phi}_c^{n+1}$.  However, the discretized coarse- and fine-mesh level-set functions can diverge to the point where our numerical-error characterization with $\mathcal{F}_{c,f}(\cdot)$ becomes invalid.  Therefore, we reset $\vv{\phi}_c^{n+1}$ periodically through the {\tt FitToFineGrid()} method by interpolating its level-set values from $\vv{\phi}_f^{n+1}$ every $R_{freq} = 3$ iterations.  This recurrent synchronization protects against large trajectory deviations while preserving the natural noise that emerges from $\vv{\phi}_c^n$'s conventional advection.


\begin{algorithm}[!t]
\SetAlgoLined
\SetKwFunction{evaluate}{Evaluate}
\SetKwFunction{reinitialize}{Reinitialize}
\SetKwFunction{semilagrangian}{SemiLagrangian}
\SetKwFunction{fittofinegrid}{FitToFineGrid}

\KwIn{coarse grid and its nodal level-set values and velocities, $\mathcal{G}_c^n$, $\vv{\phi}_c^n$, and $U_c^n$; updated fine grid and its nodal level-set values, $\mathcal{G}_f^{n+1}$ and $\vv{\phi}_c^{n+1}$; number of iterations for level-set redistancing, $\nu$; current simulation iteration, $iter$; coarse-grid reset frequency, $R_{freq}$; $CFL$ number.}
\KwResult{updated coarse grid structure, $\mathcal{G}_c^{n+1}$, and reinitialized nodal level-set values, $\vv{\phi}_c^{n+1}$.}
\BlankLine

\eIf{$R_{freq}$ divides $(iter + 1)$ evenly}{
	$(\mathcal{G}_c^{n+1},\, \vv{\phi}_c^{n+1}) \leftarrow$ \fittofinegrid{$\mathcal{G}_c^n$, $\vv{\phi}_c^n$, $\mathcal{G}_f^{n+1}$, $\vv{\phi}_f^{n+1}$}\tcp*[r]{Reset coarse grid}
}{
	$(\mathcal{G}_c^{n+1},\, \vv{\phi}_c^{n+1}) \leftarrow$ \semilagrangian{$\mathcal{G}_c^n$, $\vv{\phi}_c^n$, $U_c^n$, $U_c^n$, $CFL$}\tcp*[r]{Refer to Algorithm \href{https://www.sciencedirect.com/science/article/pii/S002199911630242X\#fg0060}{3} in \cite{Mirzadeh;etal:16:Parallel-level-set}}
}
\BlankLine
	
$\vv{\phi}_c^{n+1} \leftarrow$ \reinitialize{$\mathcal{G}_c^{n+1}$, $\vv{\phi}_c^{n+1}$, $\nu$}\;

\Return $(\mathcal{G}_c^{n+1},\, \vv{\phi}_c^{n+1})$\;

\caption{\small $(\mathcal{G}_c^{n+1},\, \vv{\phi}_c^{n+1}) \leftarrow$ {\tt AdvectCoarseGrid(}$\mathcal{G}_c^n$, $\vv{\phi}_c^n$, $U_c^n$, $\mathcal{G}_f^{n+1}$, $\vv{\phi}_f^{n+1}$, $\nu$, $iter$, $R_{freq}$, $CFL${\tt )}: Advect coarse-grid level-set values from $\vv{\phi}_c^n$ to $\vv{\phi}_c^{n+1}$ and perform reinitialization.}
\label{alg:AdvectCoarseGrid}
\end{algorithm}

The last steps in \Cref{alg:IntAdvectAndCollectSamples}'s sampling iteration involve reinitializing $\vv{\phi}_c^{n+1}$.  Such a task is subtle and requires special care in our machine learning strategy.  In particular, we should solve \cref{eq:Reinitialization} \textit{only} for values in $\vv{\phi}_c^{n+1}$ that we did not update via interpolation from $\vv{\phi}_f^{n+1}$ and for which the condition in \cref{eq:WithTheFlowCondition} is not satisfied.  Thus, one must use selective reinitialization to guard the machine-learning-corrected values with the help of the auxiliary arrays $\mathcal{C}$ and $\mathcal{W}$.  As shown in \Cref{sec:Results}, the {\tt SelectiveReinitialization()} and {\tt MLSemiLagrangian()} subroutines work together with alternating transport to ensure that the moving front remains smooth and as close as possible to its higher-resolution counterpart.

The purpose of \Cref{alg:IntAdvectAndCollectSamples} is to accumulate learning samples in a local set until we reach the final simulation time.  Then, upon exiting {\tt IntAdvectAndCollectSamples()}, we return to the {\tt GenerateDataSet()} function, where $\mathcal{D}$ appends the partial results arriving from every DNS configuration.  By the end of \Cref{alg:GenerateDataSet}, $\mathcal{D}$ contains all the tuples in the form given by \cref{eq:LearningTuple}.  In the following tasks, we employ this data set to train and assess $\mathcal{F}_{c,f}(\cdot)$.


\colorsubsubsection{Technical aspects}
\label{subsubsec:TechnicalAspects}

Training an error-correcting neural network is much more resource-intensive than constructing its learning data set.  Even though assembling $\mathcal{D}$ is algorithmically complex and can take several hours, one must perform this task just once at the beginning of the learning stage.  On the contrary, optimizing $\mathcal{F}_{c,f}(\cdot)$ is less cumbersome but requires exploring the hyperparameter space through multiple evaluations.  Every trial can last many hours depending on $|\mathcal{D}|$ and the size of the hidden layers in $\mathcal{F}_{c,f}(\cdot)$.  

In this work, we have realized \Crefrange{alg:MLSemiLagrangian}{alg:AdvectCoarseGrid} within our C++ implementation of the parallel level-set framework of \cite{Mirzadeh;etal:16:Parallel-level-set}.  As for learning, we have simplified most of the tasks with TensorFlow \cite{Tensorflow15} and Keras \cite{Keras15} in Python.  In addition, we have followed conventional practices \cite{A18, Mehta19} and split $\mathcal{D}$ into non-overlapping training, testing, and validation subsets to optimize $\mathcal{F}_{c,f}(\cdot)$ and promote generalization.  Our partitioning approach also ensures that each subset mirrors the output $\tilde{\phi}_d^*$ distribution in $\mathcal{D}$.  The specific heuristics that make this possible rely on the {\tt StratifiedKFold()} method from Scikit-learn \cite{scikit-learn11}.  Although scientists usually reserve this balancing subroutine for classification problems, we have introduced it into our workflow by employing Pandas' {\tt cut()} procedure \cite{Pandas2010}.  The {\tt cut()} function first allows us to bin the reference level-set values into discrete intervals.  Then, we call {\tt StratifiedKFold()} three times with $K = 10$ to populate the training, testing, and validation subsets with $70\%$, $10\%$, and $10\%$ of the tuples in $\mathcal{D}$.  The remaining $10\%$ is discarded as a way of performing balanced subsampling. However, we do not carry out the expensive full-fledged 10-fold cross-validation during hyperparameter tuning.  We have observed that a single fold with a couple of random initializations is sufficient, given that $\mathcal{D}$ is large.

The other critical component in our machine-learning-extended semi-Lagrangian scheme is the {\tt Preprocess()} module portrayed in \cref{fig:Overview,fig:ECNet}.  The {\tt Preprocess()} actions depend on training-subset statistics and help transform raw feature vectors into amenable representations for the numerical-error estimator \cite{scikit-learn11}.  In our hyperparameter-space exploration, we have discovered the existence of three types of preprocessing operations that favor convergence, interpretability, and precision.  \Cref{alg:Preprocess} lists these adjustments performed on data-packet attributes before assembling the input vectors in \cref{eq:Samples}.  First, we $h$-scale the distance-related data in $\mathcal{p}$, such as the departure-arrival-point distance, level-set values, level-set second-order derivatives, and curvature.  The purpose of $h$-scaling is (1) to reduce the relative difference among feature magnitudes in $\mathcal{p}$ and (2) to uncorrelate data from a particular grid resolution.  While (1) facilitates neural learning, (2) is useful for bootstrapping hyperparameter exploration across grid discretizations.  In other words, uncorrelating data-packet attributes from the mesh size can reduce the cost of neural model optimization because the hyperparameters that work for $h_c$ are a good starting point for fitting $\mathcal{F}_{c,f}(\cdot)$ to learning tuples from $\mathcal{G}'$, where $h_c' \neq h_c$.

After $h$-scaling distance data, we use the feature-type-based statistics $\mathcal{Q}$ extracted from the training subset to perform standardization.  The latter comprises centering and scaling the entries in $\mathcal{p}$ using the mean and standard deviation calculated \textit{across} features of the same kind.  For example, the {\tt Standardize()} method in \Cref{alg:Preprocess} replaces $\varphi \in \Psi \subset \mathcal{p}$ for $(\varphi - \mathcal{Q}.\mu_\phi) / \mathcal{Q}.\sigma_\phi$, where $\mathcal{Q}.\mu_\phi$ and $\mathcal{Q}.\sigma_\phi$ are the level-set mean and standard deviation computed for $\Psi \doteq \{\phi_a,\, (\phi_d)_{ij},\, \phi_d\}$, where $i,j \in \{0, 1\}$.


\begin{algorithm}[!t]
\SetAlgoLined
\SetKwFunction{getstats}{GetStats}
\SetKwFunction{standardize}{Standardize}
\SetKwFunction{pca}{PCA}

\KwIn{data packet to preprocess, $\mathcal{p}$; mesh size, $h$.}
\KwResult{preprocessed data packet as a vector $\vv{p}$.}
\BlankLine

\tcp{In-place $h$-scaling groups of variables in $\mathcal{p}$}
Divide $\mathcal{p}.\phi_a$, $\mathcal{p}.(\phi_d)_{ij}$, and $\mathcal{p}.\phi_d$ by $h$\tcp*[r]{Take $i,j \in \{0, 1\}$}
Multiply $\mathcal{p}.\phi_{xx}$ and $\mathcal{p}.\phi_{yy}$ by $h^2$\;
Multiply $\mathcal{p}.\kappa_a$ by $h$\;
Divide $\mathcal{p}.d$ by $h$\;
\BlankLine

\tcp{In-place (quasi) standardization}
$\mathcal{Q} \leftarrow$ \getstats{}\tcp*[r]{Retrieve training stats}
\standardize{$[\mathcal{p}.\phi_a$, $\mathcal{p}.(\phi_d)_{ij}$, $\mathcal{p}.\phi_d]$, $\mathcal{Q}.\mu_\phi$, $\mathcal{Q}.\sigma_\phi$}\;
\standardize{$[\mathcal{p}.\hat{\vv{u}}_a$, $\mathcal{p}.(\vv{u}_d)_{ij}]$, $\mathcal{Q}.\mu_{\vv{u}}$, $\mathcal{Q}.\sigma_{\vv{u}}$}\;
\standardize{$[\mathcal{p}.d]$, $\mathcal{Q}.\mu_d$, $\mathcal{Q}.\sigma_d$}\;
\standardize{$[\mathcal{p}.\tilde{x}_d$, $\mathcal{p}.\tilde{y}_d]$, $\mathcal{Q}.\mu_{coords}$, $\mathcal{Q}.\sigma_{coords}$}\;
\standardize{$[\mathcal{p}.(\phi_{xx})_d$, $\mathcal{p}.(\phi_{yy})_d]$, $\mathcal{Q}.\mu_{xxyy}$, $\mathcal{Q}.\sigma_{xxyy}$}\;
\standardize{$[\mathcal{p}.\kappa_a]$, $\mathcal{Q}.\mu_\kappa$, $\mathcal{Q}.\sigma_\kappa$}\;
\BlankLine

\tcp{Dimensionality reduction and whitening}
$\vv{p} \leftarrow$ \pca{$\mathcal{p}$, $\mathcal{Q}.N_{comp}$}\tcp*[r]{PCA and whitening with $N_{comp}$ components}
\BlankLine

\Return $\vv{p}$\;

\caption{\small $\vv{p} \leftarrow$ {\tt Preprocess(}$\mathcal{p}$, $h${\tt )}: Preprocess a data packet $\mathcal{p}$ using training statistics.}
\label{alg:Preprocess}
\end{algorithm}

In the last group of preprocessing operations, we use principal component analysis (PCA) and whitening alongside dimensionality reduction.  PCA and whitening are well known for accelerating convergence, and practitioners have shown that their combination leads to better results than z-scoring \cite{LeCun;EfficientBackProp;98, Larios;Gibou;HybridCurvature;2021}.  Coupling standardization and PCA is also beneficial because it prevents variance from directly reflecting the scale of the data \cite{Parker;CS170A;2016}.  As for whitening, we have incorporated it because it adds a secondary component-wise scaling that raises uncorrelated concepts to the same level of importance (on an a priori basis) \cite{A18}.  In our case, we employ Scikit-learn's {\tt PCA} class to transform standardized data packets into input vectors with $\mathcal{Q}.N_{comp} = 17$ entries.  After settling down with an optimal model, we export $\mathcal{F}_{c,f}(\cdot)$, $\mathcal{Q}$, and the {\tt PCA} object to {\tt JSON} files.  Then, we can use these files for porting the entire inference system in \cref{fig:ECNet} to different platforms.

We close this section by providing insight into training $\mathcal{F}_{c,f}(\cdot)$ with TensorFlow and Keras.  As seen in \cref{fig:ECNet}, our multilayer perceptron has $17 + 1$ linear input neurons that ingest data from the {\tt Preprocess()} module and $\tilde{\phi}_d$.  The first four hidden layers contain $N_h^i$ nonlinear, ReLU units, for $i = 1, 2, 3, 4$, with trainable connections and parameters.  The last hidden-layer neuron is linear and updatable, and the output unit merely adds its input tensors while requiring no trainable weights.  To train $\mathcal{F}_{c,f}(\cdot)$, we have used backpropagation \cite{DeepLearning;Goodfellow-et-al;2016, A18} with the Adam optimizer \cite{Adam;2015} to minimize the root mean squared error loss (RMSE) between $\tilde{\phi}_d^*$ and $\bar{\varepsilon} + \tilde{\phi}_d$.  Likewise, we have introduced kernel regularization in all hidden layers with an {\tt L2} factor of $10^{-6}$.  As for the weight updates, these take place after evaluating the RMSE over batches with 64 samples.  In addition, we monitor the mean absolute error (MAE) on the validation subset to reduce the learning rate systematically and prevent overfitting.  In particular, we stabilize optimization by attaching a callback function that halves the learning rate from $\eten{1.5}{-4}$ to $\eten{1.5}{-5}$ whenever the validation MAE does not improve for fifteen epochs.  Also, we stop backpropagation as a protective measure if the latter MAE continues to deteriorate for fifty iterations.  In our preliminary experiments, we have observed that a maximum of one thousand iterations suffices for training $\mathcal{F}_{c,f}(\cdot)$.  Thanks to Keras' facilities and callback methods, we can easily handle these and other technical aspects, like best-model checkpoints and logging.  The upcoming section asserts this complete process by producing an error-correcting neural network for a pair $(c,f)$ of grid resolutions.


\colorsection{Results}
\label{sec:Results}

We now assess the feasibility of the ideas presented in \Cref{sec:Methodology}.  Our proof of concept comprises a neural model $\mathcal{F}_{6,8}(\cdot)$ designed for a coarse grid composed of unit-square quadtrees with $\ell_c^{\max} = 6$.  As for the fine grid that supplies the reference level-set values $\phi_d^*$, we require $\ell_f^{\max} = 8$ and $h_f = 2^{-8}$.  Consequently, our error-correcting neural network, the {\tt MLSemiLagrangian()} subroutine, selective reinitialization, and the {\tt SemiLagrangian()} scheme must operate in tandem to improve interface advection with $h_c = 2^{-6} = 0.015625$.

To optimize $\mathcal{F}_{6,8}(\cdot)$, we gathered 4'051,946 samples from the methodologies outlined in \Crefrange{alg:CollectDataPackets}{alg:AdvectCoarseGrid}.  Then, we used stratified $K$-fold indexing (see \Cref{subsubsec:TechnicalAspects}) to discard about $10\%$ of them and allocate 2'836,283, 405,203, and 405,213 tuples for the training, testing, and validation subsets.  Hence, our curated learning set $\mathcal{D}'$ ended up with 3'646,699 samples.  Exploring the hyperparameter space with $\mathcal{D}'$ resulted in an optimal architecture with $N_h^i = 130$ units in the first four hidden layers.  The topology in \cref{fig:ECNet} thus gave rise to 53,561 parameters that we trained during 337 epochs.  At this stage, we also discovered that reducing input vectors from 22 to 17 dimensions was critical for learning stability and accuracy.  While fewer than 17 components in the {\tt PCA} object led to underfitting, more than these did not translate into any significant improvements.  Likewise, employing over 17 dimensions precipitated outliers' appearance because of the additional whitening normalization.  This problem is not new, and we have reported it when computing curvature in \cite{Larios;Gibou;HybridCurvature;2021}.

\begin{figure}[!t]
	\centering
	\begin{subfigure}[b]{0.4\textwidth}
		\includegraphics[width=\textwidth]{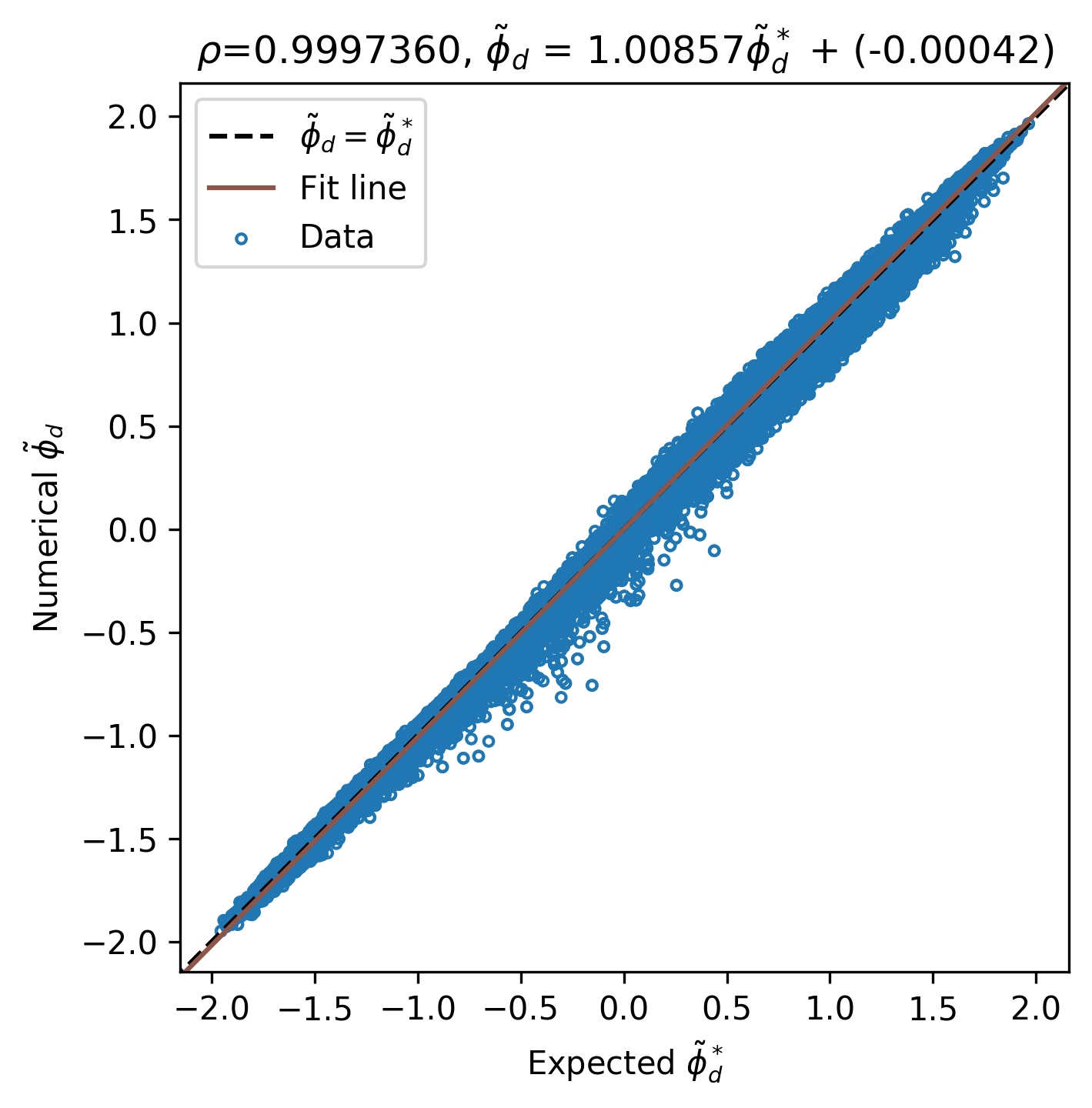}
        \caption{\footnotesize Numerical semi-Lagrangian scheme}
        \label{fig:results.training.numerics}
    \end{subfigure}
    ~
	\begin{subfigure}[b]{0.4\textwidth}
		\includegraphics[width=\textwidth]{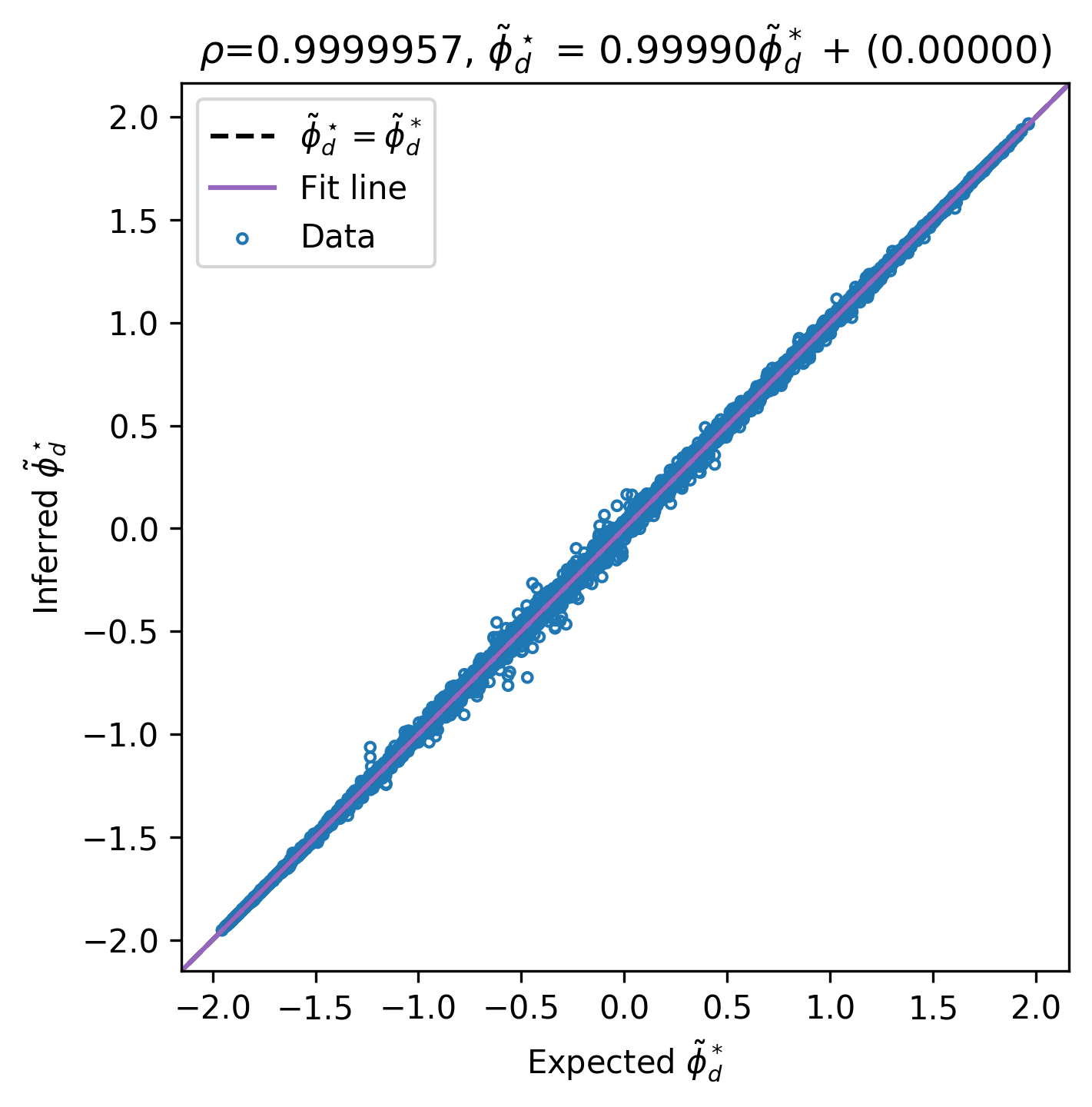}
        \caption{\footnotesize Neural network}
        \label{fig:results.training.nnet}
    \end{subfigure}
    
	\caption{Learning correlation plots over $\mathcal{D}'$ between expected and approximated $h_c$-normalized level-set values at the departure points.  (Color online.)}
	\label{fig:results.training}
\end{figure}

\Cref{fig:results.training} contrasts the fitting quality of the numerical baseline and $\mathcal{F}_{6,8}(\cdot)$ over the entire learning data set.  These results confirm that the error-correcting function aids in narrowing the gap between the estimated and the reference level-set values during semi-Lagrangian advection.  \Cref{tbl:results.training.stats} supports this observation, where MaxAE corresponds to the maximum absolute error.  Our neural network helps reduce the MAE by $91\%$ and lowers the maximum error by a factor of $2.4$ in $\mathcal{D}'$.  These figures corroborate that the conventional scheme in \crefrange{eq:SLLevelSetValue}{eq:SLIntermediateVelocity} can benefit from data-driven corrections even though we cannot represent $\bar{\varepsilon}$ exactly.  

The following tests verify whether $\mathcal{F}_{6,8}(\cdot)$ can offset numerical dissipation and decrease artificial mass loss in physical simulations.  These experiments include rotating and deforming a disk, measuring the effect of tangential shear flows, and solving two instances of the Stefan problem.  For compatibility with $\mathcal{F}_{6,8}(\cdot)$, we have discretized $\Omega$ using quadtree Cartesian grids with unit-square macrocells and $\ell_c^{\max} = 6$.  Also, we ensure the simulations meet the conditions in \cref{fig:Sample} by enforcing $CFL = 1$ and scaling velocity fields to fit within the unit ball (i.e., $\Delta t = h_c$) whenever possible.  For consistency, we reinitialize $\phi(\vv{x})$ every time step using ten iterations, as noted in \Cref{subsubsec:DataSetGeneration}.  In all cases, we measure the level-set error norms defined by

\begin{equation}
\ell^1 = \frac{1}{\mathcal{N}}\sum_{\mathcal{n} \in \mathcal{N}}\left|\phi(\mathcal{n}.\vv{x}) - \phi^*(\mathcal{n}.\vv{x})\right| \quad
\textrm{and} \quad 
\ell^\infty = \max_{\mathcal{n} \in \mathcal{N}}\left|\phi(\mathcal{n}.\vv{x}) - \phi^*(\mathcal{n}.\vv{x})\right|,
\label{eq:errornorms}
\end{equation}
where $\mathcal{N}$ is the set of nodes adjacent to $\Gamma$ (i.e., $|\phi(\vv{x})| \leqslant \sqrt{2}h_c$), and $\phi^*(\cdot)$ is the expected value.  These collective statistics over a queried solution state are analogous to the MAE and MaxAE for $\phi_d$ introduced in the sections above.

\begin{table}[!t]
	\centering
	\small
	\bgroup
	\def\arraystretch{1.1}%
	\begin{tabular}{|l|c|c|c|}
		\hline
		~ & MAE & MaxAE & RMSE \\
		\hline \hline
		Neural network                   & $\eten{7.990813}{-4}$ & $\eten{2.535921}{-1}$ & $\eten{1.744662}{-3}$ \\ \hline
		Numerical semi-Lagrangian scheme & $\eten{9.113001}{-3}$ & $\eten{6.033318}{-1}$ & $\eten{1.471008}{-2}$ \\
		\hline
	\end{tabular}
	\egroup
	\caption{Learning output statistics over $\mathcal{D}'$ for $h_c$-normalized level-set values at the departure points.}
	\label{tbl:results.training.stats}
\end{table}

We close this preamble with a remark about the hybrid inference system’s technical implementation.  In the following, we have turned to Lohmann's {\tt json} suite \cite{json;2021} and some utilities from Hermann's {\tt frugally-deep} library \cite{frugally-deep;2021} to import $\mathcal{F}_{6,8}(\cdot)$ and the {\tt Preprocess()} dependencies into C++.  {\tt frugally-deep} provides a user-friendly interface for neural forward evaluations; however, we have reconstructed the internal mechanisms in $\mathcal{F}_{6,8}(\cdot)$ as a collection of weight matrices in OpenBLAS \cite{openblas;2021}.  Compared with {\tt frugally-deep}, OpenBLAS offers up to sixty times faster inference (with compiler optimization enabled\footnote{We used the flags {\tt -O2 -O3 -march=native} for C++14.}) but requires arranging inputs in a specific form.  To accommodate OpenBLAS constraints, we clustered data into batches first.  Then, we performed predictions in \Cref{alg:MLSemiLagrangian} (\cref{alg:MLSemiLagrangian.prediction}) with one 32-bit {\tt sgemm}\footnote{\textbf{s}ingle-precision \textbf{ge}neral \textbf{m}atrix-\textbf{m}atrix multiplication: $C = \alpha A B + \beta C$, where $\alpha = 1$, and $\beta = 0$ in our case.} operation per simulation step.  This lower-precision function has not affected our system's accuracy since $\mathcal{F}_{6,8}(\cdot)$ learned from $\mathcal{D}'$ with the default single-floating-point TensorFlow configuration.  Lastly, each of the coming tests has run in a computer with 16 GB RAM, 2.2GHz processor frequency, and no multi-threading.  The reported wall times correspond to the shortest durations of ten repetitions for every experiment.  We have made available $\mathcal{F}_{6,8}(\cdot)$ and related objects used in this section at \url{https://github.com/UCSB-CASL/ECNet}.


\colorsubsection{Rotation}
\label{subsec:Rotation}

Let $\phi_{rot}(\vv{x})$ be a starting level-set function, such as \cref{eq:CircularLevelSetFunction}, with a circular interface of radius $r = 0.15$ centered at $(0, 0.75)$ within $\Omega \equiv [-1, +1]^2$.  If one considers the divergence-free velocity field

\begin{equation}
\vv{u}_{rot}(\vv{x}) = 
\left(\begin{array}{c}
	u(x,y) \\
	v(x,y) \\
\end{array}\right) = \frac{1}{\sqrt{2}}\left(\begin{array}{c}
	-y \\
	x
\end{array}\right),
\label{eq:rotation.velocityfield}
\end{equation}
the disk completes one revolution at $t^{end} = 2\pi\sqrt{2}$ since $\max{||\vv{u}_{rot}(\vv{x})||} = 1$, $\forall \vv{x} \in \Omega$.  \Cref{tbl:results.rotation.stats} shows our hybrid strategy's performance and the error for the interface location.  In addition, it includes results for the baseline semi-Lagrangian scheme of \cite{Mirzadeh;etal:16:Parallel-level-set} at various resolutions.  Note that our approach comprises not only the {\tt MLSemiLagrangian()} module but also a custom selective reinitialization.  As interleaved advection in \Cref{alg:IntAdvectAndCollectSamples}, selective redistancing protects a subset of the machine-learning-corrected level-set values when solving \cref{eq:Reinitialization}.  Furthermore, we alternate between the numerical scheme and the {\tt MLSemiLagrangian()} subroutine to regularize the inferred trajectory, as seen in \cref{fig:Overview}.

\begin{table}[!t]
	\centering
	\small
	\bgroup
	\def\arraystretch{1.1}%
	\begin{tabular}{|l|c|c|c|c|r|r|}
		\hline
		Method       & $\ell_c^{\max}$ & $\ell^1$ error     & $\ell^\infty$ error & Disk area & Area loss (\%) & Time (sec.) \\
		\hline \hline
		Ours                       & 6 & $\eten{4.672}{-4}$ &  $\eten{9.317}{-4}$ & $\eten{7.027}{-2}$ & 0.59  &  $3.874$ \\
		\hline
		\multirow{3}{*}{Numerical} & 6 & $\eten{3.380}{-3}$ &  $\eten{4.481}{-3}$ & $\eten{6.740}{-2}$ & 4.65  &  $3.578$ \\
 		                           & 7 & $\eten{8.545}{-4}$ &  $\eten{1.195}{-3}$ & $\eten{6.985}{-2}$ & 1.18  & $13.665$ \\
 		                           & 8 & $\eten{2.152}{-4}$ &  $\eten{3.083}{-4}$ & $\eten{7.048}{-2}$ & 0.30  & $53.956$ \\
		\hline
	\end{tabular}
	\egroup
	\caption{Rotation accuracy and performance assessment.  Reported errors include measurements taken at grid points for which $|\phi_{rot}(\vv{x})| \leqslant \sqrt{2}h_c$ at the end of one revolution.}
	\label{tbl:results.rotation.stats}
\end{table}

\begin{figure}[!t]
	\centering
    \begin{subfigure}[b]{0.32\textwidth}
		\includegraphics[width=\textwidth]{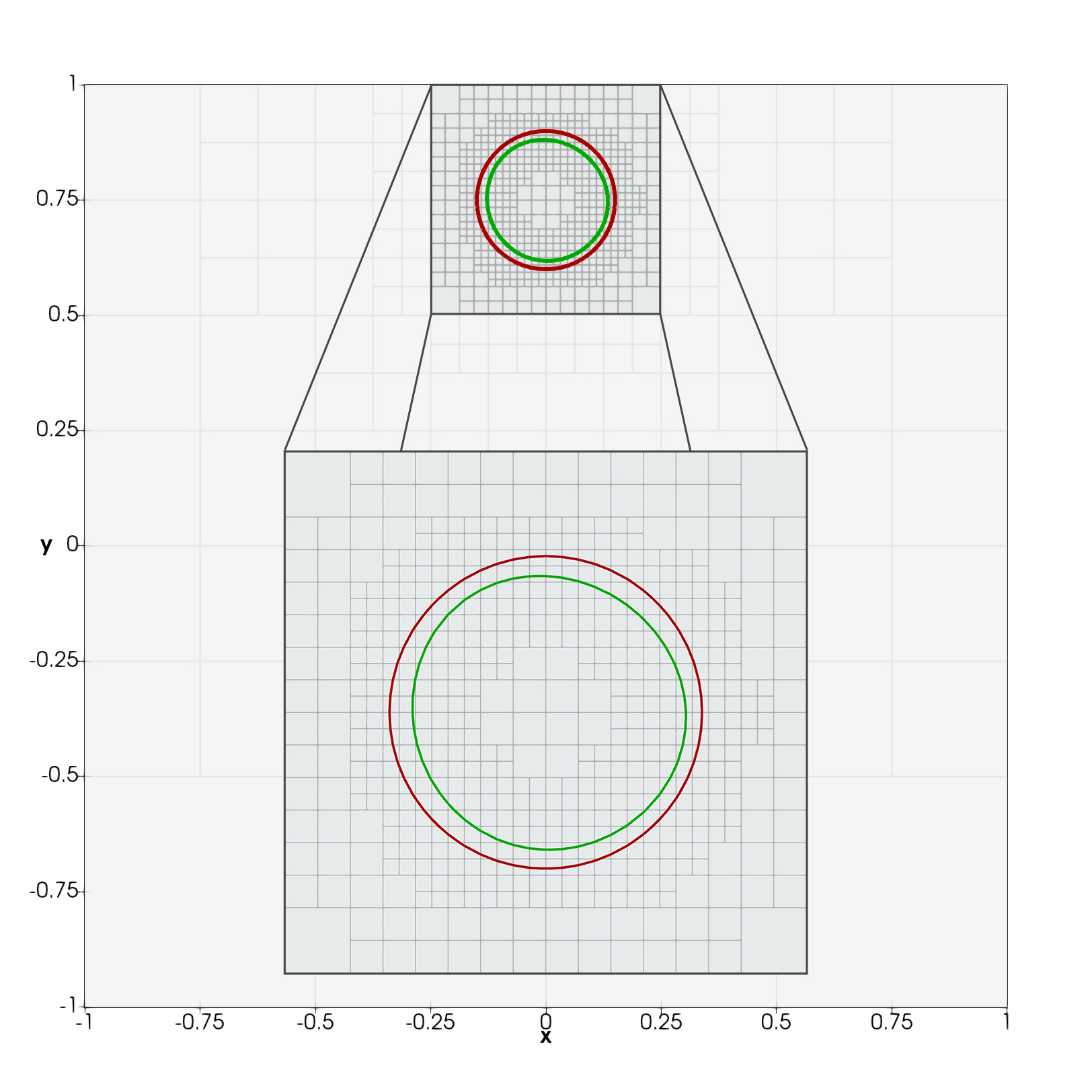}
        \label{fig:results.rotation.5.num6}
    \end{subfigure}
    ~
	\begin{subfigure}[b]{0.32\textwidth}
		\includegraphics[width=\textwidth]{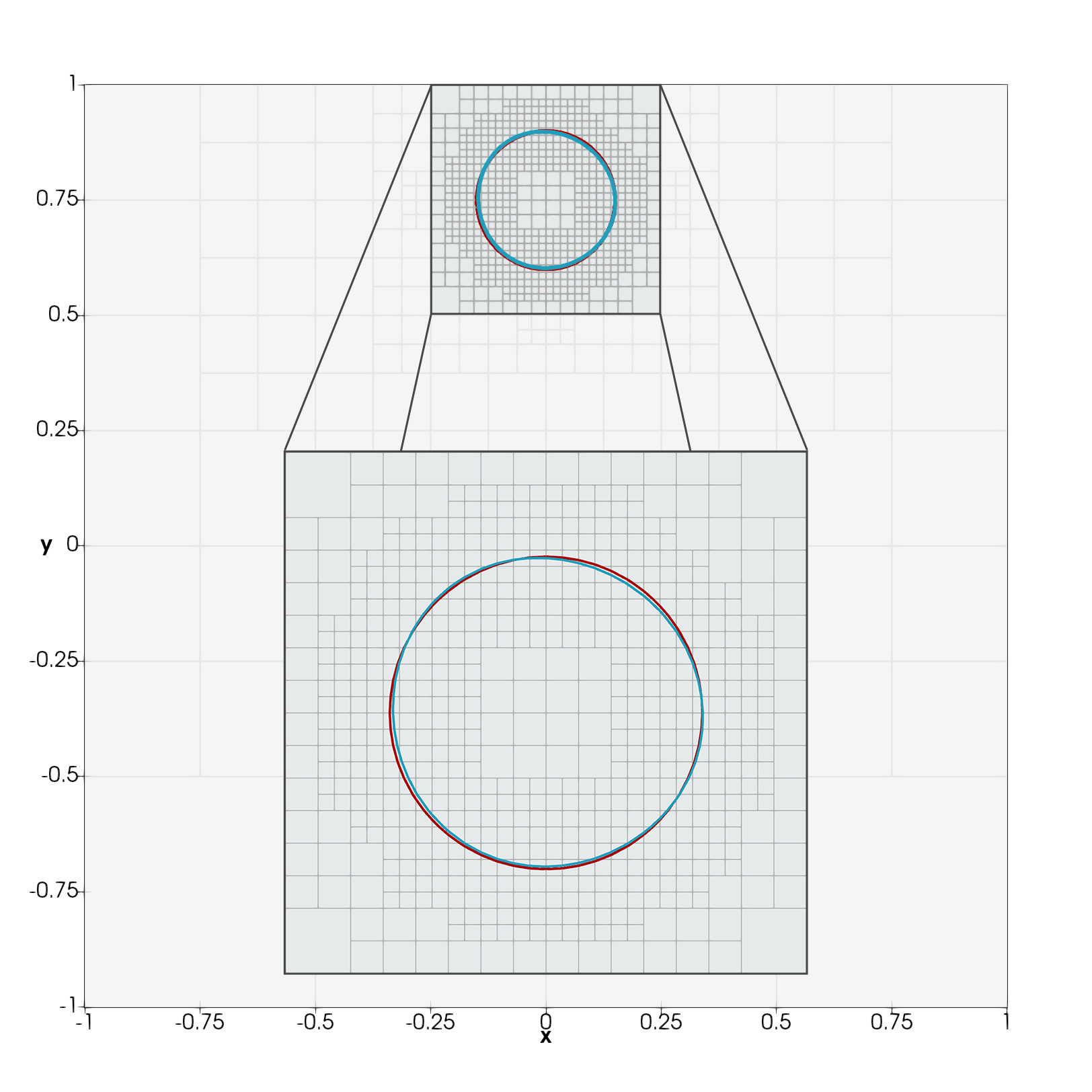}
        \label{fig:results.rotation.5.nnet6}
    \end{subfigure}
    ~
    \begin{subfigure}[b]{0.32\textwidth}
		\includegraphics[width=\textwidth]{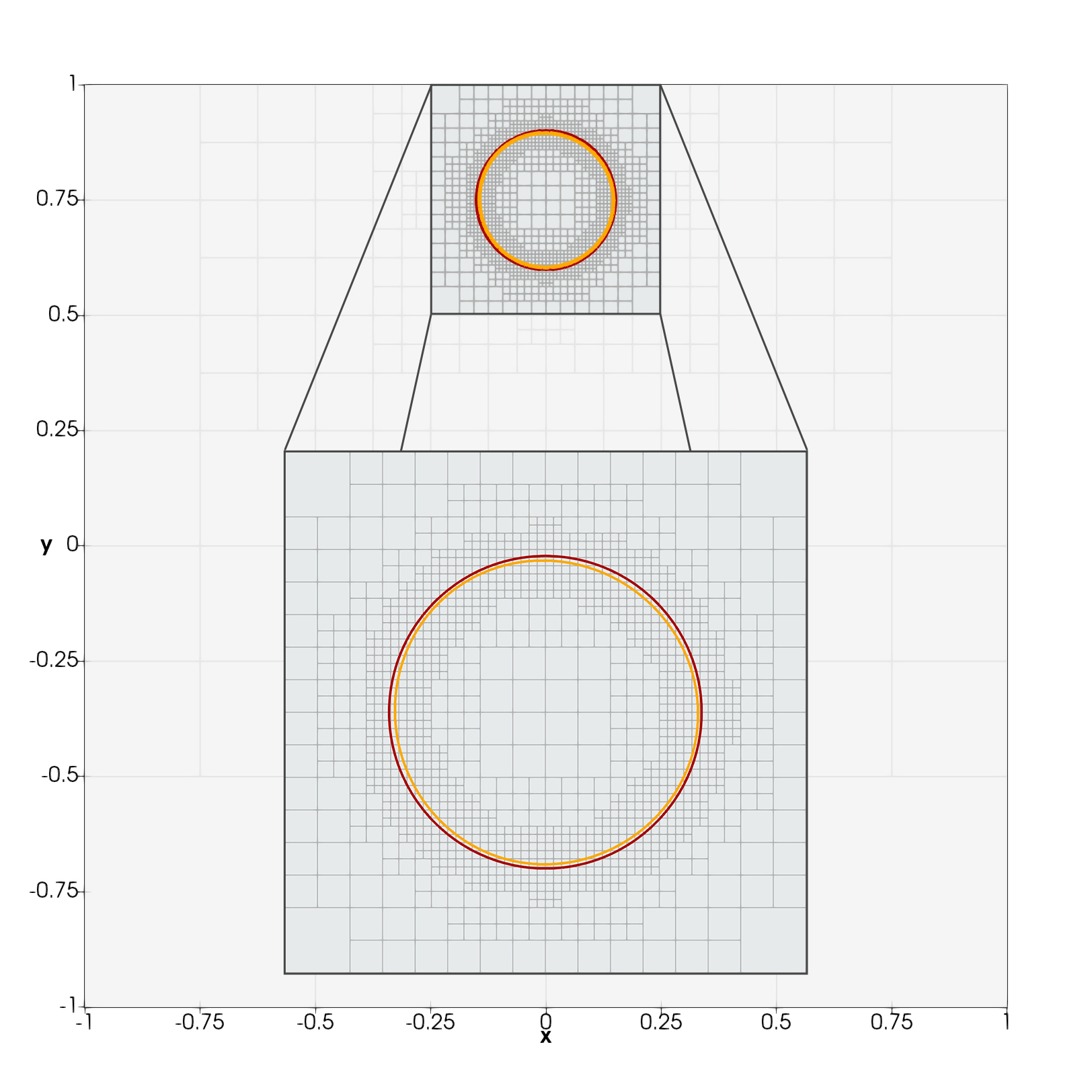}
        \label{fig:results.rotation.5.num7}
    \end{subfigure}
    \\
    \begin{subfigure}[b]{0.32\textwidth}
		\includegraphics[width=\textwidth]{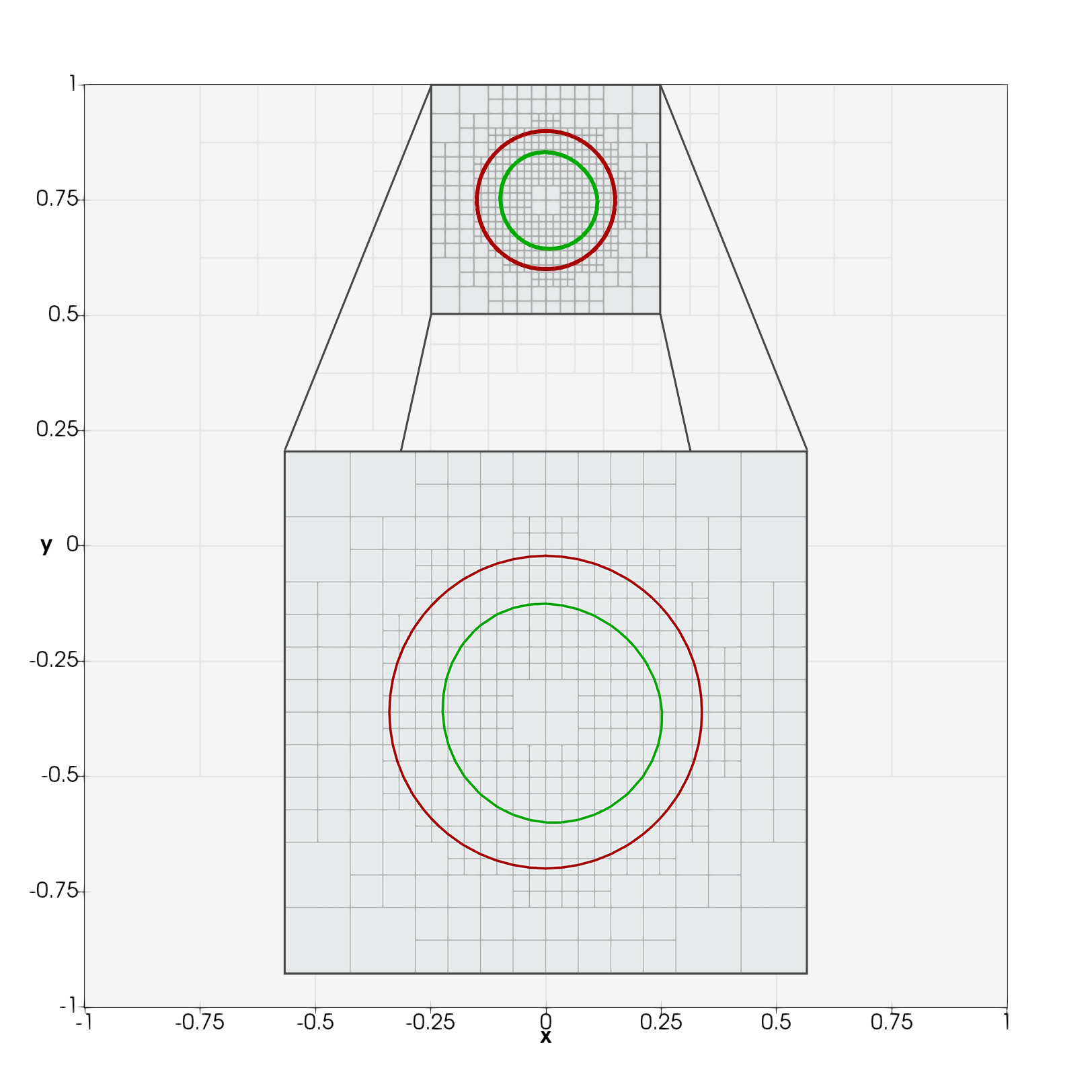}
        \caption{\footnotesize Numerical baseline for $h_c = 2^{-6}$}
        \label{fig:results.rotation.10.num6}
    \end{subfigure}
    ~
	\begin{subfigure}[b]{0.32\textwidth}
		\includegraphics[width=\textwidth]{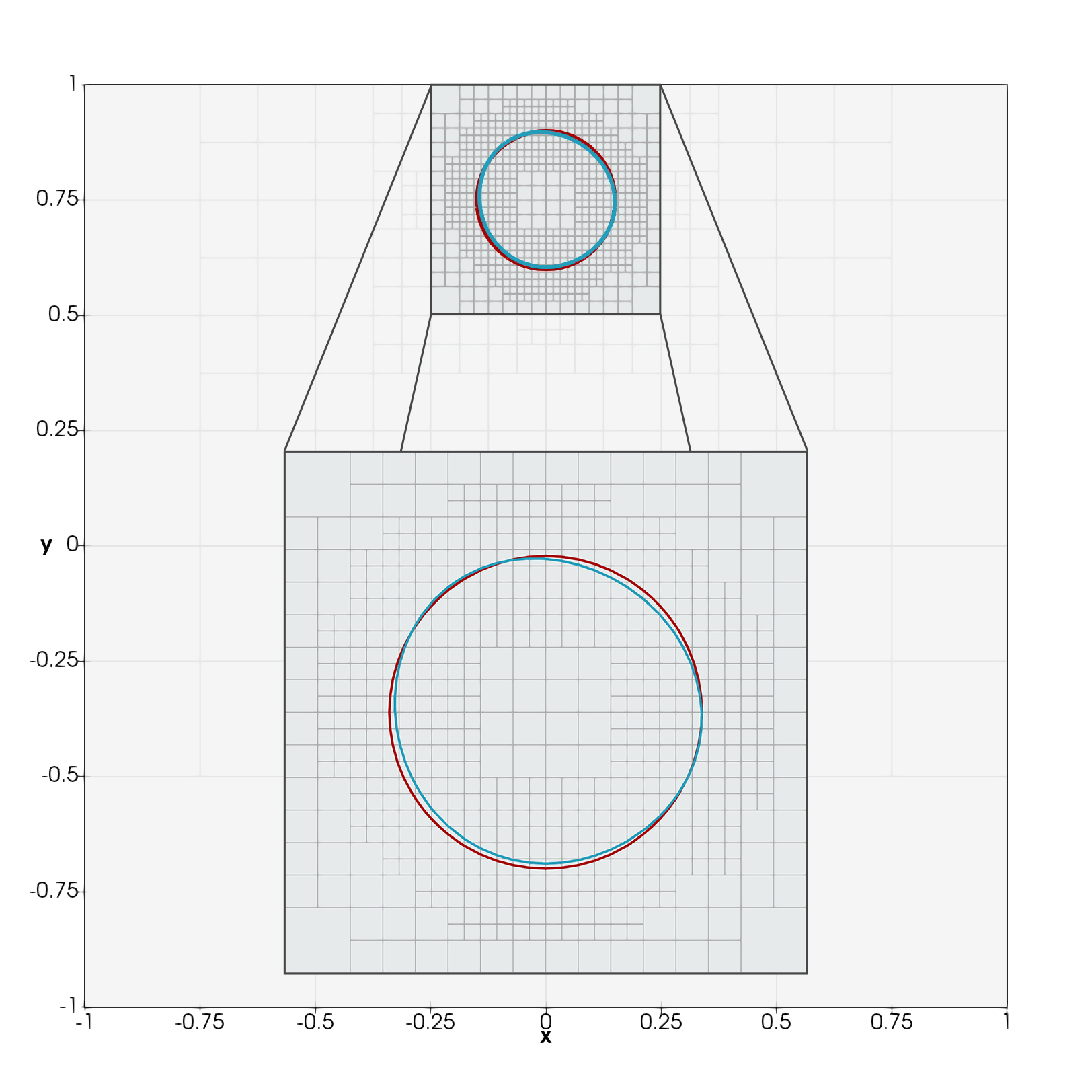}
        \caption{\footnotesize Hybrid approach for $h_c = 2^{-6}$}
        \label{fig:results.rotation.10.nnet6}
    \end{subfigure}
    ~
    \begin{subfigure}[b]{0.32\textwidth}
		\includegraphics[width=\textwidth]{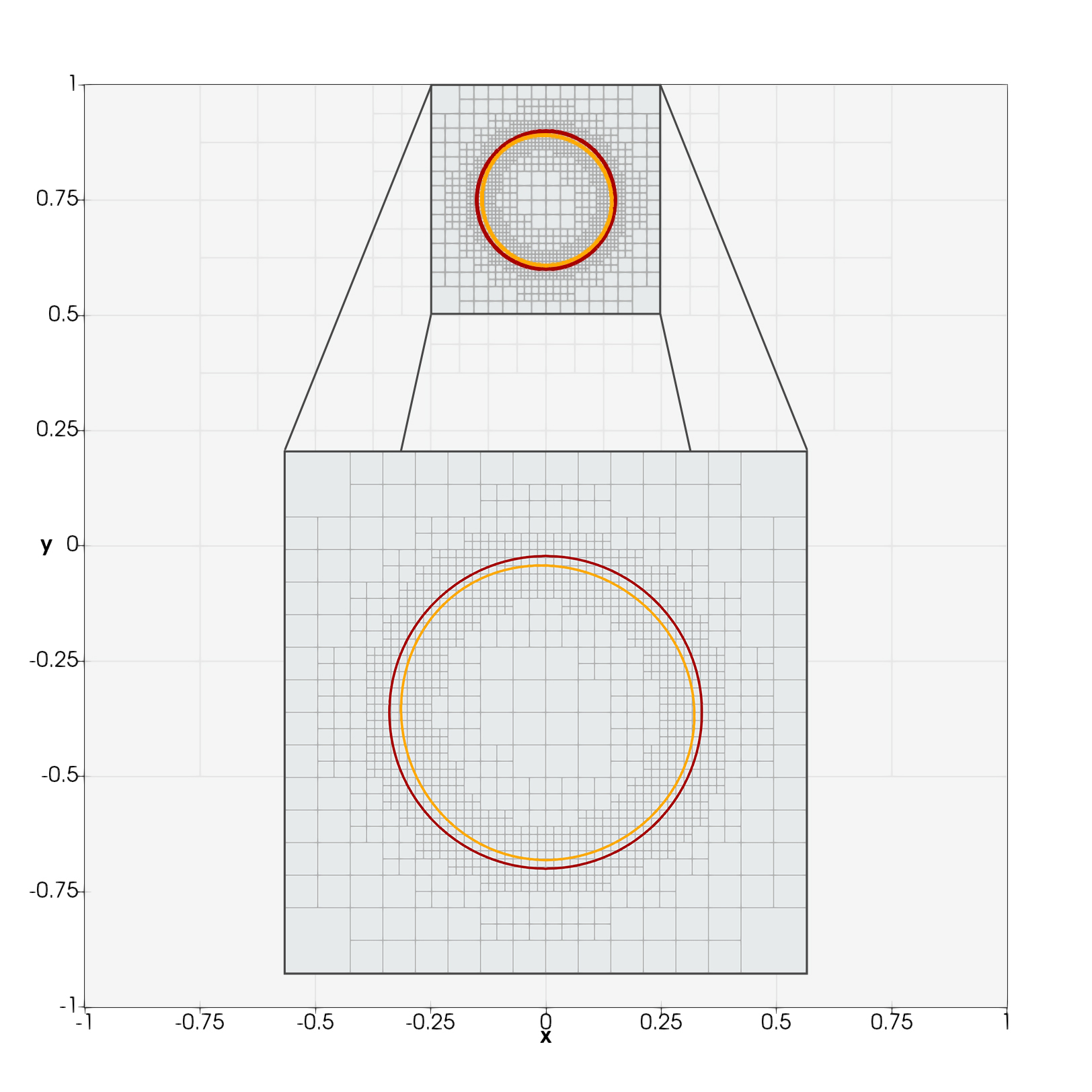}
        \caption{\footnotesize Numerical scheme for $h_c = 2^{-7}$}
        \label{fig:results.rotation.10.num7}
    \end{subfigure}
    
	\caption{Rotation-problem zero level sets at the end of five (i.e., $t^{end} = 10\pi\sqrt{2}$) (top row) and ten (i.e., $t^{end} = 20\pi\sqrt{2}$) (bottom row) revolutions.  The left and right columns show the numerical scheme in green and orange for $\ell_c^{\max} = 6$ and $7$.  The center column provides the machine-learning-corrected solution in blue for $\ell_c^{\max} = 6$.  In all cases, the analytical disk appears in red.  (Color online.)}
	\label{fig:results.rotation.5.10}
\end{figure}

Compared with the baseline at $\ell_c^{\max} = 6$, the statistics in \cref{tbl:results.rotation.stats} prove that $\mathcal{F}_{6,8}(\cdot)$ helps counteract numerical diffusion.  In particular, the average error reduces by a factor of 7.2, and the maximum error drops by around 79\%.  Also important is the fact that artificial area loss decreases from 4.65\% to 0.59\%.  Further, \cref{tbl:results.rotation.stats} reveals that our approach outperforms the numerical solver at twice the resolution.  For instance, the hybrid system preserves area more accurately than the baseline at $\ell_c^{\max} = 7$ but requires less than one-third of the computation time.  For completeness, the last row of \cref{tbl:results.rotation.stats} includes the results for the semi-Lagrangian scheme at the reference training resolution (i.e., $\ell_c^{\max} = 8$).  If one contrasts those measurements with ours, it is easy to see that the conventional method delivers smaller $\ell^1$ and $\ell^\infty$ error norms, as expected.  This behavior echoes the findings of Zhuang \etal in \cite{Zhuang;etal;LrndDiscForPassSclrAdvctn2D;2021}, where an advecting neural model trained on a reference grid at $8\times$ the resolution achieved roughly the same accuracy as the baseline at $4\times$ the resolution.  Extrapolating that rule into our work, we should anticipate area-preservation and level-set accuracy improvements analogous to the baseline outcomes at half the mesh size (i.e., $h_c = 2^{-7}$ ).  \Cref{fig:results.rotation.5.10} gives a visual perspective to our results corresponding to five and ten revolutions.  We also include the standard solution for $\ell_c^{\max} = 7$ to ease the comparison.

\begin{figure}[!t]
	\centering
	\begin{subfigure}[b]{5.2cm}
		\includegraphics[width=\textwidth]{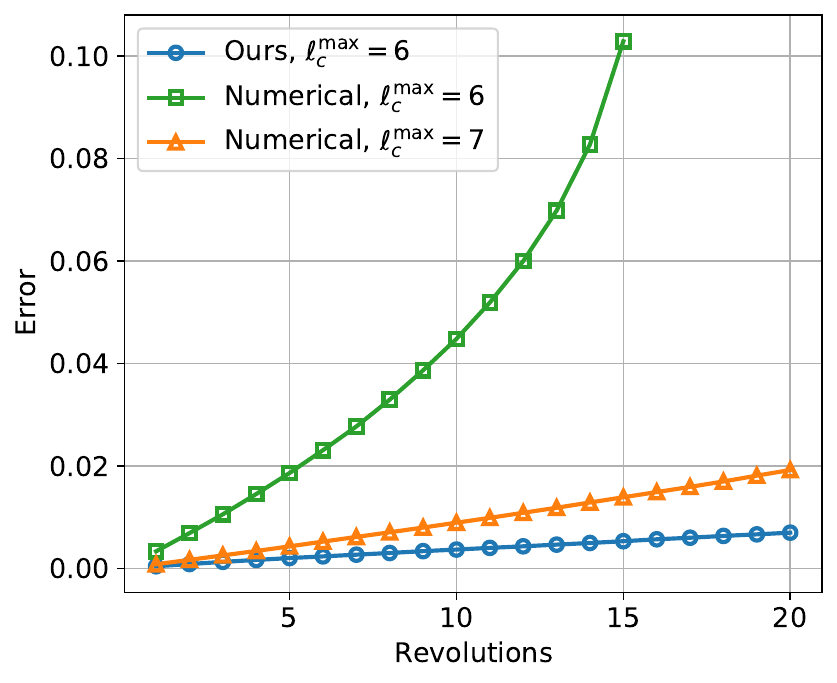}
        \caption{\footnotesize Mean absolute error}
        \label{fig:results.rotation.stress.stats.mae}
    \end{subfigure}
    ~
	\begin{subfigure}[b]{5.2cm}
		\includegraphics[width=\textwidth]{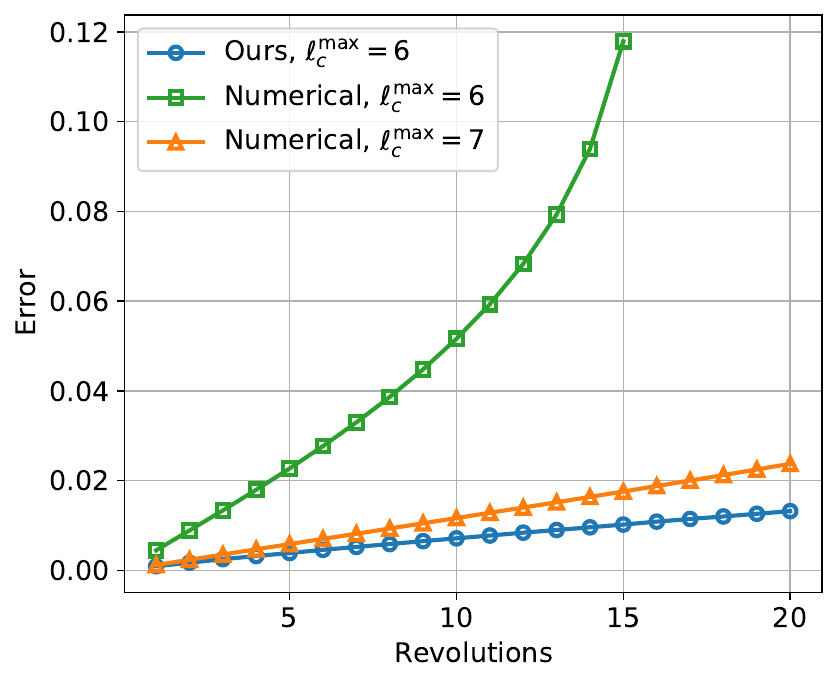}
        \caption{\footnotesize Maximum absolute error}
        \label{fig:results.rotation.stress.stats.maxae}
    \end{subfigure}
    ~
    \begin{subfigure}[b]{5.1cm}
		\includegraphics[width=\textwidth]{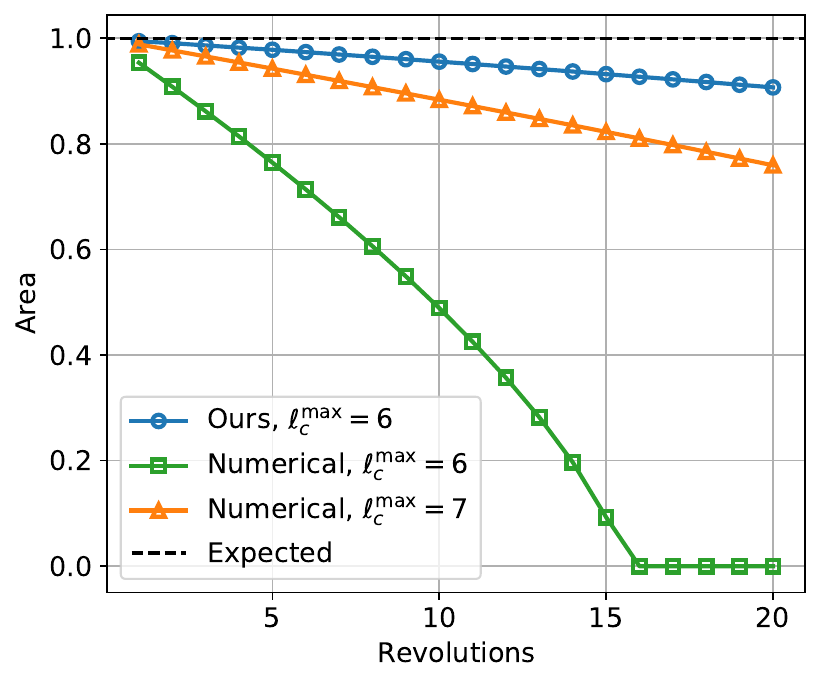}
        \caption{\footnotesize Area conservation}
        \label{fig:results.rotation.stress.stats.area}
    \end{subfigure}
    
	\caption{Stress test for the rotation experiment using the semi-Lagrangian scheme and our hybrid advection system for $h_c = 2^{-6}$.  For comparison, we include the results for the numerical approach with $h_c = 2^{-7}$.  The left and center plots show the $\ell^1$ and $\ell^\infty$ error norms over $\phi$ at the end of each revolution.  The right plot describes the evolution of the (normalized) area, taking $\eten{7.068583}{-2}$ as a reference.   (Color online.)}
	\label{fig:results.rotation.stress.stats}
\end{figure}

We conclude this experiment with a stress test.  We compare the accuracy of our approach with the ordinary semi-Lagrangian scheme at the same ($\ell_c^{\max} = 6$) and twice the grid resolution ($\ell_c^{\max} = 7$).  This time, the simulation lasts up to twenty revolutions.  The resulting statistics appear in \cref{fig:results.rotation.stress.stats}.

\Cref{fig:results.rotation.stress.stats} confirms that the machine-learning-augmented semi-Lagrangian routine is superior to the baseline scheme at a compatible grid discretization with $h_c = 2^{-6}$.  Here, the corrected trajectory remains close to the analytical solution, whereas the numerical interface vanishes right after the fifteenth revolution.  Our strategy also outperforms the numerical accuracy at twice the resolution.  Specifically, $\mathcal{F}_{6,8}(\cdot)$ helped reduce area loss by a factor of 2.6 by the end of the twentieth rotation.  The latter validates that our hybrid framework minimizes the effects of unwelcome diffusion on a comparable level with a conventional solver in a discretized domain with $h_c = 2^{-7}$.  Once more, these machine learning benefits come at a fraction of the cost required to solve the same problem in a mesh with eight-level unit-square quadtrees.  

\begin{figure}[!t]
	\centering
	\begin{subfigure}[b]{0.32\textwidth}
		\includegraphics[width=\textwidth]{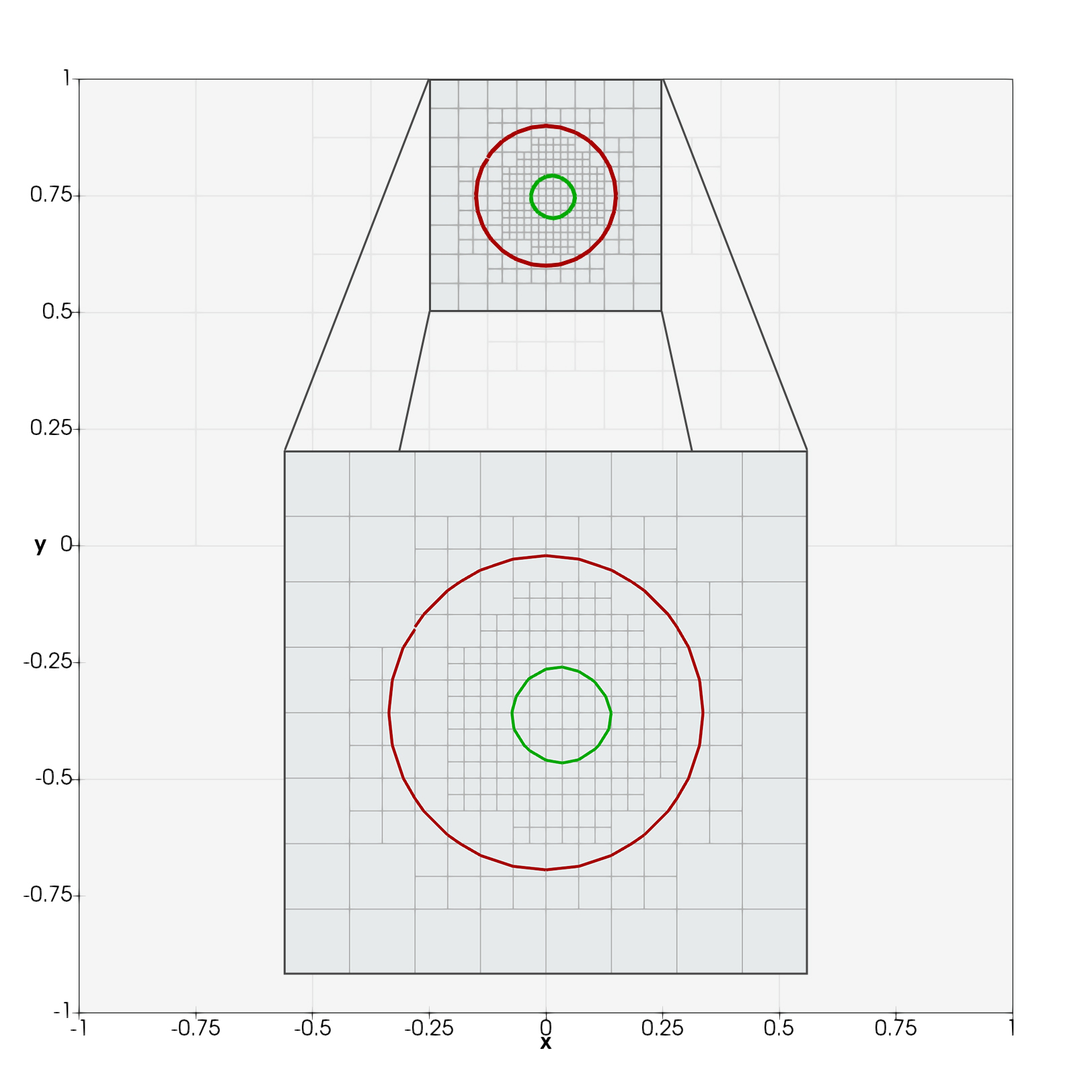}
        \caption{\footnotesize Numerical baseline for $h_c = 2^{-6}$}
        \label{fig:results.rotation.15.num6}
    \end{subfigure}
    ~
	\begin{subfigure}[b]{0.32\textwidth}
		\includegraphics[width=\textwidth]{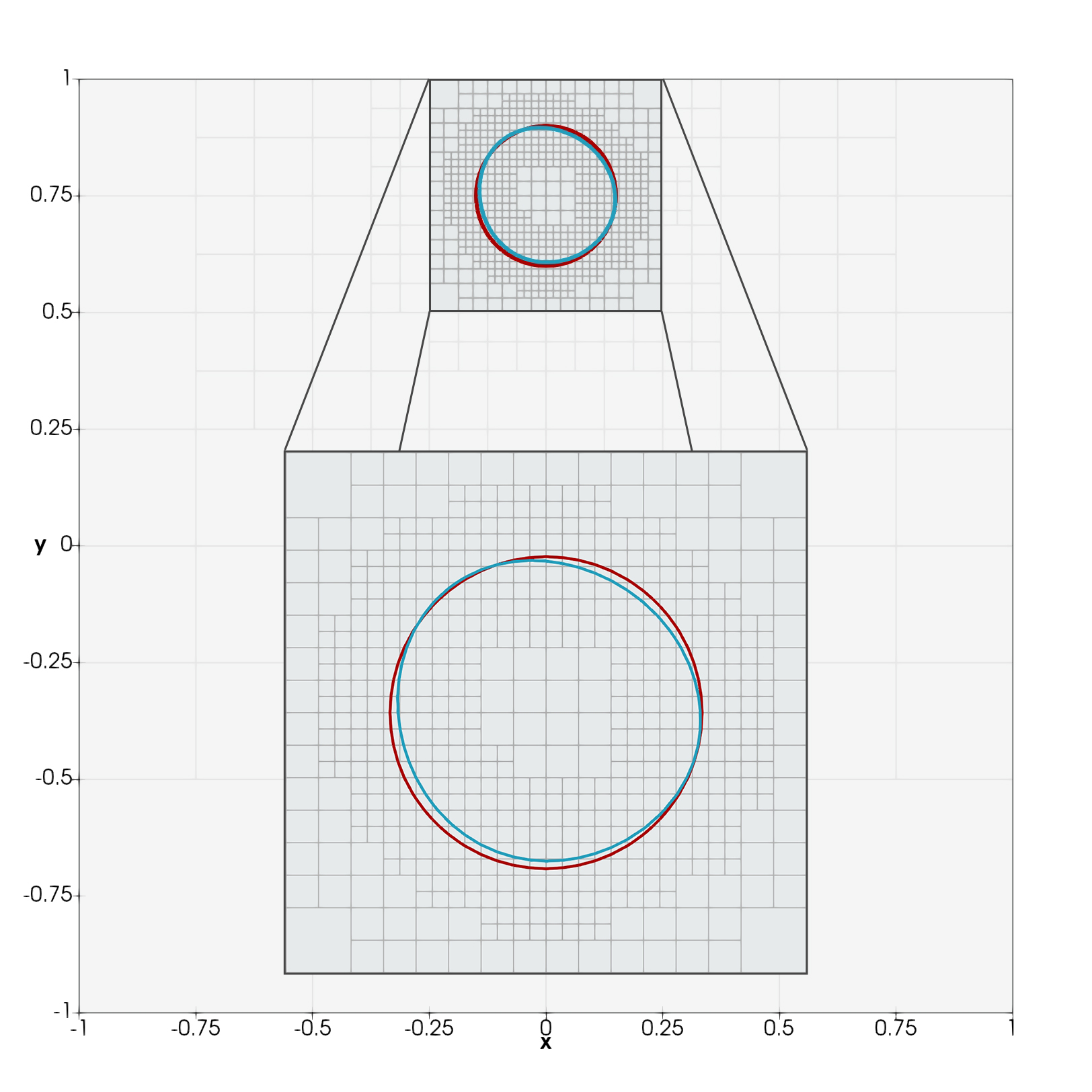}
        \caption{\footnotesize Hybrid approach for $h_c = 2^{-6}$}
        \label{fig:results.rotation.15.nnet6}
    \end{subfigure}
    ~
    \begin{subfigure}[b]{0.32\textwidth}
		\includegraphics[width=\textwidth]{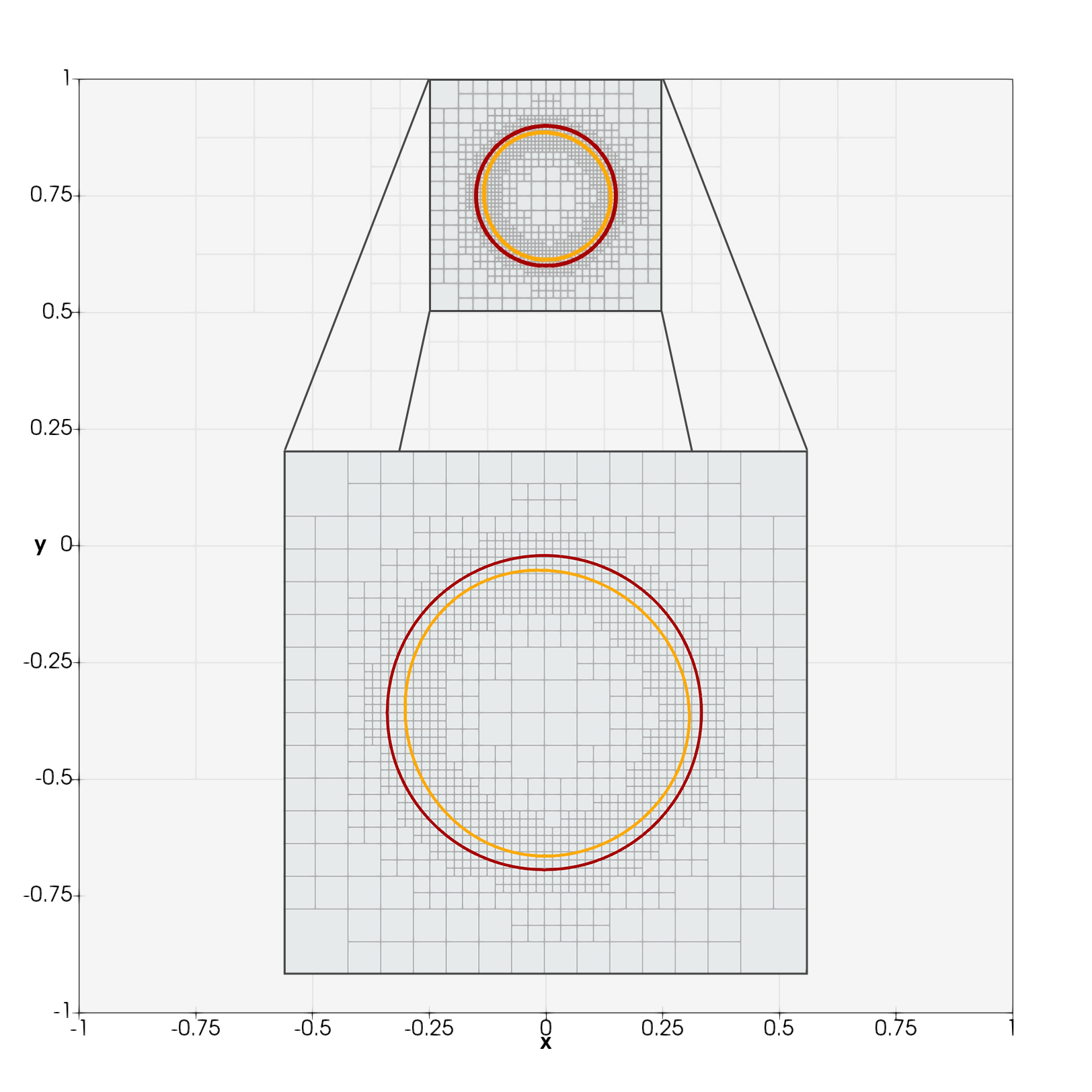}
        \caption{\footnotesize Numerical scheme for $h_c = 2^{-7}$}
        \label{fig:results.rotation.15.num7}
    \end{subfigure}
    
	\caption{Rotation-problem zero level sets at the end of the fifteenth revolution (i.e., $t^{end} = 30\pi\sqrt{2}$).  The left and center panels show the numerical baseline in green and the machine-learning-corrected solution in blue for $h_c = 2^{-6}$.  For comparison, we include the numerical results for a grid with twice the resolution (i.e., $h_c = 2^{-7}$) in the right panel in orange.  The analytical disk appears in red for all cases.  (Color online.)}
	\label{fig:results.rotation.15}
\end{figure}

\Cref{fig:results.rotation.15} closes this case study with zero-isocontour illustrations at the end of the fifteenth revolution for the three tested systems.  Finally, \cref{fig:results.rotation.stress.stats,fig:results.rotation.15} verify that our approach remains stable for all simulation times, independently from the chosen $t^{end} = 0.5$ in \Cref{alg:GenerateDataSet}.  However, as seen in \cref{fig:results.rotation.15.nnet6}, bias artifacts can emerge as $t^n \rightarrow \infty$.  To tackle this phenomenon, we plan to enforce physical constraints, introduce temporal information during training, or consider a different numerical transport method.

\colorsubsection{Vortex}
\label{subsec:Vortex}

Next, we consider the more challenging vortex deformation flow proposed by \cite{Bell;Colella;Glaz:89:A-second-order-proje}.  Let $\phi_{vtx}(\vv{x})$ be an initial level-set function, such as \cref{eq:CircularLevelSetFunction}, with a circular interface of radius $r = 0.15$ centered at $(0.5, 0.75)$ within $\Omega \equiv [0, 1]^2$.  If we define the divergence-free velocity field

\begin{equation}
\vv{u}_{vtx}(\vv{x}) = 
\left(\begin{array}{c}
	u(x,y) \\
	v(x,y) \\
\end{array}\right) = \left(\begin{array}{c}
	-\sin^2(\pi x) \sin(2\pi y) \\
	\sin^2(\pi y) \sin(2\pi x)
\end{array}\right),
\label{eq:vortex.velocityfield}
\end{equation}
the circle deforms into a (clockwise) spiral with $\max{||\vv{u}_{vtx}(\vv{x})||} = 1$, $\forall \vv{x} \in \Omega$.  In this experiment, we evolve the starting interface forward with \cref{eq:vortex.velocityfield} until $t^{mid} = 0.625$ and then backward using $-\vv{u}_{vtx}(\vv{x})$.  We want to examine how accurately we can reconstruct the original disk at $t^{end} = 1.25$.

\begin{table}[!t]
	\centering
	\small
	\bgroup
	\def\arraystretch{1.1}%
	\begin{tabular}{|l|c|c|c|c|r|r|}
		\hline
		Method       & $\ell_c^{\max}$ & $\ell^1$ error     & $\ell^\infty$ error & Disk area & Area loss (\%) & Time (sec.) \\
		\hline \hline
		Ours                       & 6 & $\eten{9.369}{-4}$ &  $\eten{3.751}{-3}$ & $\eten{7.054}{-2}$ & 0.20  & $0.656$ \\
		\hline
		\multirow{3}{*}{Numerical} & 6 & $\eten{1.329}{-3}$ &  $\eten{5.686}{-3}$ & $\eten{6.986}{-2}$ & 1.17  & $0.602$ \\
 		                           & 7 & $\eten{3.367}{-4}$ &  $\eten{2.361}{-3}$ & $\eten{7.034}{-2}$ & 0.49  & $2.369$ \\
 		                           & 8 & $\eten{9.882}{-5}$ &  $\eten{9.487}{-4}$ & $\eten{7.059}{-2}$ & 0.14  & $9.715$ \\
		\hline
	\end{tabular}
	\egroup
	\caption{Vortex-deformation accuracy and performance assessment.  Reported errors include measurements taken at grid points for which $|\phi_{vtx}(\vv{x})| \leqslant \sqrt{2}h_c$ at $t^{end} = 1.25$.}
	\label{tbl:results.vortex.stats}
\end{table}

\begin{figure}[!t]
	\centering
	\begin{subfigure}[b]{0.45\textwidth}
		\includegraphics[width=\textwidth]{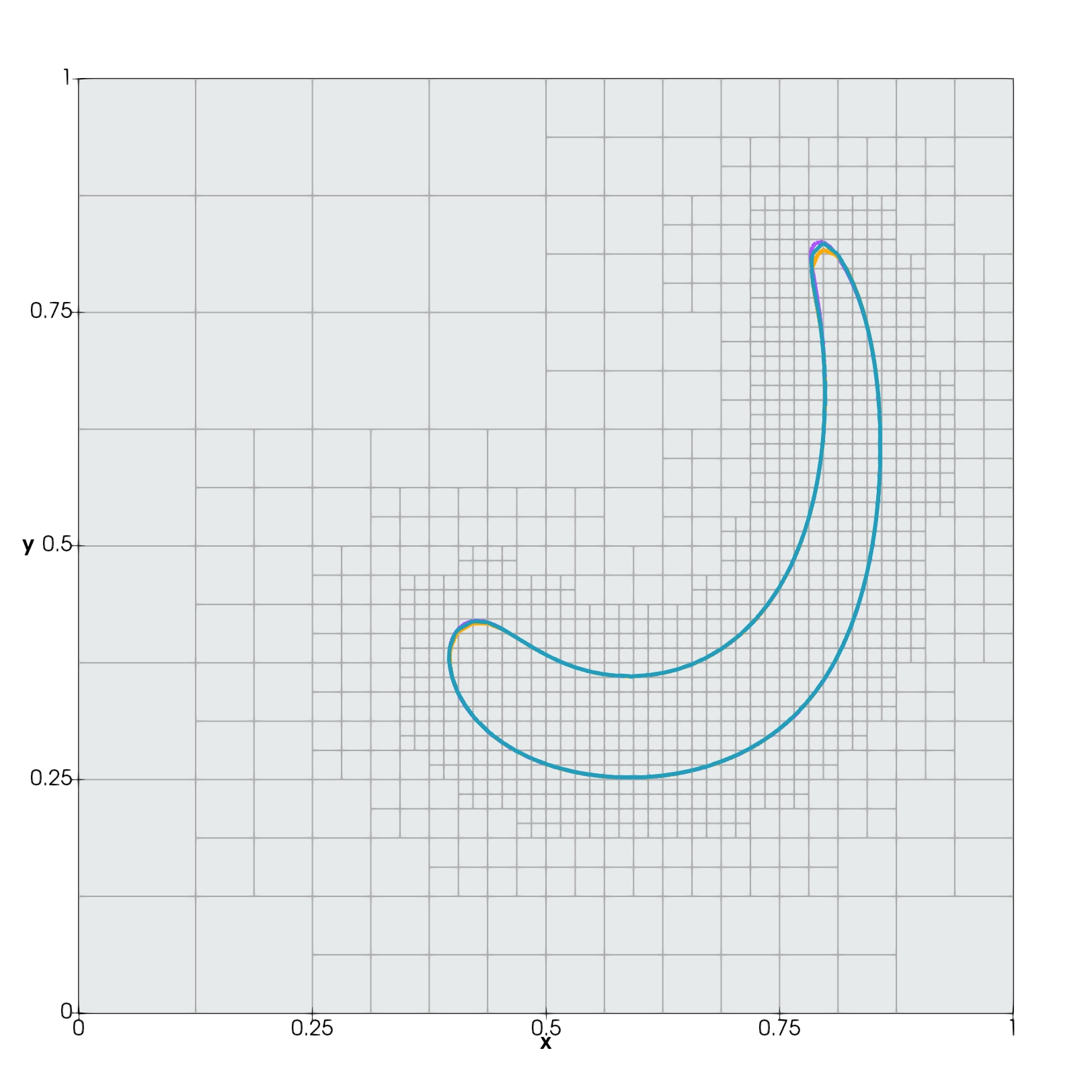}
        \caption{\footnotesize Maximal deformation at time $t^{mid} = \frac{1}{2}t^{end}$}
        \label{fig:results.vortex.middle}
    \end{subfigure}
    ~
	\begin{subfigure}[b]{0.45\textwidth}
		\includegraphics[width=\textwidth]{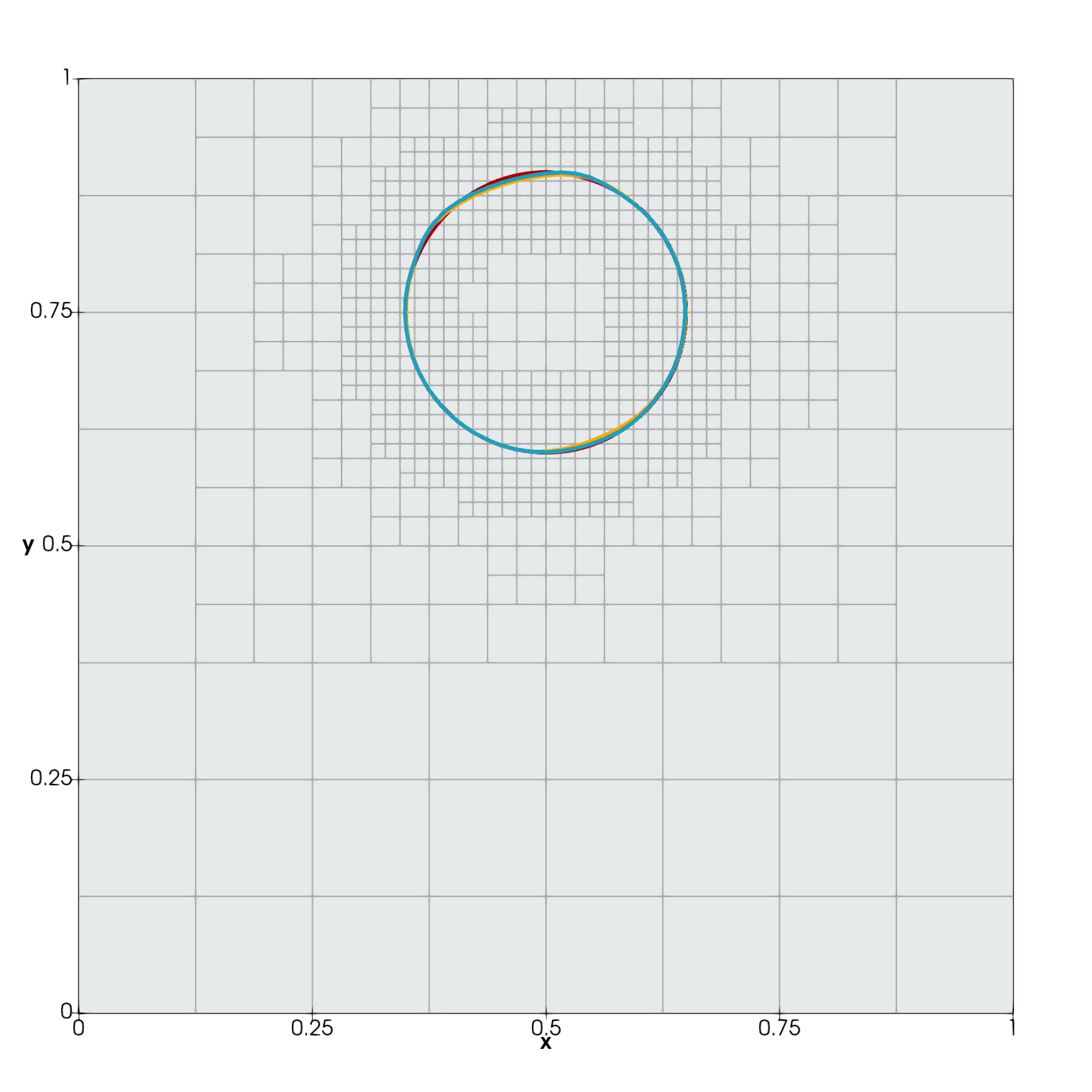}
        \caption{\footnotesize Original disk reconstructed at time $t^{end}$}
        \label{fig:results.vortex.final}
    \end{subfigure}
    
	\caption{Zero level sets for the vortex experiment at the intermediate and final simulation times.  The numerical baseline appears in orange for $\ell_c^{\max} = 6$ and purple for $\ell_c^{\max} = 7$.  We show the machine-learning-corrected trajectory in blue for $\ell_c^{\max} = 6$.  For comparison, the analytical contour is only visible in the right panel in red.  (Color online.)}
	\label{fig:results.vortex}
\end{figure}

\begin{table}[!t]
	\centering
	\small
	\bgroup
	\def\arraystretch{1.1}%
	\begin{tabular}{|l|c|c|c|r|}
		\hline
		Method    & $\ell^1$ error     & $\ell^\infty$ error & Disk area          & Area loss (\%) \\
		\hline \hline
		Ours      & $\eten{3.232}{-3}$ &  $\eten{1.164}{-2}$ & $\eten{7.082}{-2}$ & -0.20 \\
		\hline
		Numerical & $\eten{4.185}{-3}$ &  $\eten{1.439}{-2}$ & $\eten{7.021}{-2}$ &  0.68 \\
		\hline
	\end{tabular}
	\egroup
	\caption{Degeneracy in the vortex-deformation experiment.  Reported errors include measurements taken at grid points for which $|\phi_{vtx}(\vv{x})| \leqslant \sqrt{2}h_c$ at $t^{end} = 2$ for $h_c = 2^{-6}$.}
	\label{tbl:results.vortex.degeneracy.stats}
\end{table}

\Cref{tbl:results.vortex.stats} summarizes this experiment's interface-location accuracy and performance metrics for several discretizations.  According to these statistics, our method is superior to the numerical baseline at $\ell_c^{\max} = 6$, but it is not as good as the conventional semi-Lagrangian advection at twice the resolution (at least regarding the MAE).  In particular, $\mathcal{F}_{6,8}(\cdot)$ helps reduce the $\ell^1$ error norm by $30\%$, the maximum error by one-third, and the area loss by $83\%$ when $h_c = 2^{-6}$.  
    
\Cref{fig:results.vortex} confirms the findings in \cref{tbl:results.vortex.stats} with visual evidence from the intermediate and final simulation states.  Again, the machine learning solution requires a minimal fraction of the cost for realizing the numerical approach on finer grids.  However, this time, it is more difficult for the proposed system to recover the accuracy lost to under-resolution.  For example, if the simulation prolongs to $t^{end} = 2$, the hybrid strategy degenerates to being only slightly better than its numerical counterpart (see \cref{tbl:results.vortex.degeneracy.stats}).  In this regard, we hypothesize that the observed behavior stems from the lack of physical constraints to enforce smoothness along the moving boundary.  The mass gain in \cref{tbl:results.vortex.degeneracy.stats} is probably a symptom of this problem, revealing our system's inability to redistribute area in highly deforming flows.  Thus, this vortex test exposes the need to integrate more accurate smoothing mechanisms and improve the training protocol by encompassing more patterns from under-resolved regions.


\colorsubsection{Circular-vortex-patch problem}
\label{subsec:CircularVortexPatch}

The following experiment measures the effects of tangential shear flows on a stationary interface.  Let $\Gamma$ be the zero-isocontour of the level-set function $\phi_{cvp}(\vv{x})$ in \cref{eq:CircularLevelSetFunction}.  Also, consider the simplified circular-vortex-patch problem, where a fixed disk $D_{cvp}$ of radius $r_{cvp} = 0.6$, randomly centered at $\vv{x}_{cvp} = (x_{cvp},\, y_{cvp})^T \in [-h_c/2, +h_c/2]^2$, encloses a rotational velocity field given by

\begin{equation}
\vv{u}_{cvp}(\vv{x}) = 
\left(\begin{array}{c}
	u(x,y) \\
	v(x,y) \\
\end{array}\right) = \left\{
\begin{array}{ll}
	\dfrac{1}{r_{cvp}}\left(\begin{array}{c}
	-(y - y_{cvp}) \\
	(x - x_{cvp})
\end{array}\right), & \textrm{if }\|\vv{x} - \vv{x}_{cvp}\| < r_{cvp},\\
	\vv{0},         & \textrm{otherwise}.
\end{array}\right.
\label{eq:shearflow.cvp.velocityfield}
\end{equation}
In other words, $D_{cvp} \subset \Omega \equiv [-1, 1]^2$ contains a divergence-free velocity field, where $\|\vv{u}_{cvp}(\vv{x})\| \leqslant 1$ for all $\vv{x} \in D_{cvp}$, and $\vv{u}_{cvp}$ evaluates to zero everywhere else.  Here, we assess how well we can preserve the initial circular profile $\Gamma(0) \equiv D_{cvp}$ throughout a numerical simulation.

\begin{table}[!t]
	\centering
	\small
	\bgroup
	\def\arraystretch{1.1}
	\begin{tabular}{|l|c|c|c|c|r|r|}
		\hline
		Method   & $\ell_c^{\max}$ & $\ell^1$ error & $\ell^\infty$ error & Disk area & Area loss (\%) & Time (sec.) \\
		\hline \hline
		Ours                       & 6 & $\eten{6.839}{-3}$  &  $\eten{9.071}{-3}$ & $1.105$   & $2.28$         &  $5.207$ \\ 
		\hline
		\multirow{3}{*}{Numerical} & 6 & $\eten{8.960}{-4}$  &  $\eten{1.239}{-4}$ & $1.127$   & $0.31$         &  $4.578$ \\ 
 		                           & 7 & $\eten{5.709}{-4}$  &  $\eten{7.469}{-4}$ & $1.129$   & $0.19$         & $19.035$ \\ 
 		                           & 8 & $\eten{2.542}{-4}$  &  $\eten{3.523}{-4}$ & $1.130$   & $0.09$         & $80.418$ \\ 
		\hline
	\end{tabular}
	\egroup
	\caption{Circular-vortex-patch-problem accuracy and performance assessment.  Reported errors include measurements taken at grid points for which $|\phi_{cvp}(\vv{x})| \leqslant \sqrt{2}h_c$ at $t^{end} = 2\pi r_{cvp}$.}
	\label{tbl:results.shearflow.cvp.stats}
\end{table}

The error and performance statistics taken at $t^{end} = 2k\pi r_{cvp}$, for $k = 1$, appear in \cref{tbl:results.shearflow.cvp.stats} for our system and the {\tt SemiLagrangian()} method at three resolutions.  $t^{end}$ is the time it would take for a particle to complete $k$ revolutions inside $D_{cvp}$, but $\Gamma(t)$ should not deform for any $t \geqslant 0$.  As seen in \cref{tbl:results.shearflow.cvp.stats}, our error-correcting neural network cannot produce satisfactory results for this short period.  When $\ell_c^{\max} = 6$, for example, our system cannot generalize to this kind of velocity data and worsens the trajectory quality by a factor of $7.3$ in the $\ell^\infty$ error norm.  However, \cref{tbl:results.shearflow.cvp.stats} does not tell the whole story, as we illustrate next.

\begin{figure}[!t]
	\centering
	\begin{subfigure}[b]{5.2cm}
		\includegraphics[width=\textwidth]{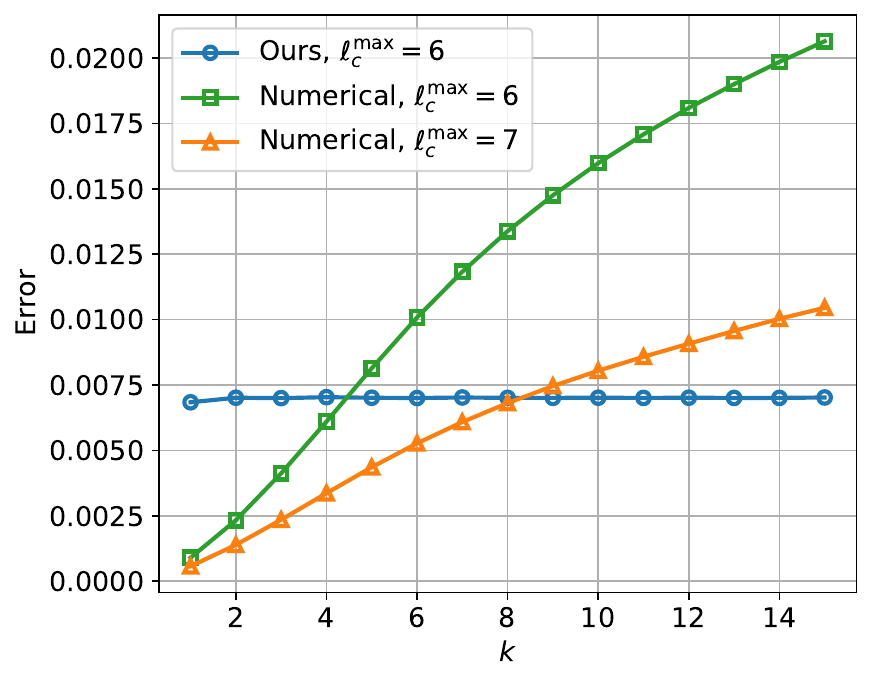}
		\caption{\footnotesize Mean absolute error}
		\label{fig:results.shearflow.cvp.stress.stats.mae}
	\end{subfigure}
    ~
	\begin{subfigure}[b]{5.2cm}
		\includegraphics[width=\textwidth]{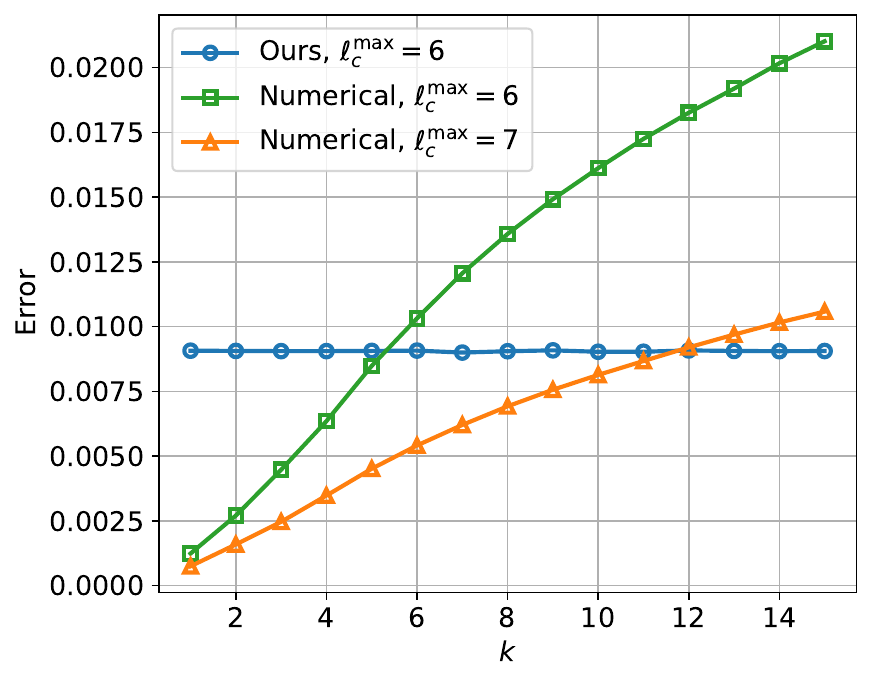}
		\caption{\footnotesize Maximum absolute error}
		\label{fig:results.shearflow.cvp.stress.stats.maxae}
	\end{subfigure}
	~
	\begin{subfigure}[b]{5.1cm}
		\includegraphics[width=\textwidth]{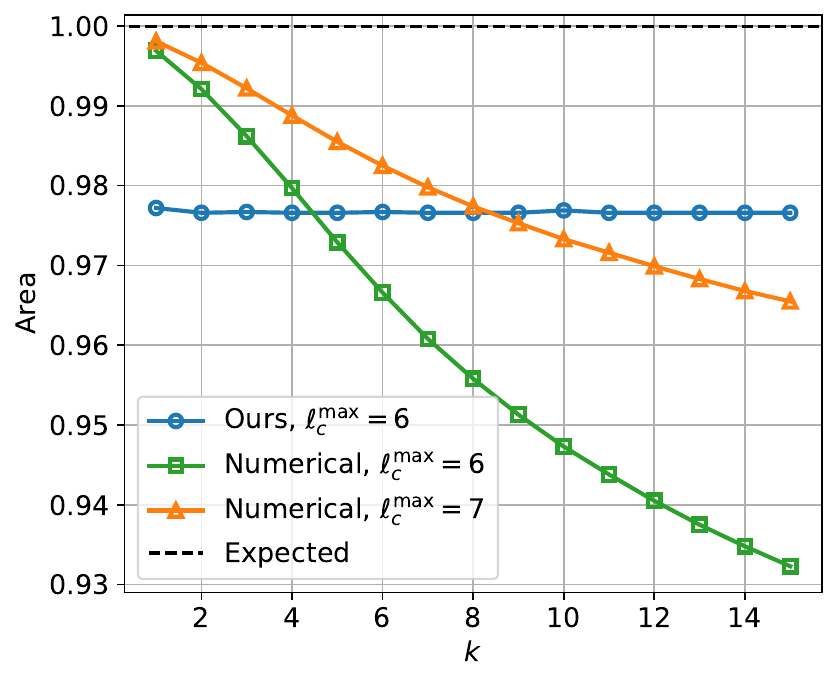}
		\caption{\footnotesize Area conservation}
		\label{fig:results.shearflow.cvp.stress.stats.area}
	\end{subfigure}
    
	\caption{Stress test for the circular-vortex-patch problem using the semi-Lagrangian scheme and our hybrid advection system for $h_c = 2^{-6}$.  For comparison, we include the results for the numerical approach with $h_c = 2^{-7}$.  The left and center plots show the $\ell^1$ and $\ell^\infty$ error norms over $\phi$ at the $k$th revolution.  The right plot describes the evolution of the (normalized) area, taking $1.130973$ as a reference.  (Color online.)}
	\label{fig:results.shearflow.cvp.stress.stats}
\end{figure}

\begin{figure}[!t]
	\centering
	\begin{subfigure}[b]{0.32\textwidth}
		\includegraphics[width=\textwidth]{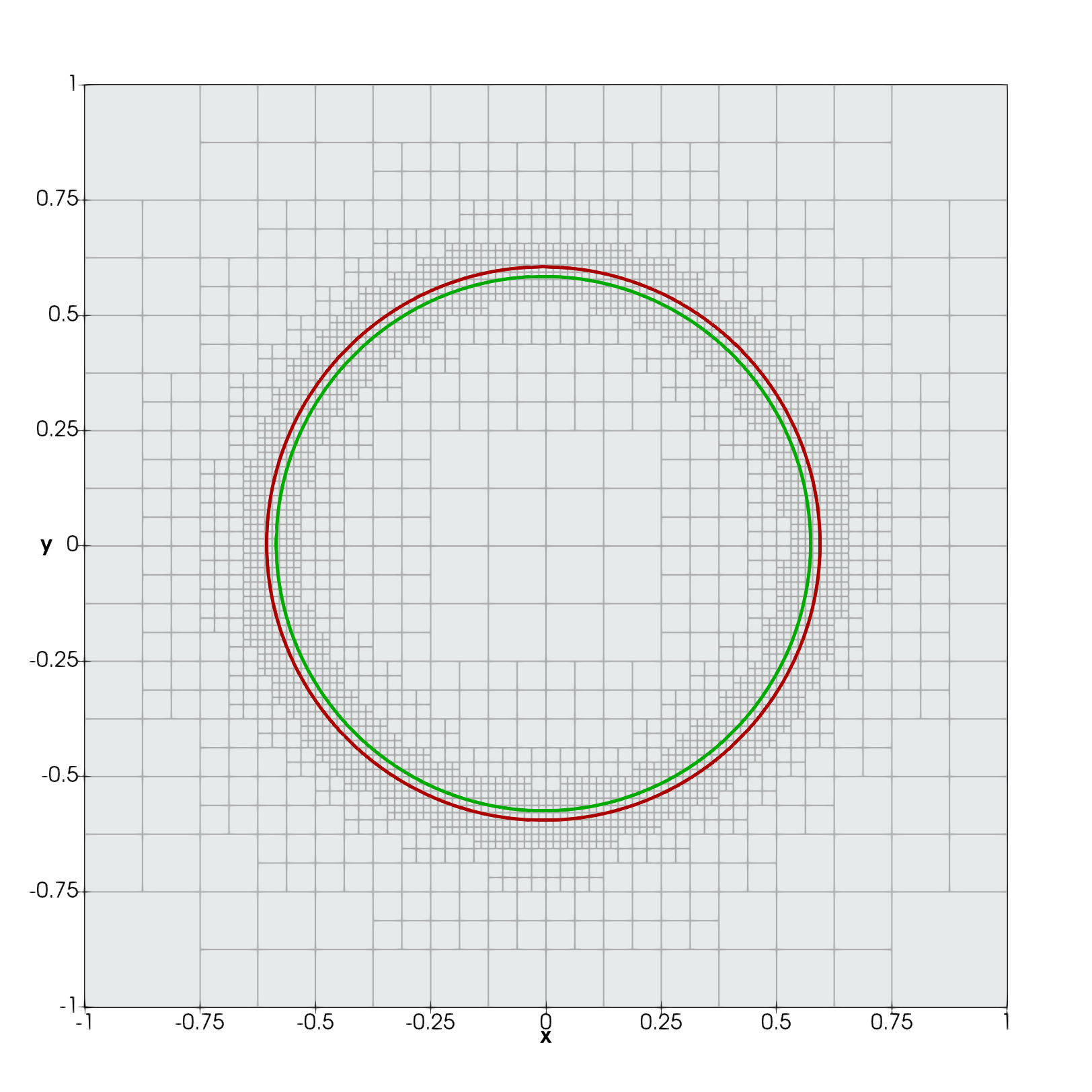}
		\caption{\footnotesize Numerical baseline for $h_c = 2^{-6}$}
		\label{fig:results.shearflow.cvp.num6}
	\end{subfigure}
	~
	\begin{subfigure}[b]{0.32\textwidth}
		\includegraphics[width=\textwidth]{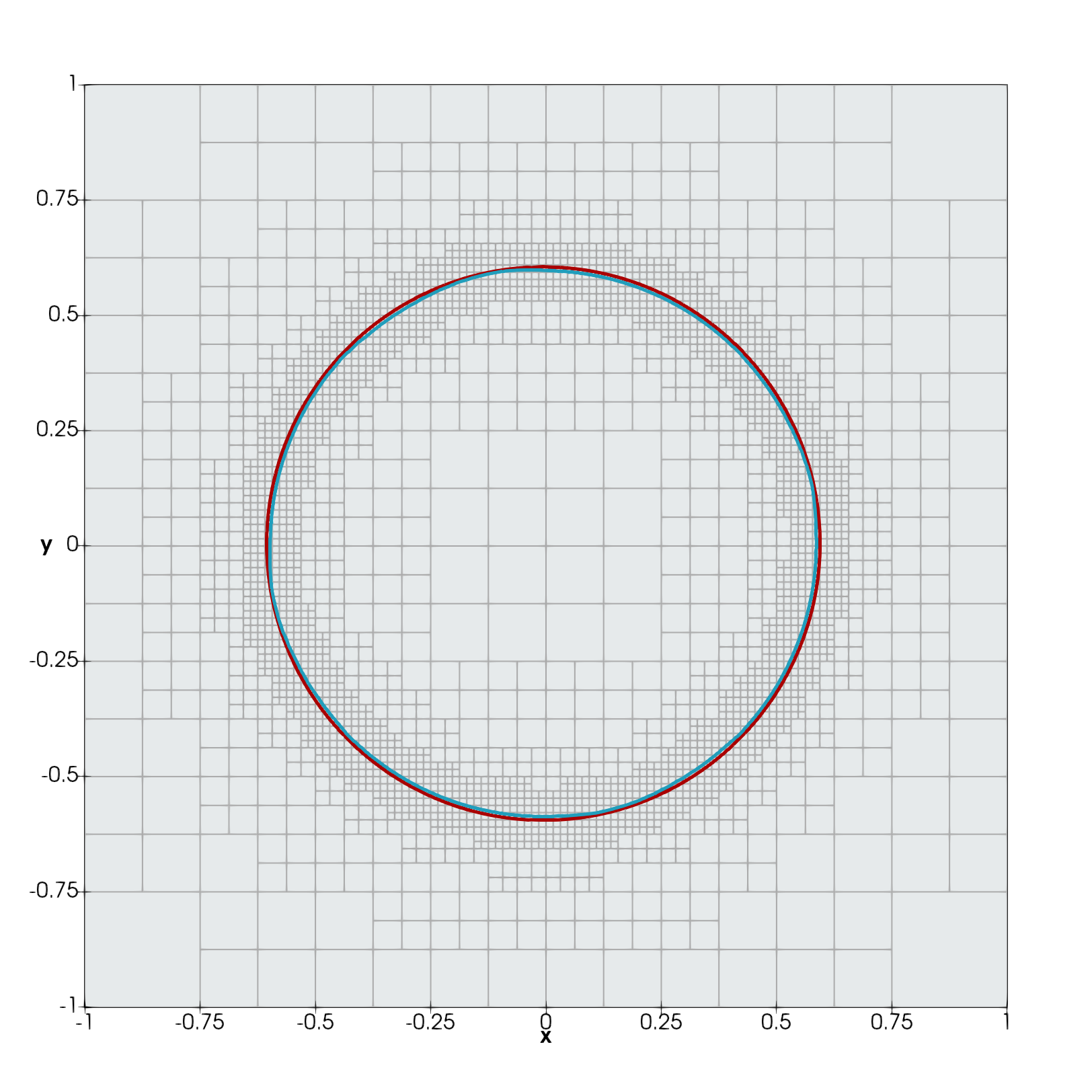}
		\caption{\footnotesize Hybrid system for $h_c = 2^{-6}$}
		\label{fig:results.shearflow.cvp.stress.nnet6}
	\end{subfigure}
	~
	\begin{subfigure}[b]{0.32\textwidth}
		\includegraphics[width=\textwidth]{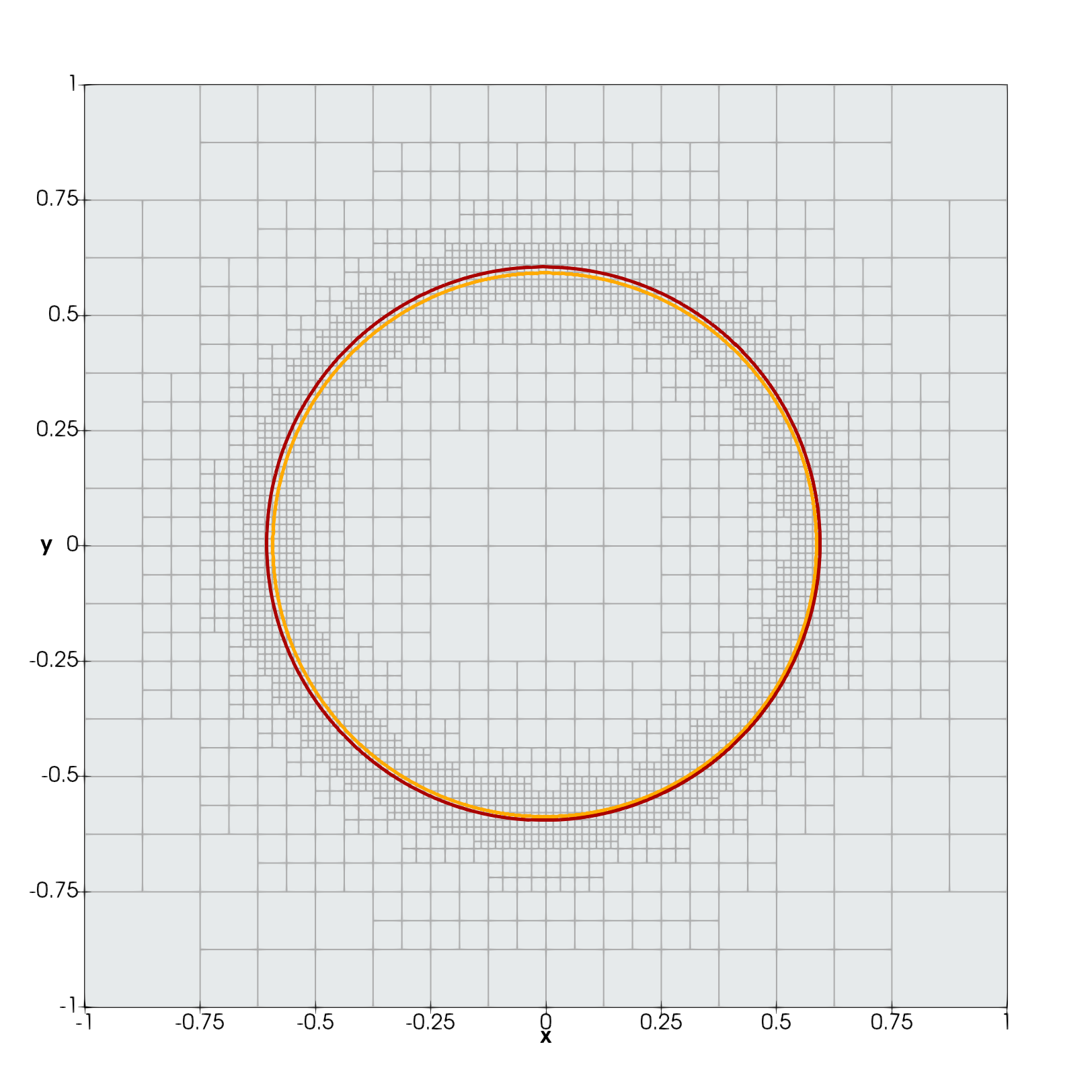}
		\caption{\footnotesize Numerical solution for $h_c = 2^{-7}$}
		\label{fig:results.shearflow.cvp.stress.num7}
	\end{subfigure}
    
	\caption{Circular-vortex-patch zero level sets at $t^{end} = 30\pi r_{cvp}$.  The baseline appears in the left and right panels in green and orange for $\ell_c^{\max}=6,$ and $7$.  Our solution is shown in the center panel in blue.  We display the analytical contour (i.e., $D_{cvp}$) in all charts in red.  (Color online.)}
	\label{fig:results.shearflow.cvp.stress}
\end{figure}

Suppose we stress the baseline and the hybrid solvers with $k = 15$ revolutions so that $t^{end} = 30\pi r_{cvp}$.  If we track the boundary evolution and the error norms in \cref{eq:errornorms}, we get the charts shown in \cref{fig:results.shearflow.cvp.stress.stats}.  \Cref{fig:results.shearflow.cvp.stress} supplements these results by contouring $\Gamma$ and $D_{cvp}$ at the end of the fifteenth rotation.  Clearly, our strategy outperforms the baseline in the long run, even at twice the grid resolution.  This phenomenon has been reported in \cite{Zhuang;etal;LrndDiscForPassSclrAdvctn2D;2021}, where neurally computed transport profiles gradually reached a steady state and then maintained those shapes forever.  In our case, the alternating mechanism discussed above seems to lock the interface in equilibrium with the velocity field, which keeps $\Gamma$'s area stable and independent of time.  Although the blue circle in \cref{fig:results.shearflow.cvp.stress.nnet6} is not as regular as the orange disk in \cref{fig:results.shearflow.cvp.stress.num7}, our {\tt MLSemiLagrangian()} subroutine can reduce advection costs by a factor of $3.65$, as seen in \cref{tbl:results.shearflow.cvp.stats}.

Our findings in this experiment expose a certain degree of overfitting in our error-correcting strategy.  In the beginning, $\mathcal{F}_{c,f}(\cdot)$ cannot recognize the tangential shear-flow patterns.  Thus, the machine-learning-corrected trajectory's precision declines fast as $\Gamma$ undergoes mild deformations.  These deformations and velocity patterns eventually match $\mathcal{F}_{c,f}(\cdot)$'s training tuples.  Then, the mass-conservation scheme kicks in to preserve the blob's area indefinitely.  In a sense, our neural network transforms the disk into a profile it knows how to handle.  Therefore, if we must improve {\tt MLSemiLagrangian()}'s robustness, we should start by integrating more velocity patterns and interface types into the training process.  Also, we should make our method consistent by adding detection mechanisms to fall back to the numerical approximation whenever the latter is sufficiently accurate.  This task might involve investigating neural classifiers (e.g., \cite{TroubledCellIndicator18, ShockDetector20, Buhendwa;Bezgin;Adams;IRinLSwithML;2021}) or some gradient-dependent quality metric (e.g., \cite{Macklin;Lowengrub;ImprovedCurvatureAppTumorGrowth;2006, Lervag;CalcCurvatureLSM;2014, Ervik;Lervag;Munkejord;LOLEX;2014}) to recognize well-resolved stencils.


\colorsubsection{Stefan problem}
\label{subsec:StefanProblem}

We now assess the performance of our hybrid algorithm when solving the phase transition of a single-component liquid melt to a solid structure.  Here, we consider a diffusion-dominated crystallization process that we can model as a Stefan problem.  To state this FBP, suppose we split the computational domain $\Omega$ into two subdomains, $\Omega_s$ and $\Omega_l$, separated by a solidification front $\Gamma$.  The Stefan problem describes the evolution of the scalar temperature field $T$, decomposed into $T_s$ for the solid phase $\Omega_s$ and $T_l$ for the liquid phase $\Omega_l$, as\footnote{The subscripts $s$ and $l$ indicate the solid and liquid phases.}

\begin{equation}
\left\{
\begin{split}
	\pde{T_l}{t} &= D_l \Delta T_l, \quad \textrm{in }\Omega_l, \\
	\pde{T_s}{t} &= D_s \Delta T_s, \quad \textrm{in }\Omega_s.
\end{split}
\right.
\label{eq:StefanProblem}
\end{equation}
Generally, the $D_l$ and $D_s$ diffusion constants can be discontinuous across $\Gamma$.  Moreover, the temperature on the solid-liquid interface satisfies the Gibbs--Thomson boundary condition \cite{Alexiades;Solomon;Wilson:88:The-formation-of-a-s, Alexiades;Solomon:93:Mathematical-Modelin}

\begin{equation}
T_s = T_l = T_\Gamma = -\epsilon_c\kappa - \epsilon_v \left(\vv{u}_{sp} \cdot \vv{n}\right).
\label{eq:GibbsThomsonCondition}
\end{equation}
In this expression, $\vv{n}$ is the outward unit vector normal to the solidification front, $\vv{u}_{sp}$ denotes the interface velocity, $\kappa$ is the local curvature at the interface, and $\epsilon_c$ and $\epsilon_v$ are the surface tension and molecular kinematics supercooling coefficients.  As for the normal velocity component, we can compute it in terms of the jump in the heat flux at $\Gamma$ with

\begin{equation}
\vv{u}_{sp} \cdot \vv{n} = -\left[ \nabla_{\vv{n}}T \right] = -\left( D_l\nabla_{\vv{n}}T_l - D_s\nabla_{\vv{n}}T_s \right),
\label{eq:StefanNormalVelComponent}
\end{equation}
where $\nabla_{\vv{n}}T$ is the directional derivative $\vv{n} \cdot \nabla T$.  For the practical implementation of the Stefan problem in the parallel level-set framework, we refer the reader to Algorithm \href{https://www.sciencedirect.com/science/article/pii/S002199911630242X\#fg0140}{5} in \cite{Mirzadeh;etal:16:Parallel-level-set}.  The latter extends the adaptive-Cartesian-grid-based Algorithm 2 in \cite{Chen;Min;Gibou:09:A-numerical-scheme-f}.  In addition, we have assumed $D_s = D_l = 1$ and used {\tt PETSc}'s BiCGStab {\tt KSPSolve} function \cite{Balay;Brown;Buschelman;etal:12:PETSc-Web-page} to solve the system arising from \cref{eq:StefanProblem}.


\colorsubsubsection{Frank-sphere}
\label{subsubsec:FrankSphere}

First, we evaluate our system's response to the Frank-sphere problem \cite{FrankSphere;1950}, which has a known analytical solution.  In this FBP, a region described by a disk with $T_s = 0$ grows into a supercooled liquid phase.  As time progresses, the radius expands according to $R(t) = s_0 \sqrt{t}$, where $s_0$ is the base parameter.  Likewise, we define the temperature field by

\begin{equation}
T(r, t) = T(s) = \left\{
\begin{array}{ll}
	0                                            & \textrm{if } s \leqslant s_0 \\
	T_\infty\left(1 - \frac{F(s)}{F(s_0)}\right) & \textrm{if } s > s_0
\end{array}
\right.,
\label{eq:FrankSphere}
\end{equation}
where $r$ is the distance to the disk's center, $s = r/\sqrt{t}$, and $F(s) = E_1(s^2/4)$.  Further, $E_1(z) = \int_z^\infty{\frac{\textrm{e}^{-t}}{t}\textrm{d}t}$ is the exponential integral, and F(s) is the heat equation's similarity solution.  Also, $T_\infty$ denotes the temperature infinitely far away from $\Gamma$.  To determine $T_\infty$, we combine the jump condition in \cref{eq:StefanNormalVelComponent} with the relation

\begin{equation}
\vv{u}_{sp} \cdot \vv{n} = \ode{R}{t} = \frac{s_0}{2\sqrt{t}}
\label{eq:FrankSphereNormalVelComponent}
\end{equation}
at the freezing front so that

\begin{equation}
T_\infty = \frac{s_0 F(s_0)}{2F'(s_0)}.
\label{eq:FrankSphereTInfty}
\end{equation}

In this experiment, $\phi_{fs}(\vv{x}) = \|\vv{x} - \vv{x}_{fs}\| - r_{fs}$ is our initial level-set function embedded in $\Omega \equiv [-1, +1]^2$ with a circular interface centered at a random location $\vv{x}_{fs}$ inside $[-h_c/2, +h_c/2]^2$.  If we define $s_0 = 0.5$ and $t^0 = 0.25$, we get $T_\infty = -0.15015425523242384$, and the freezing front radius $r_{fs}$ starts at $0.25$.  With these parameters, we ensure that $\max{||\vv{u}_{sp}(\vv{x})||} \leqslant 1$ for all $\vv{x} \in \Omega$ upon quadratic extension from $\Gamma$.  In addition, we enforce the time-step restriction $\Delta t = h_c$ and the Dirichlet boundary condition $T_\Gamma = 0$.  

\begin{figure}[t]
	\centering
	\begin{subfigure}[b]{0.32\textwidth}
		\includegraphics[width=\textwidth]{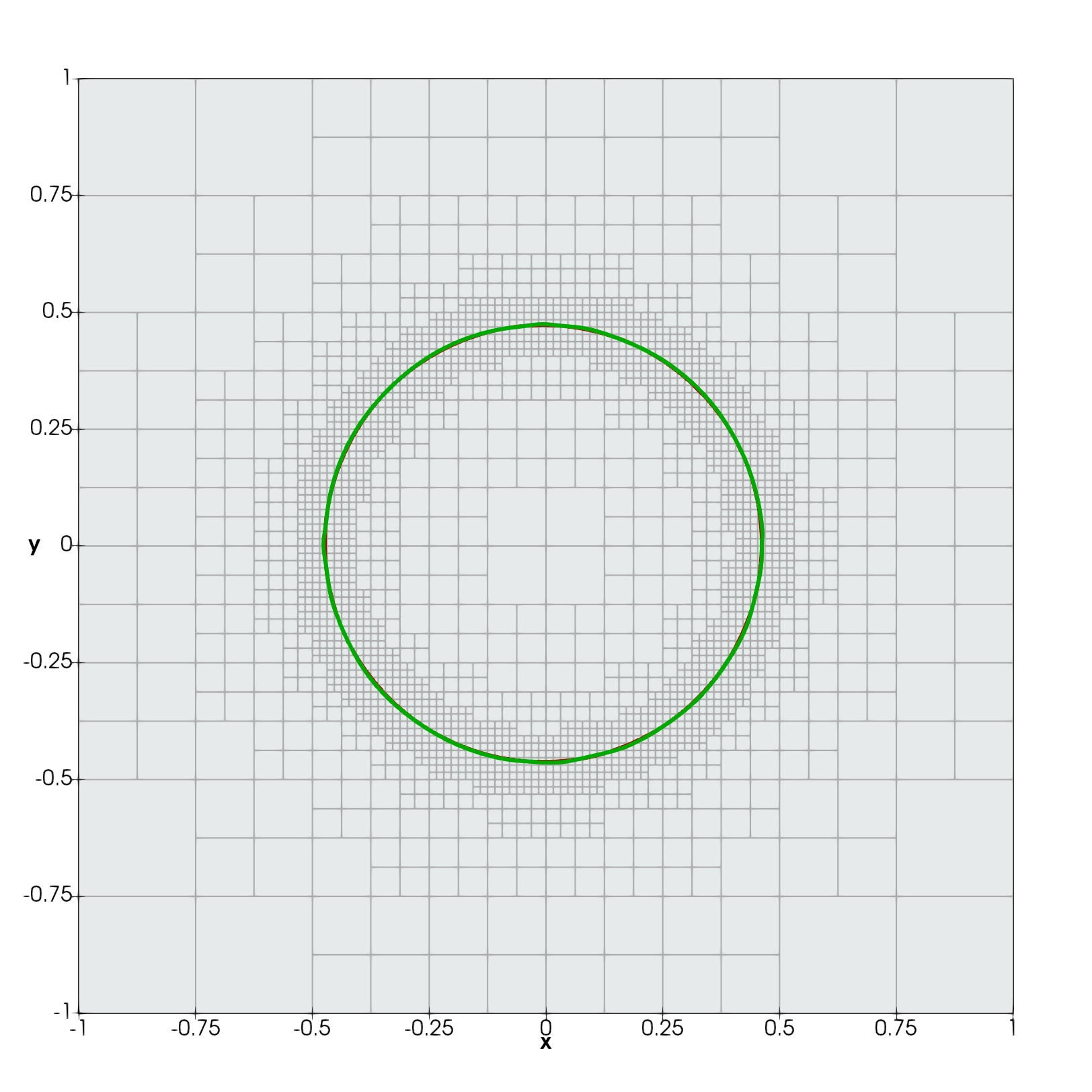}
        \caption{\footnotesize Numerical baseline for $h_c = 2^{-6}$}
        \label{fig:results.franksphere.num6}
    \end{subfigure}
    ~
	\begin{subfigure}[b]{0.32\textwidth}
		\includegraphics[width=\textwidth]{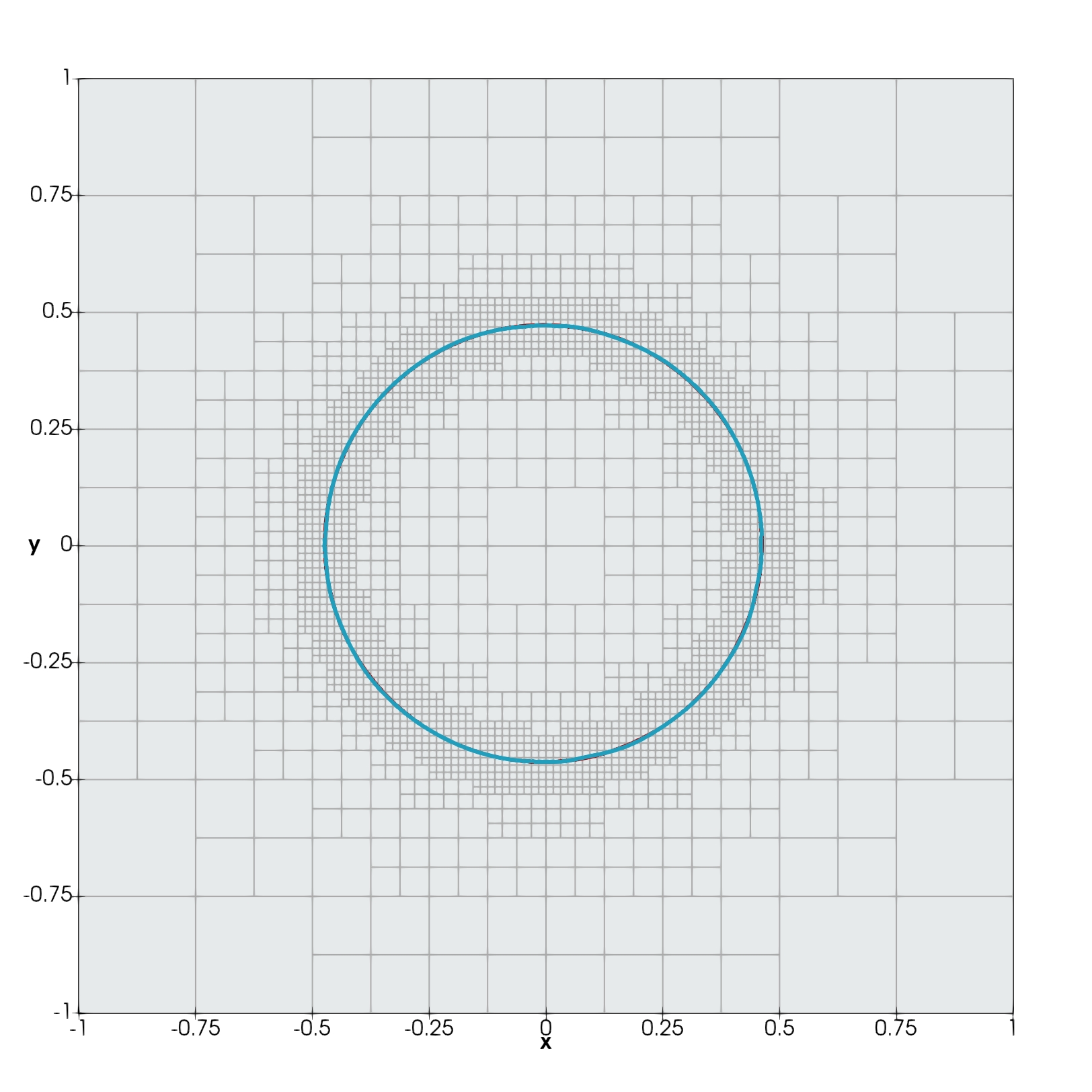}
        \caption{\footnotesize Hybrid approach for $h_c = 2^{-6}$}
        \label{fig:results.franksphere.nnet6}
    \end{subfigure}
    ~
	\begin{subfigure}[b]{0.32\textwidth}
		\includegraphics[width=\textwidth]{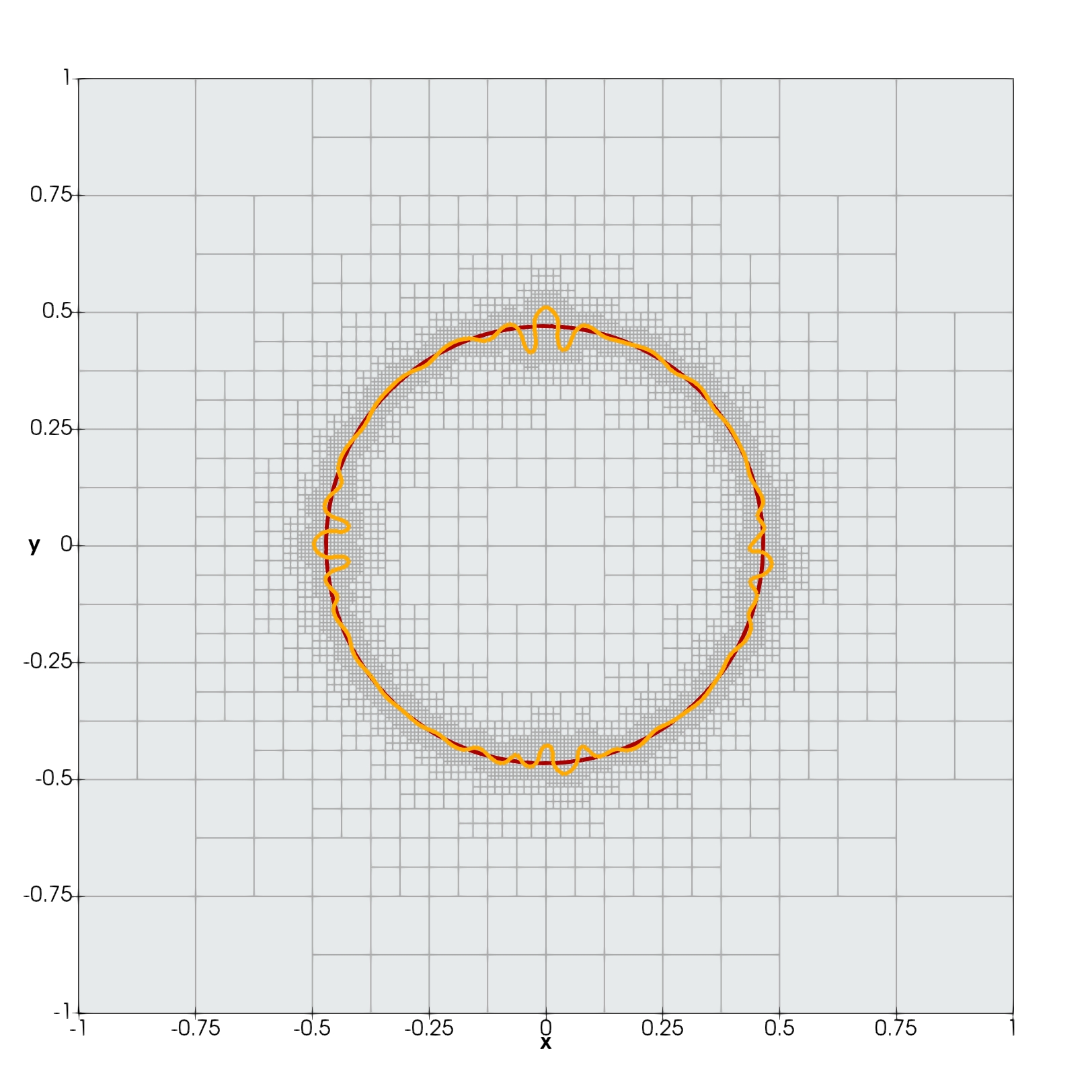}
        \caption{\footnotesize Numerical scheme for $h_c = 2^{-7}$}
        \label{fig:results.franksphere.num7}
    \end{subfigure}
    
	\caption{Frank-sphere zero level sets at $t^{end} = 0.875$.  The left and center panels show the numerical baseline in green and the machine-learning-corrected solution in blue for a discretization with $\ell_c^{\max} = 6$.  For comparison, we include the numerical result for a grid with twice the resolution in the right panel in orange.  In all cases, the analytical disk appears in red.  (Color online.)}
	\label{fig:results.franksphere}
\end{figure}

\Cref{tbl:results.franksphere.stats} summarizes the interface-location and temperature error statistics for Frank-sphere simulations running until $t^{end} = 0.875$.  The table also shows the area loss and the running wall time at resolutions varying from $\ell^{\max}_c = 6$ to $\ell^{\max}_c = 8$.  Because of the inherent instability in the Frank-sphere problem, the free boundary develops dendritic structures as the mesh size decreases.  However, for $h_c = 2^{-6}$, our strategy agrees with the baseline trajectory while exhibiting better area conservation and temperature accuracy.  \Cref{fig:results.franksphere} supplements these observations by contrasting the zero-isocontours computed with the numerical solver and the hybrid system.  This test certifies the feasibility of our approach for building solutions for FBPs where the velocity field is not necessarily divergence-free.

\begin{table}[t]
	\centering
	\small
	\bgroup
	\def\arraystretch{1.1}%
	\begin{tabular}{|l|c|c|c|c|c|r|r|}
		\hline
		Method & $\ell_c^{\max}$ & $\ell^1$ error & $\ell^\infty$ error & Temperature error & Disk area & Area loss (\%) & Time (sec.) \\
		\hline \hline
		Ours                       & 6 & $\eten{5.853}{-4}$ &  $\eten{1.762}{-3}$ & $\eten{4.225}{-4}$ & $\eten{6.876}{-1}$ & -0.05  &  4.342 \\
		\hline
		\multirow{3}{*}{Numerical} & 6 & $\eten{8.112}{-4}$ &  $\eten{3.625}{-3}$ & $\eten{6.982}{-4}$ & $\eten{6.892}{-1}$ & -0.29  &  3.497 \\
 		                           & 7 & $\eten{9.253}{-3}$ &  $\eten{5.614}{-2}$ & $\eten{9.137}{-3}$ & $\eten{6.846}{-1}$ &  0.38  &  9.253 \\
 		                           & 8 & $\eten{5.039}{-2}$ &  $\eten{1.511}{-1}$ & $\eten{3.405}{-2}$ & $\eten{6.879}{-1}$ & -0.10  & 92.190 \\
		\hline
	\end{tabular}
	\egroup
	\caption{Frank-sphere accuracy and performance statistics.  Reported errors include measurements taken at grid points for which $|\phi_{vtx}(\vv{x})| \leqslant \sqrt{2}h_c$ at $t^{end} = 0.875$.}
	\label{tbl:results.franksphere.stats}
\end{table}

\colorsubsubsection{Unstable solidification}
\label{subsubsec:UnstableSolidification}

At last, we evaluate our approach's response to curvature-driven, anisotropic crystallization.  To this end, we place a solid seed into a supercooled liquid and solve \cref{eq:StefanProblem} with a nonzero Gibbs--Thomson condition (i.e., $T_\Gamma \neq 0$).  Here, we consider the domain $\Omega \equiv [-1, +1]^2$ with a temperature field initialized uniformly as $T_l = -0.08$ for the liquid phase and $T_s = 0$ for the seed.  The nucleus has a characteristic three-fold freezing front given by the polar-rose equation

\begin{equation}
r(\theta) = a \cos\left( p\theta \right) + b,
\label{eq:AnisotropicPolarRose}
\end{equation}
where $a = 0.02$, $b = 0.09$, $p = 3$, and $\theta \in [0, 2\pi)$.  Furthermore, we have applied the condition in \cref{eq:GibbsThomsonCondition} using 

\begin{equation}
\epsilon_c = \eten{2}{-4}\left[1 + 0.6\cos\left( 4 \left(\theta_{\vv{n}} + \tfrac{\pi}{3} \right)\right)\right] \quad \textrm{and} \quad \epsilon_v = 0,
\label{eq:AnisotropicConditions}
\end{equation} 
where $\theta_{\vv{n}}$ is the interface normal-vector angle with respect to the horizontal.  Also, we have imposed adiabatic boundary conditions for the four sides of $\Omega$.  With this configuration, we have carried out the unstable-solidification simulation from $t^0 = 0$ to $t^{end} = 3$.

\begin{figure}[t]
	\centering
	\begin{subfigure}[b]{0.32\textwidth}
		\includegraphics[width=\textwidth]{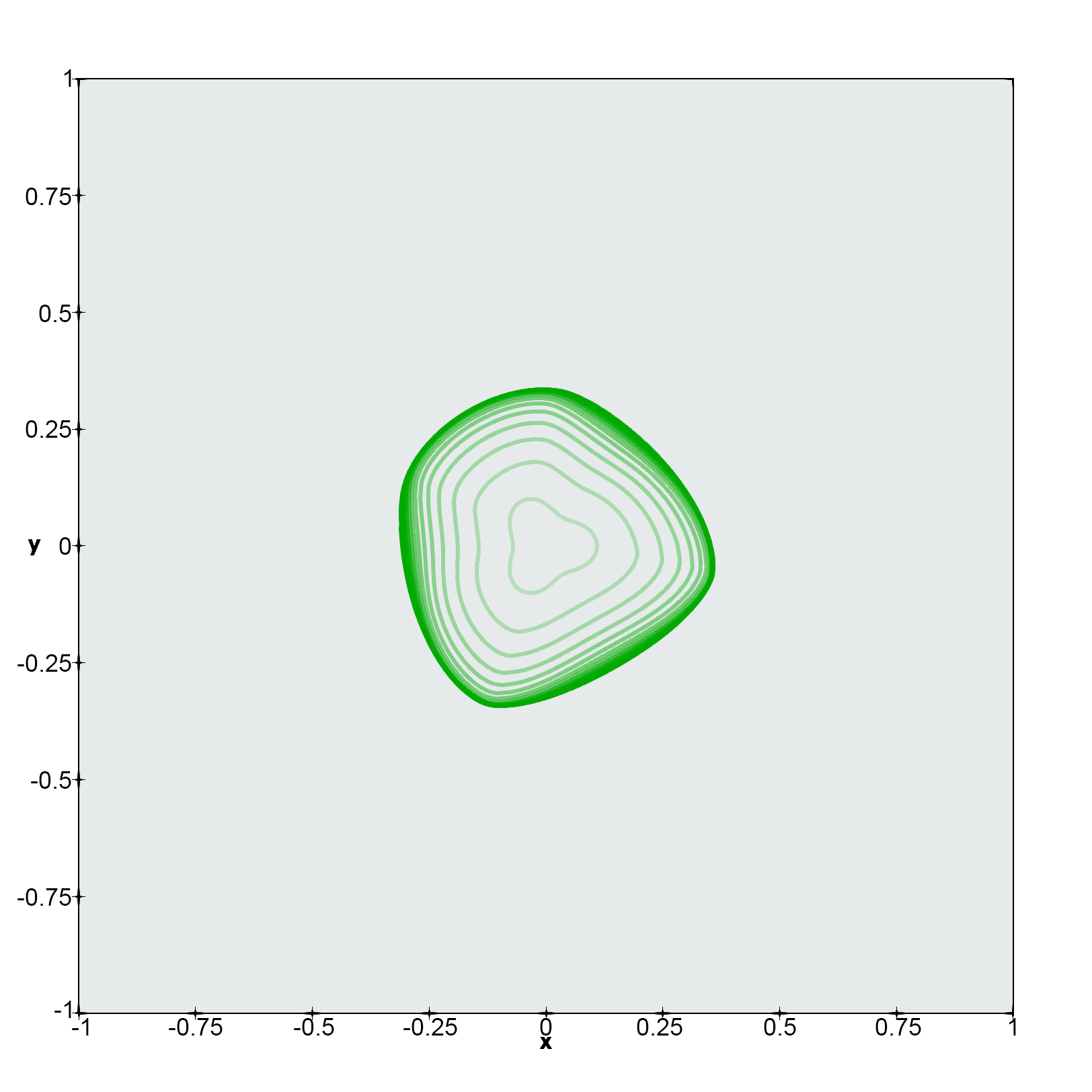}
        \caption{\footnotesize Numerical baseline for $h_c = 2^{-6}$}
        \label{fig:results.anisotropic.num6}
    \end{subfigure}
    ~
	\begin{subfigure}[b]{0.32\textwidth}
		\includegraphics[width=\textwidth]{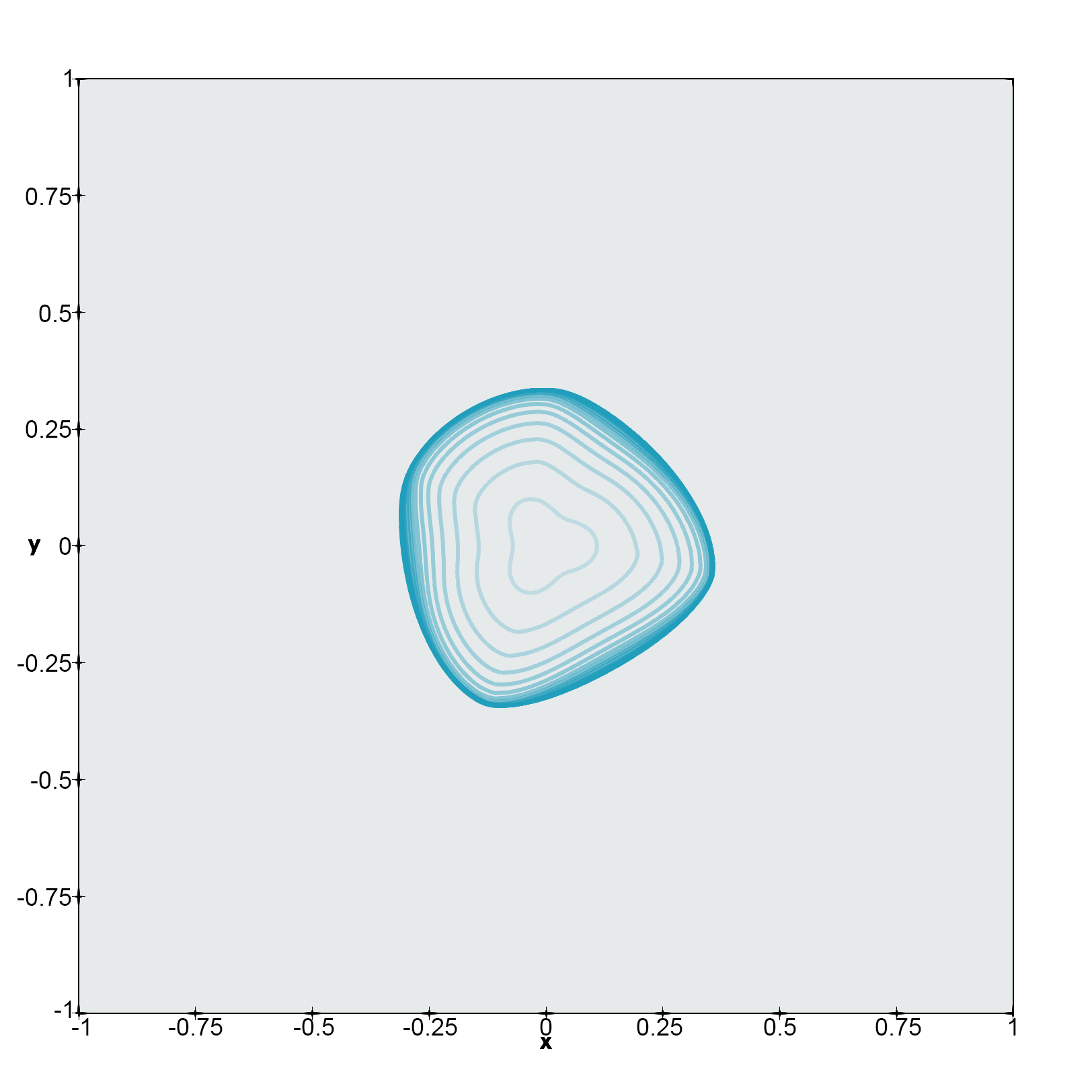}
        \caption{\footnotesize Hybrid approach for $h_c = 2^{-6}$}
        \label{fig:results.anisotropic.nnet6}
    \end{subfigure}
    ~
    \begin{subfigure}[b]{0.32\textwidth}
		\includegraphics[width=\textwidth]{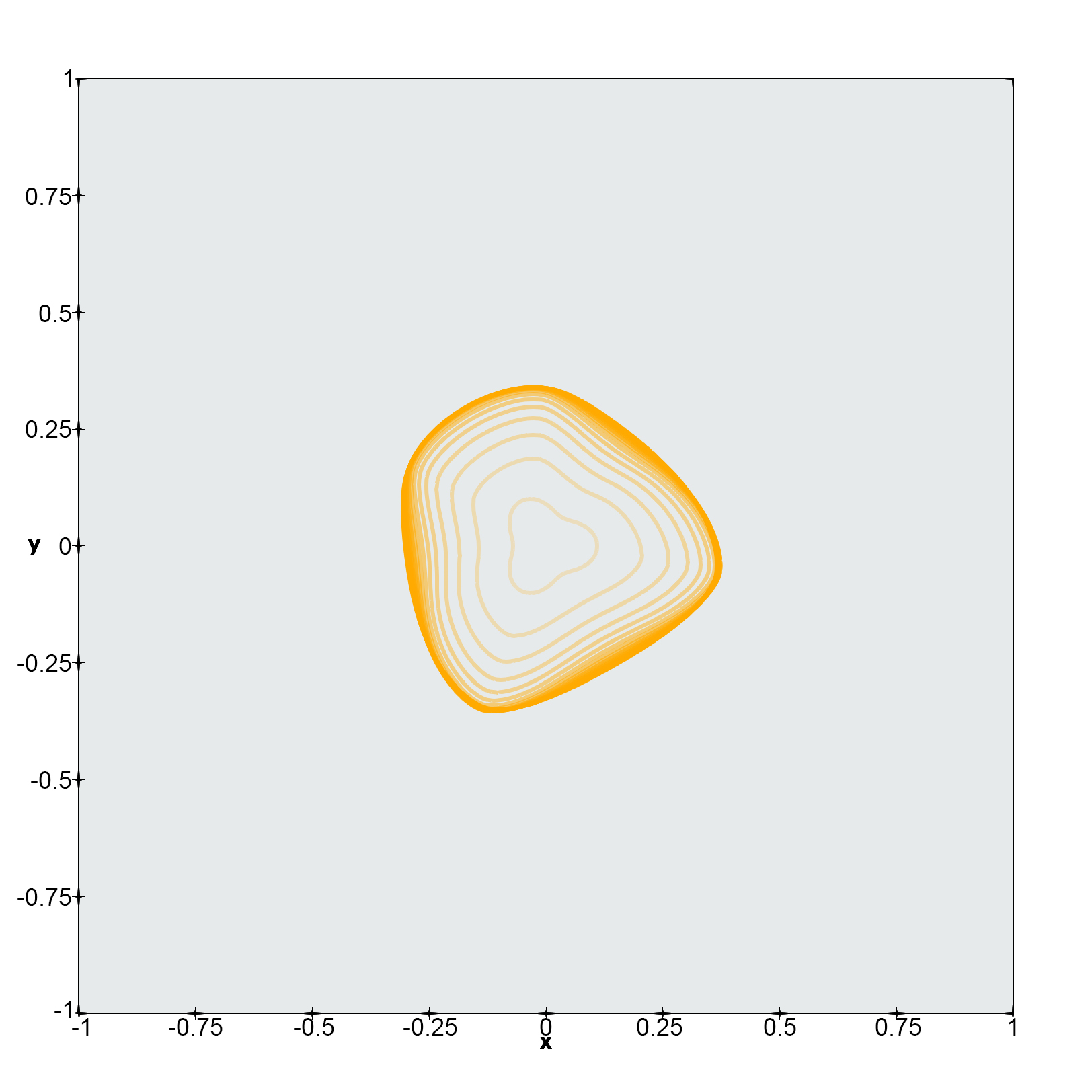}
        \caption{\footnotesize Numerical scheme for $h_c = 2^{-7}$}
        \label{fig:results.anisotropic.num7}
    \end{subfigure}
    
	\caption{Unstable-solidification zero-level-set growth histories with sixteen steps from $t^0 = 0$ to $t^{end} = 3$ for $\ell_c^{\max} = 6, 7$.  The (green and orange) numerical trajectories appear on the left and right panels.  We show our hybrid approach solution in the center chart in blue.  (Color online.)}
	\label{fig:results.anisotropic}
\end{figure}

\begin{figure}[t]
	\centering
	\includegraphics[width=7.5cm]{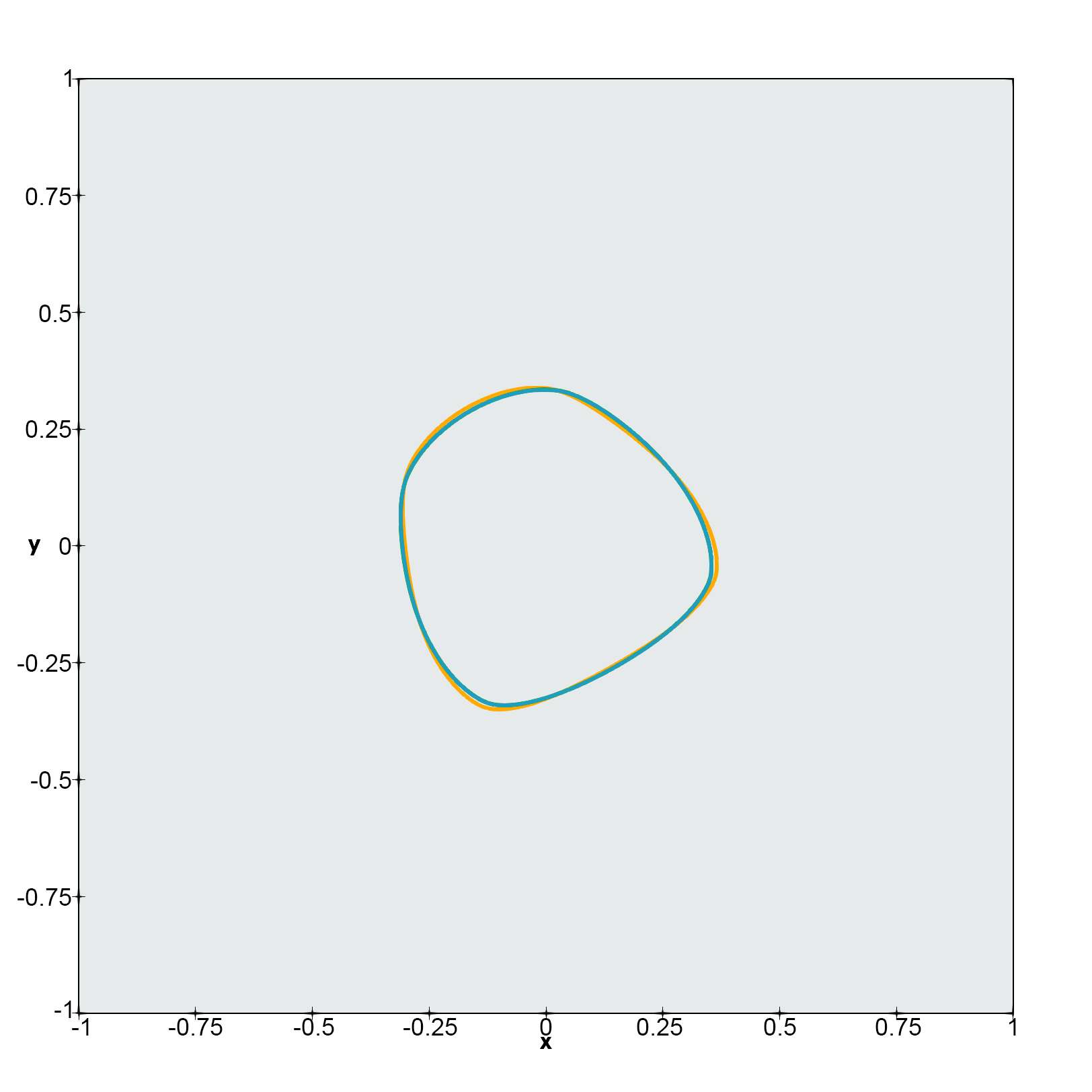}
	\caption{Comparison of the unstable-solidification zero level sets at time $t^{end} = 3$.  The numerical interfaces appear in green ($\ell_c^{\max} = 6$) and orange ($\ell_c^{\max} = 7$), while our hybrid approach solution is shown in blue for $h_c = 2^{-6}$.  The machine-learning-corrected and the standard zero-isocontours practically overlap.  (Color online.)}
	\label{fig:results.anisotropic.final}
\end{figure}

\Cref{fig:results.anisotropic,fig:results.anisotropic.final} compare the numerical and machine-learning-corrected growth histories and final zero level sets for $h_c = 2^{-6}$ and $h_c = 2^{-7}$.  In this case, we have used an adaptive time step 

\begin{equation}
\Delta t = h_c \min{\left(1, \tfrac{1}{\max{||\vv{u}_{sp}||}}\right)},
\label{eq:AnisotropicTimeStep}
\end{equation}
which always leads to an $\mathcal{F}_{6,8}(\cdot)$-compatible $\Delta t = h_c$ for $\ell_c^{\max} = 6$ but not for $\ell_c^{\max} = 7$.  Qualitatively, the results in \cref{fig:results.anisotropic,fig:results.anisotropic.final} show that our hybrid solver converges to the baseline solutions at both levels of refinement.  These plots thus suggest that our hybrid strategy yields trajectories that agree with the numerical patterns obtained with $\ell_c^{\max} = 6$ and $\ell_c^{\max} = 7$ for curvature-driven simulations.  But, to understand better the machine learning effects on crystallization, further research is necessary to relax the velocity maximum-unit-norm restriction.  Possible paths to address this problem involve increasing the variety of speed patterns in $\mathcal{D}$ or introducing velocity embeddings or encodings.

\colorsection{Conclusions}
\label{sec:Conclusions}

We have introduced a hybrid machine learning strategy to improve the accuracy of semi-Lagrangian schemes for relatively coarse adaptive Cartesian grids.  Our approach comprises a multilayer perceptron $\mathcal{F}_{c,f}(\cdot)$ that processes velocity, level-set, and positional information to quantify the error in the interface trajectory.  To develop this model, we have postulated the problem according to the image super-resolution construct \cite{Dong;Loy;He;SuperResolution;2014}.  The latter requires employing a high-resolution mesh to supply reference level-set values for neural optimization.  Once trained, $\mathcal{F}_{c,f}(\cdot)$ operates alongside selective reinitialization and alternates with standard advection to reduce numerical diffusion.  Unlike other machine learning solutions for passive scalar transport (e.g., \cite{Farimani;Gomes;Pande;DLPhysTransPhenomena;2017, Xie;etal;TempoGAN;2018, Liu;etal;DLMthdsSuperResoltnReconstTurbFlows;2020, Pathak;etal;MLToAugCoarseGridCFD;2020, Zhuang;etal;LrndDiscForPassSclrAdvctn2D;2021}), our approach is localized in time and space.  In particular, we have avoided complex architectures and costly evaluations by concentrating our effort only on vertices next to the interface.  Also, our error-correcting neural network consumes data from the current time frame, thus relieving the system from buffering past states.  Together, these features make our strategy attractive to parallel level-set frameworks (e.g., \cite{Mirzadeh;etal:16:Parallel-level-set}), where communication among computing nodes can be expensive.

We have also elaborated on the methodologies to assemble the learning data set and train the error-correcting neural network.  Our main contribution is a novel machine-learning-augmented semi-Lagrangian algorithm that blends neural inference with numerically advected level-set values in an alternating fashion.  To validate its correctness, we have examined the entire strategy with a few representative test cases.  These included rotating and warping a disk, evaluating the effects of tangential shear flows, and two instances of the Stefan problem.  Our results confirm that $\mathcal{F}_{c,f}(\cdot)$ can dampen numerical diffusion and improve mass loss in simple advection problems.  More precisely, our strategy's accuracy is often as good as or better than the numerical scheme at twice the resolution while requiring only a fraction of the cost.  Likewise, we have shown that our framework produces feasible solidification fronts for crystallization processes.  However, bias artifacts and significant deterioration have occurred in lengthy or highly deforming simulations or in the presence of tangential shear flows.  In addition, the stringent velocity maximum-unit-norm condition imposes severe limitations for FBPs involving rapid interface changes.

Unwelcome artifacts, unexpected results, and limitations in our case studies reveal several opportunities for future investigation.  For example, we could enrich the learning data set by incorporating more than a single interface type (e.g., sine waves, as in \cite{LALariosFGibou;LSCurvatureML;2021}) and more velocity patterns, emphasizing under-resolved sampling.  Other promising avenues for enhancing robustness involve binding physical constraints to $\mathcal{F}_{c,f}(\cdot)$ \cite{Raissi17a, Raissi18, Raissi;PINN;2019, Beucler;etal;EnforcingAnalyticConstNnets;2021}, integrating velocity embeddings, adversarial training \cite{GAdversarialNets14, Farimani;Gomes;Pande;DLPhysTransPhenomena;2017, Xie;etal;TempoGAN;2018}, enforcing consistency by detecting well-resolved stencils \cite{Macklin;Lowengrub;ImprovedCurvatureAppTumorGrowth;2006, Lervag;CalcCurvatureLSM;2014, Ervik;Lervag;Munkejord;LOLEX;2014, TroubledCellIndicator18, ShockDetector20, Buhendwa;Bezgin;Adams;IRinLSwithML;2021}), and even replacing the semi-Lagrangian formulation with a more accurate method.  Similarly, we should consider temporal data only if it does not set back {\tt MLSemiLagrangian()}'s performance.  Then, we could evaluate our solver on more complex flows, such as the Kirchhoff elliptical vortex \cite{Kirchhoff;Vortex;1876} and Chaplygin's vortical structures \cite{Chaplygin;CilyndricalVortex;1899, Chaplygin;VortexMotionFluid;1903, Meleshko;VanHeist;OnChaplyginVortex;1994}.  As a separate, challenging task, it also remains to migrate our strategy to higher resolutions and three-dimensional domains.  In the meantime, we echo the arguments and motivation in \cite{Pathak;etal;MLToAugCoarseGridCFD;2020} and hope our contributions open up further interdisciplinary research between machine learning and scientific computing.


\color{navy}\sffamily\section*{Acknowledgements}\rmfamily\color{black}

Use was made of computational facilities purchased with funds from the National Science Foundation (CNS-1725797) and administered by the Center for Scientific Computing (CSC).  The CSC is supported by the California NanoSystems Institute and the Materials Research Science and Engineering Center (MRSEC; NSF DMR 1720256) at UC Santa Barbara.



{\footnotesize
\biboptions{sort&compress}
\bibliographystyle{unsrt}
\bibliography{references}}

\end{document}